\RequirePackage{fix-cm}
\documentclass[twocolumn]{svjour3}          
\smartqed  

\usepackage{graphicx}
\usepackage{times}
\usepackage{helvet}
\usepackage{courier}
\usepackage{amsmath,epsfig}
\usepackage{graphicx}
\usepackage{bm}
\usepackage{float,caption,multirow}
\usepackage{makecell}
\usepackage{url,caption}
\usepackage{multirow}
\usepackage{makecell}
\usepackage{wrapfig}
\usepackage{amssymb}
\usepackage{epstopdf}
\usepackage{booktabs} 
\usepackage{epsfig}

\usepackage{multirow}
\usepackage{xcolor}
\usepackage{helvet,url}
\usepackage{courier}
\usepackage{epsfig}
\usepackage{picinpar}
\usepackage{graphicx}
\usepackage{epstopdf}
\usepackage{bm}
\usepackage{epsfig}
\usepackage{picinpar}
\usepackage{bm}
\usepackage{float,caption,multirow}
\usepackage{marvosym}
\usepackage{subcaption}

\begin{document}
\sloppy
\title{Practical Video Object Detection via Feature Selection and Aggregation}




\author{Yuheng Shi$^\dagger$         \and
        Tong Zhang$^\dagger$       \and
        Xiaojie Guo               
}

\institute{\Letter\ Xiaojie Guo \at
             \email{xj.max.guo@gmail.com}
             \and
           Yuheng Shi$^\dagger$ \at 
              \email{yuheng@tju.edu.cn}            
           \and
           Tong Zhang$^\dagger$ \at
           \email{cstongzhang@tju.edu.cn}
           \and
             The authors are from the College of Intelligence and Computing, Tianjin University, Tianjin 300350, China. $^\dagger$ These authors contributed equally to this work.\\
             This work was supported by the National Natural Science Foundation of China under Grants No.62072327 and 62372251.
}

\date{Received: date / Accepted: date}

\maketitle
\abstract{
Compared with still image object detection, video object detection (VOD) needs to particularly concern the high across-frame variation in object appearance, and the diverse deterioration in some frames. In principle, the detection in a certain frame of a video can benefit from information in other frames. Thus, how to effectively aggregate features across different frames is key to the target problem. Most of contemporary aggregation methods are tailored for two-stage detectors, suffering from high computational costs due to the dual-stage nature. On the other hand, although one-stage detectors have made continuous progress in handling static images, their applicability to VOD lacks sufficient exploration. To tackle the above issues, this study invents a very simple yet potent strategy of feature selection and aggregation, gaining significant accuracy at marginal computational expense. Concretely, for cutting the massive computation and memory consumption from the dense prediction characteristic of one-stage object detectors, we first condense candidate features from dense prediction maps. Then, the relationship between a target frame and its reference frames is evaluated to guide the aggregation. Comprehensive experiments and ablation studies are conducted to validate the efficacy of our design, and showcase its advantage over other cutting-edge VOD methods in both effectiveness and efficiency. Notably, our model reaches \emph{a new record performance, i.e., 92.9\% AP50 at over 30 FPS on the ImageNet VID dataset on a single 3090 GPU}, making it a compelling option for large-scale or real-time applications. The implementation is simple, and accessible at \url{https://github.com/YuHengsss/YOLOV}.}



\begin{figure}[t]
\centering
\includegraphics[width=1.0\linewidth]{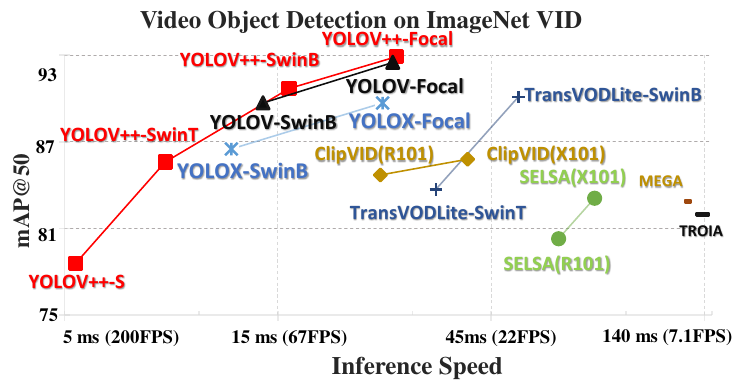}
\caption{Performance comparison in accuracy (AP50) and inference speed (FPS) on a 3090 GPU device. }
\label{fig:performance}
\end{figure}

\section{Introduction}\label{sec1}

Object detection, as a key component in a wide spectrum of vision-based intelligent applications~\cite{dalal2005histograms,felzenszwalb2008discriminatively}, aims to simultaneously locate and classify objects within images. Along with the emergence of deep learning~\cite{krizhevsky2012imagenet,dosovitskiy2020image}, numerous still image object detection models have been proposed, which can be broadly categorized into two-stage and one-stage object detectors according to their detection procedures. 

\begin{figure*}[t]
\centering
\subfloat{\includegraphics[width=\dimexpr\textwidth/4\relax]{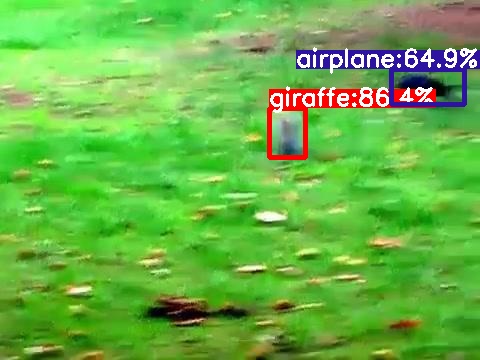}}\hfill
\subfloat{\includegraphics[width=\dimexpr\textwidth/4\relax]{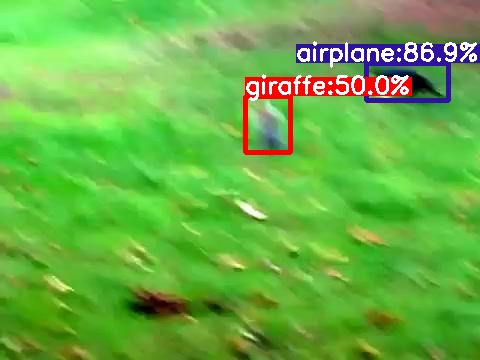}}\hfill
\subfloat{\includegraphics[width=\dimexpr\textwidth/4\relax]{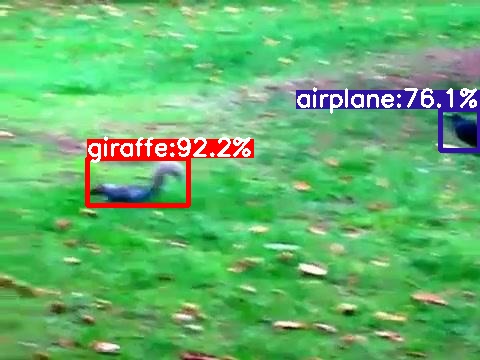}}\hfill
\subfloat{\includegraphics[width=\dimexpr\textwidth/4\relax]{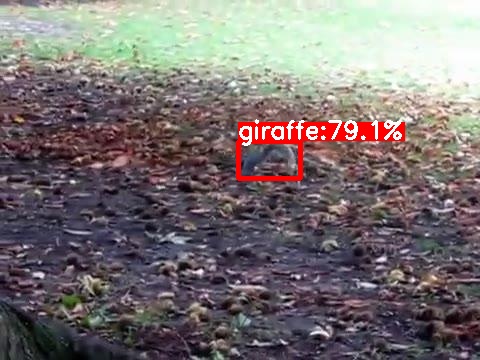}}

\subfloat{\includegraphics[width=\dimexpr\textwidth/4\relax]{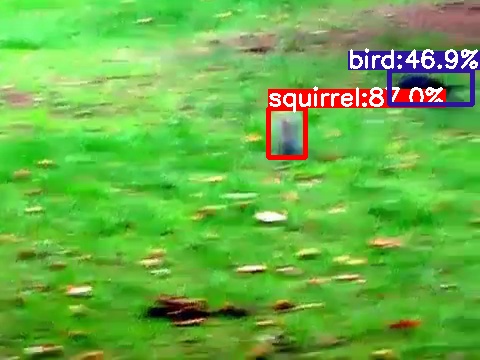}}\hfill
\subfloat{\includegraphics[width=\dimexpr\textwidth/4\relax]{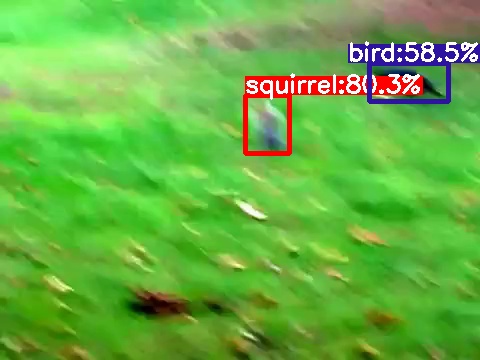}}\hfill
\subfloat{\includegraphics[width=\dimexpr\textwidth/4\relax]{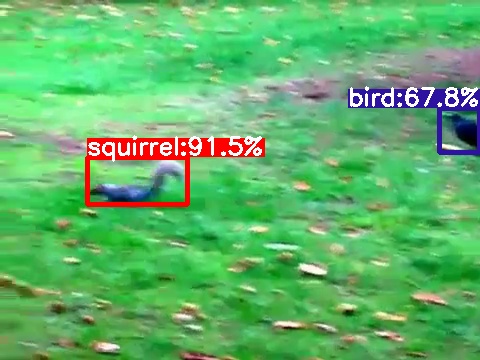}}\hfill
\subfloat{\includegraphics[width=\dimexpr\textwidth/4\relax]{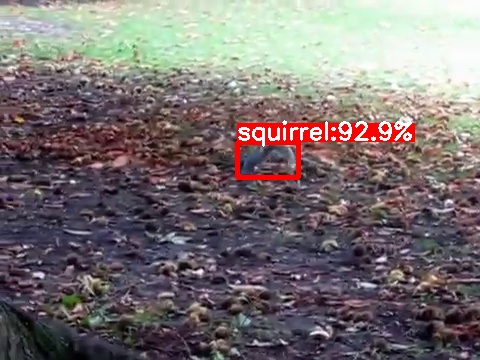}}
\caption{A comparison between the base detector YOLOX (1st row) and our method (2nd row). The frames suffer from various kinds of interference, like non-rigid motions, motion blur and challenging poses, making the base detector fail to accomplish the task. Our method can precisely predict the objects.}
\label{fig:Base_compare}
\end{figure*}

As for two-stage detectors, they initially select possible object regions (\emph{a.k.a.} proposals), followed by classifying these regions. Region-based CNN (R-CNN)~\cite{girshick2014rich,girshick2015fast,ren2015faster} pioneer the technical route of two-stage object detection with a variety of follow-ups~\cite{he2016deep,lin2017feature,dai2016r,cai2018cascade,he2017mask,liu2018path}, remarkably boosting the accuracy of detection. Leveraging region-level features, these detectors for static images can be readily adapted to more complex tasks such as segmentation and video object detection~\cite{wu2019sequence,deng2019relation,chen2020memory,gong2021temporal}. However, the underlying two-stage nature poses a bottleneck in efficiency for practical scenarios. 
While for one-stage object detectors, the location and classification are jointly and directly produced through dense prediction from feature maps. The YOLO family~\cite{redmon2016you,redmon2017yolo9000,bochkovskiy2020yolov4} and SSD~\cite{liu2016ssd} are representatives in this category. 
Subsequent developments~\cite{lin2017focal,ge2021yolox,ge2021ota,tian2019fcos,xu2022pp,Wang_2023_CVPR} have significantly enhanced the precision and efficiency of one-stage detectors. Bypassing the requirement of region proposals as in two-stage approaches, one-stage detectors offer superior speed, making them attractive for real-time applications. 
Motivated by the success of the attention mechanism~\cite{vaswani2017attention} in natural language processing (NLP) tasks, subsequent studies have adapted it for vision tasks, such as image classification~\cite{dosovitskiy2020image,DeiT2021,liu2021swin} and object detection~\cite{carion2020end,zhu2020deformable}, with remarkable success. 
Specifically, DETR~\cite{carion2020end} is arguably the first attempt that applies the Transformer encoder-decoder architecture on the task of object detection, which streamlines the detection pipeline by viewing object detection as a direct set prediction problem. 
Further, Deformable DETR~\cite{zhu2020deformable} was devised to accelerate the training process with a deformable attention module.
Although DETRs~\cite{carion2020end,zhu2020deformable} reduce the complexity compared to two-stage detectors such as R-CNN~\cite{girshick2014rich,girshick2015fast,ren2015faster}, the encoder in DETRs emerges as a computational bottleneck, as demonstrated in~\cite{lin2022d}. 

Video object detection can be viewed as an advanced version of still image object detection. In comparison with still image object detection, VOD shall particularly concern the high across-frame variation in object appearance (\emph{e.g.}, non-rigid object motions and rare poses), and the diverse deterioration (\emph{e.g.}, motion blur and part occlusion) in some frames. Intuitively, one can process video sequences via feeding frames one-by-one into still image object detectors. But, by this means, the temporal information across frames will be wasted, which could be instrumental in mitigating or eliminating the mentioned ambiguity present in a single image.
As depicted in Fig.~\ref{fig:Base_compare}, degradation such as motion blur, non-rigid object transformation, and challenging poses often appears in video frames, significantly increasing the difficulty of detection. For instance, via solely looking at the first frame in Fig.~\ref{fig:Base_compare}, it is hard or even impossible for human beings to tell where and what the objects are. Conversely, video sequences can provide richer information than single still images. In other words, other frames in the same sequence can possibly support the prediction for a certain frame. Hence, \emph{how to effectively aggregate temporal messages from different frames is crucial to the accuracy}. 

Temporal information aggregation can be achieved through various methods, such as box-level post-processing~\cite{han2016seq,sabater2020robust,belhassen2019improving}, optical flow-based~\cite{zhu2017flow,zhu2018towards,wang2018fully}, and tracking-based techniques~\cite{feichtenhofer2017detect,zhang2018integrated}. Recently, attention-based methods~\cite{wu2019sequence,deng2019relation,chen2020memory,gong2021temporal,zhou2022transvod} have come to prominence due to the simplicity and exceptional ability to capture long-range dependencies. 
The implementation of attention for feature aggregation varies across different detection frameworks. In~\cite{wu2019sequence,chen2020memory,gong2021temporal}, Faster R-CNN~\cite{ren2015faster} is adopted as the baseline detector, and candidate proposals filtered by the Region Proposal Network (RPN) are typically aggregated using self-attention~\cite{vaswani2017attention}. Conversely, in~\cite{zhou2022transvod,he2021end,ptseformer,fujitake2022video} Deformable DETR~\cite{zhu2020deformable} is chosen as the baseline, information interaction between  queries from different frames is facilitated through cross-attention. A common characteristic of these two types of detector is the sparsity of generated proposals—typically around 300 per frame. In contrast, one-stage detectors like YOLOX~\cite{ge2021yolox} generate a dense set of proposals. For example, the number of proposals reaches 8,400 for a $640 \times 640$ image, intensively surpassing the quantity in Faster R-CNN and Deformable DETR. Given the quadratic complexity of attention with respect to the sequence length, directly applying attention to aggregate features among vast proposals from a one-stage detector would result in substantial computational costs, negating its inference speed advantage. The characteristic of dense prediction becomes a barrier to the adaptation of one-stage detectors for VOD. In contrast to the widely used two-stage and DETR series detectors, the application of one-stage detectors in VOD has been barely explored. Driven by these considerations, a pertinent question arises: \emph{Can we selectively apply attention to only a subset of proposals from one-stage detectors to construct a practical (both accurate and fast) video object detector}?

\textbf{Contributions.} This paper answers the above question via designing a simple yet potent strategy for selecting and aggregating features produced by one-stage detectors (we use YOLOX~\cite{ge2021yolox} in this work to validate our primary claims). The major contributions of this work can be summarized as the following points: 1) As aforementioned, a substantial amount of predictions impedes efficient across-frame aggregation, we introduce a Feature Selection Module (FSM) to reject low-quality candidates, which considerably cuts down the computational expense of feature aggregation; 2) To establish the connection between features of reference frames and those of the keyframe, we introduce a Feature Aggregation Module (FAM) with a feature similarity measurement to form an affinity matrix that is utilized to guide the aggregation. To further alleviate the shortcoming of commonly-used cosine similarity, an average pooling operator on reference features is customized. These two operations cost limited computational resources with significant gains in accuracy; 3)  To demonstrate the efficacy of our design and show its superiority over other state-of-the-art alternatives, extensive experiments together with ablation studies are conducted. Equipped with the proposed strategies, our model can achieve a new record accuracy 92.9$\%$ AP50 on the ImageNet VID dataset with 30+ FPS on a single 3090 GPU (please see Fig.~\ref{fig:performance} for details) without bells and whistles, which is attractive for practical scenarios. By further post-processing, its accuracy goes up to 93.2 $\%$ AP50.

Our previous version (YOLOV) was published in~\cite{shi2023yolov}. In~\cite{shi2023yolov}, FSM adopts Non-Maximum Suppression (NMS) to reduce the redundancy, while FAM only refines the classification score. 
By contrast, this version (YOLOV++) alternatively discards NMS to maintain all the candidates above a pre-defined confidence threshold, for the sake of decreasing the risk of sub-optimal selection by NMS and activating the label assignment strategy in the baseline detector. 
In addition, YOLOV++ is enabled to simultaneously adjust both the classification and IOU scores in FAM. These two technical modifications exhibit their advantages in further boosting the accuracy with marginal overheads. Besides, this manuscript presents deeper analysis on the VOD problem with more comprehensive experiments to verify the efficacy of our design, demonstrate its superiority over other SOTA methods, and reveal the generalization ability to be applied on various detectors with different backbones.

\section{Related Work}
As a long-standing and popular topic in computer vision, object detection has been always drawing significant attention from the community with remarkable progress made over last years, particularly with the emergence of deep learning. In what follows, we will briefly review representative works in still image object detection and video object detection, which are closely related to this study.

\subsection{Still Image Object Detection}

Thanks to the development of hardware, large-scale datasets~\cite{lin2014microsoft,krizhevsky2012imagenet,shao2019objects365} and sophisticated network structures~\cite{simonyan2014very,he2016deep,xie2017aggregated,wang2020cspnet,senet,mobilenet,convnext,efficientnet,liu2021swin,dosovitskiy2020image}, the performance of object detection has continuously improved. Modern object detectors can be generally divided into two-stage and one-stage schemes. The general pipeline of two-stage detectors, pioneered by R-CNN~\cite{girshick2014rich} and its variants including Faster R-CNN~\cite{ren2015faster}, R-FCN~\cite{dai2016r}, and Mask R-CNN~\cite{he2017mask}, begins by selecting candidate regions through Region Proposal Network (RPN), followed by feature extraction using modules like RoIPooling~\cite{ren2015faster} and RoIAlign~\cite{he2017mask}. The bounding box regression and classification are then finished through an extra detection head. The methods in this group can achieve relatively accurate detection results but at high time price. 

In contrast, one-stage detectors such as the YOLO series~\cite{redmon2016you,redmon2017yolo9000,bochkovskiy2020yolov4}, SSD~\cite{liu2016ssd}, RetinaNet~\cite{lin2017focal}, and FCOS~\cite{tian2019fcos}, simplify the detection procedure via abandoning the proposal generation, which perform dense prediction on feature maps and directly give the position and class probability. These one-stage detectors are usually faster but less accurate than the mentioned two-stage ones owing to the end-to-end manner. With recent innovations in one-stage detection~\cite{wang2021scaled,ge2021yolox,Wang_2023_CVPR,xu2022pp}, the accuracy turns to be more and more competitive with the two-stage ones. 
However, the dense prediction characteristic of these models presents challenges for feature aggregation via the attention mechanism when applied to video data, thereby constraining their utility in VOD. Our research seeks to investigate the feasibility of selectively aggregating features over only a subset of proposals from one-stage detectors.

Moreover, the Transformer encoder-decoder architecture~\cite{vaswani2017attention}, as a novel technical paradigm, gives birth to sparse object detectors~\cite{zhang2022dino,carion2020end,zhu2020deformable,liu2022dab}. DETR~\cite{carion2020end} represents the initial successful application of the Transformer to object detection, utilizing object queries for direct sparse prediction without need of manually designed components. Its follow-up, say Deformable DETR~\cite{zhu2020deformable}, ameliorates DERT by introducing a deformable attention module to focus on a limited number of key sampling points, and thereby speeding up the training. Unfortunately, the computational demand of DERT-like models are relatively heavy.

\subsection{Video Object Detection}

Compared to still image object detection, degradation may frequently occur in partial video frames. When one frame is polluted, temporal information from other frames could be used for better detection. 
One branch of existing VOD methods concentrates on box-level post-processing \cite{han2016seq,belhassen2019improving,sabater2020robust,deng2019relation}. Specifically, Seq-NMS~\cite{han2016seq} leverages high-confidence detections from adjacent frames in  the same video sequence to boost the scores of less confident detections. Seq-Bbox-Matching~\cite{belhassen2019improving} matches detected bounding boxes across frames to form tubelets, and incorporates tubelet-level linking to infer missed detections and enhance detection recall. Additionally, BLR~\cite{deng2019relation} treats the post-processing task of linking bounding boxes as a pathfinding optimization problem, integrating learned relationships between objects into the post-processing phase. Similarly, REPP introduces a learning-based method for evaluating similarity between detections across frames as a preliminary step to prediction refinement. These methods initially concatenate predictions from multiple frames into object tubelets, and then adjust the confidences of object within the same tubelet using manually designed techniques.

Another branch aims to strengthen the features of keyframe, expecting to alleviate degradation via utilizing the features from (selected) reference frames. The methods derived from this idea can be roughly classified as optical flow-based~\cite{zhu2017flow,zhu2018towards}, attention-based~\cite{wu2019sequence,deng2019relation,chen2020memory,gong2021temporal,sun2021mamba,he2022queryprop,ptseformer}, and tracking-based~\cite{feichtenhofer2017detect,zhang2018integrated,liu2023objects} approaches. 
In the realm of optical flow-based methods, Deep Feature Flow~\cite{zhu2017deep} introduces optical flow for aligning image-level features, while FGFA~\cite{zhu2017flow} utilizes optical flow to aggregate features along motion paths. Additionally, MANet~\cite{wang2018fully} implements a comprehensive strategy including both pixel-level and instance-level calibration. However, the computational cost of estimating optical flow limits the efficiency of these methods. Furthermore, such approaches are generally ineffective to capture long-range temporal information.
Attention-based methods significantly alleviate the issues encountered by the optical flow-based ones. As a representative, SESLA~\cite{wu2019sequence} proposes a long-range feature aggregation scheme according to the semantic similarity between region-level features. Inspired by the relation module from~\cite{hu2018relation} for still image detection, RDN~\cite{deng2019relation} captures the relationship between objects in both spatial and temporal contexts. Furthermore, MEGA~\cite{chen2020memory} designs a memory enhanced global-local aggregation module for better modeling the relationship between objects. Alternatively, TROIA~\cite{gong2021temporal} executes the ROI alignment for fine-grained feature aggregation, while HVR-Net~\cite{han2020mining} integrates intra-video and inter-video proposal relations for further improvement. Moreover, MBMBA~\cite{sun2021mamba} enlarges the reference feature set by introducing memory bank. QueryProp~\cite{he2022queryprop} notices the high computational cost of video detectors and tries to speed up the process through a lightweight module. ClipVID~\cite{deng2023identity} employs guided attention within a reference box to selectively focus on pertinent locations in feature maps. More recently, inspired by DERT, TransVOD~\cite{zhou2022transvod} builds an end-to-end VOD framework based on a spatial-temporal Transformer architecture to effectively detect and link objects as they move and change throughout a video sequence. 
In addition to the attention-based methods, D\&T~\cite{feichtenhofer2017detect} solves VOD in a tracking manner by constructing correlation maps of features from different frames.  Tracklet-Conditioned~\cite{zhang2018integrated} combines detection results of the current frame with trajectory information computed from previous frames. Objects~\cite{liu2023objects} provides a solution to VOD by predicting object locations from a static keyframe and leveraging object motion as a supervisory signal. 
Although the above approaches boost the precision of detection, they mostly rely on heavy base detectors and suffer from the relatively slow inference speed. Besides, EOVOD~\cite{sun2022efficient} utilizes one-stage detectors as the core of its VOD framework. Despite the fast inference speed, its accuracy falls far behind the SOTA performance.

\section{Propoposed Method}

\subsection{Problem Analysis }

\begin{table*}[t]
    \centering
    \caption{Computational consumption comparison between different advanced video object detectors. Without loss of generality, the feature dimension is uniformly set to 256, YOLOX adopts the commonly-used resolution of $640\times640$ for one-stage detectors, and the inference time is tested on a single 3090 GPU. }
    \label{tab:computation_cost}
    \begin{tabular}{lccccc}
    \hline\noalign{\smallskip}
        Method                        & Base Detector                     & \# Processed Frames  & \# Candidates/frame  & Memory(GB)     & Time(ms)  \\
        \noalign{\smallskip}
        \hline
        \noalign{\smallskip}
        SELSA~\cite{wu2019sequence}   & Faster R-CNN~\cite{ren2015faster}  & 21     & 300  &  1.8    & 7.4       \\
        TransVOD~\cite{he2021end}     & Deformable DETR~\cite{zhu2020deformable} & 14     & 300  &  1.3    & 6.2       \\
        YOLOX-L + Attention                  & YOLOX~\cite{ge2021yolox}           & 4      & 8400 &  12.1   & 50.2      \\
        \hline
    \end{tabular}
    
\end{table*}

Considering the characteristics of videos (various degradation \emph{vs.} rich temporal information), instead of individually processing frames, how to seek supportive information from other frames for a target one (keyframe) plays a key role in boosting the accuracy of video detection. Several attempts~\cite{deng2019relation,chen2020memory,wu2019sequence,he2022queryprop} with noticeable improvement in accuracy corroborate the importance of temporal aggregation to the problem. 
Prior to launching our analysis and motivation, \textbf{the attention mechanism} shall be introduced for ease of exposition, which has shown the adaptability and effectiveness in feature aggregation. Self-attention was first introduced in~\cite{vaswani2017attention}, as a key part in Transformers. For an input sequence $X = \{\mathbf{x}_1, \mathbf{x}_2, ..., \mathbf{x}_n\}$, a linear transformation is first applied to obtain the Query ($Q$), Key ($K$), and Value ($V$) as follows~\cite{vaswani2017attention}: 
\begin{equation}\label{qkv}
\begin{aligned}
Q = XW_Q, 
K = XW_K, 
V = XW_V,
\end{aligned}
\end{equation}
where $W_Q$, $W_K$, and $W_V$ are the parameters to learn. Then, the normalized dot product of Query and Key is passed through a Softmax function to obtain the attention weight ($A$) through:
\begin{equation}\label{eq:att}
A = \text{Softmax}(\frac{QK^T}{\sqrt{d}}),
\end{equation}
where $d$ is the dimension of the key.
 As can be seen from Eq.~\eqref{eq:att}, the complexity is $O(n^2 d)$ with $n$ the length of the input sequence. 

(Faster) R-CNN and (Deformable) DETR are two popular architectures that well fit the feature aggregation via attention. In the R-CNN based (two-stage) framework, massive candidate regions are first ``selected" as proposals by the Region Proposal Network. Then each proposal is determined as an object or not and which class it belongs to, together with NMS for redundancy reduction.  While for the DETR series, the quantity of detection boxes corresponds to the number of queries initialized within the network decoder. The amount of candidate predictions per frame by the above detector is quite limited (\emph{e.g.}, 300 proposals in Faster R-CNN). As previously discussed, their main drawback is slow inference speed. The computational bottleneck of the R-CNN category mainly comes from dealing with substantial low-confidence region candidates.  
Despite improvements in the attention mechanism, the extended input sequence length proportionally burdens the encoder in Deformable DETR, adversely affecting its efficiency.
Alternatively, one-stage detectors yield dense predictions, with the number of prediction boxes contingent on the size of feature maps and the quantity of anchors. Taking the YOLOX~\cite{ge2021yolox} detector as an example, given an input image of $640 \times 640$, its amount of detection boxes is $8,400$. Directly dealing with feature maps of a one-stage detector through the attention for temporal information aggregation inevitably results in tremendous memory and computation consumption. To be more clear, Table~\ref{tab:computation_cost} reports the computational resource demands of video object detectors with various architectures, in each of which, only one layer of self-attention for is used for feature aggregation. In contrast to the Faster R-CNN and Deformable DETR (sparse) detectors, YOLOX (dense) armed with the self-attention incurs a great deal of memory usage and long inference time. Moreover, the employment of self-attention on YOLOX for aggregating features from merely $4$ frames escalates the inference time beyond $50$ms and necessitates upwards of 12 GB of GPU memory.
For comparison, the base detector YOLOX-L spends 27.2ms to sequentially process $4$ images, which is approximately half the time required in the feature aggregation phase. This is to say, directly aggregating dense candidates of multiple frames from one-stage detectors through self-attention undermines their speed advantage. Also, the improvement in accuracy will be marginal, because the number of frames to aggregate is restricted considering the computational cost.

Advanced label assignment strategies in one-stage detection~\cite{ge2021ota,zhang2020bridging,kim2020probabilistic} advocate the so-called center prior, which renders only the central part of the object in the feature map as a positive sample, while the majority of remaining features are considered as background. Again, these background features are filtered out in the first stage of two-stage detectors like Faster R-CNN and do not participate in the subsequent feature aggregation process in VOD. In light of the above analysis and observation, we propose a simple yet effective Feature Selection Module to select important regions after the one-stage detection, thereby circumventing the processing of a substantial number of low-quality candidates. This manner significantly reduces the number of features requiring aggregation in a single frame, and thus expedites the training and inference process. Furthermore, we incorporate a confidence prior into the temporal feature aggregation to construct an affinity matrix, supplanting the original cosine similarity matrix. The simultaneous consideration of the semantic similarity and the quality of reference features enhances the performance of VOD. In the next subsection, we will detail our design. 

\begin{figure*}[t]
\centering
\includegraphics[width=1.0\linewidth]{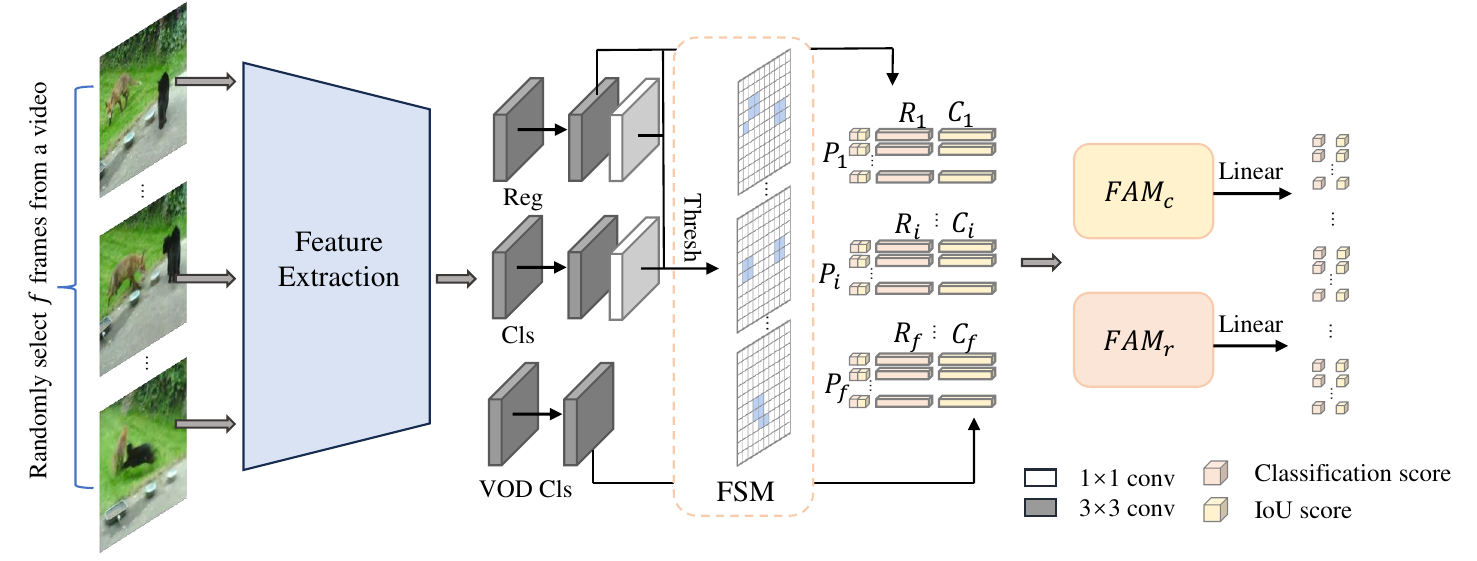}
\caption{A schematic illustration of our framework.}
\label{fig:yolov}
\end{figure*}

\subsection{Our Design}

Figure~\ref{fig:yolov} exhibits the overall framework of our method. Schematically, we randomly sample $f$ frames from the same video and feed them into the base detector to extract features. According to the prediction of YOLOX Head, our Feature Selection Module selects dense foreground proposals for further refinement. All the features selected from FSM are fed into our Feature Aggregation Module to generate classification and IoU scores. In summary, the first step is prediction (having a large number of regions with low confidences discarded), while the second step can be viewed as feature refinement (taking advantage of other frames by aggregation). By this principle, our design can simultaneously benefit from the efficiency of one-stage detectors and the accuracy gained from temporal aggregation. We choose the YOLOX~\cite{ge2021yolox} as base to present our main claims throughout the paper. The generalization ability of proposed strategy to other base detectors such as FCOS~\cite{tian2019fcos} and PPYOLOE~\cite{xu2022pp} is also demonstrated in the experiment section.

\subsubsection{FSM: Feature Selection Module}

The aggregation of background information has been intuitively considered to have a minimal contribution towards enhancing detection performance in images of compromised quality. In the realm of sparse detectors, exemplified by video object detectors utilizing Faster R-CNN as the foundational detector, a significant portion of the background features is eliminated by the RPN, leaving primarily those features deemed as foreground for temporal aggregation. Conversely, in one-stage detectors, foreground predictions are initially picked out through a predefined confidence threshold, followed by the NMS to reduce redundant predictions, finally producing a set of sparse foreground predictions.

\begin{table}[t]
\centering
\caption{Average proposal number per frame (N) and class-agnostic recall (AR) at IoU=0.5 for YOLOX-S and YOLOX-SwinTiny models using two different feature selection manners (TopK + NMS \emph{vs.}  Thresh) on the validation set of ImageNet VID. The confidence threshold is set to 0.001 in the Thresh pipeline.}
\label{tab:AR_calculation}
\begin{tabular}{l|cccc}
\hline
Model & Pipeline & $N$ & AR (\%) \\
\hline
YOLOX-S & TopK + NMS & 30 & 95.8 \\
\hline
YOLOX-S & Thresh & 83.0 & 95.3 \\
\hline
YOLOX-SwinTiny & TopK + NMS & 30 & 97.2 \\
\hline
YOLOX-SwinTiny & Thresh & 33.8 & 95.6 \\
\hline
\end{tabular}

\end{table}

To refine the output of one-stage detectors, we first select the top $K$ predictions (\emph{e.g.}, $K=750$) based on confidence scores. Then, a predefined number of predictions ($N=30$, for instance) are chosen by the NMS. The features corresponding to these selected predictions are gathered for further refinement. This feature selection strategy, referred to as TopK+NMS, aims to secure sparse foreground predictions. A pertinent question arises regarding the efficacy of this strategy in maintaining a sufficiently high foreground recall rate, which is crucial for maximizing the potential performance. 
To address this concern, we evaluate the class-agnostic recall at an IoU threshold of 0.5 for sparse foreground predictions using the validation set of the ImageNet VID dataset. This evaluation is conducted on models trained with YOLOX-S and YOLOX-SwinTiny, as documented in Tab.~\ref{tab:AR_calculation}. The findings reveal that both the two models achieve 95\%+ recalls, demonstrating that TopK+NMS can effectively encompass the majority of the foreground areas. 

TopK+NMS successfully filters out sparse foreground features, but due to the label assignment strategy (\emph{e.g.}, OTA~\cite{ge2021ota}) in one-stage detectors being specially tailored for dense predictions, using the previous label assignment scheme for sparse predictions after temporal information aggregation leads to poor performance, as indicated in Tab.~\ref{table:effectiveness of ours}. Specifically, when using OTA to assign labels to the features filtered by TopK+NMS, the accuracy of supervising the classification score and IoU score of the aggregated features is not as good as the accuracy of only adjusting the classification score. 
To address this concern, a feasible approach is only retaining dense foreground predictions. To achieve this goal, we divide the feature map into foreground and background parts through a simple confidence threshold, which is simply dubbed as Thresh in Tab.~\ref{tab:AR_calculation}. Besides, we also report the class-agnostic recall and the average proposals per frame in Tab.~\ref{tab:AR_calculation}. The recall rate slightly decreases when using the threshold to extract the foreground, but it is still above 95$\%$. The value of $N$ varies with the backbones adopted, but it is generally within 100, which is even smaller than the sparse detectors. We show that the foreground features obtained in this way can fit well with the label assignment strategy of dense detection, bringing additional performance improvement.

\begin{figure}
    \centering
    \includegraphics[width=1\linewidth]{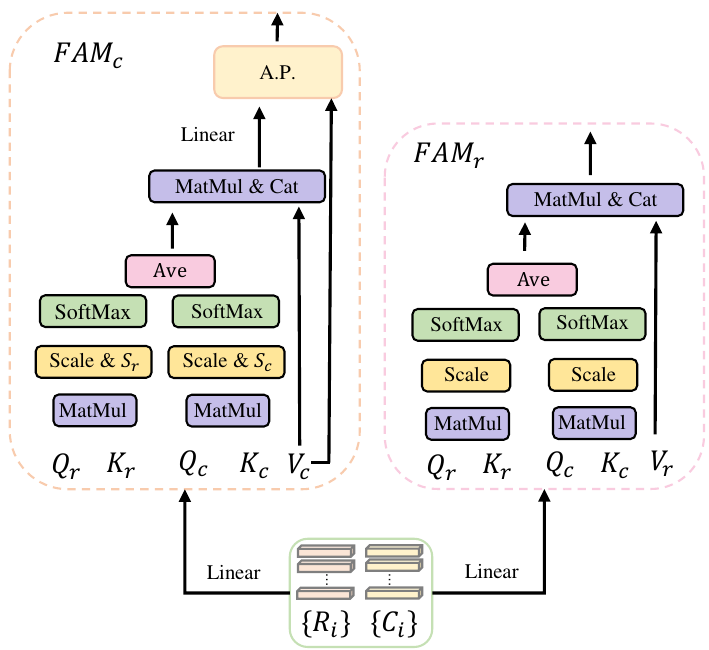}
    \caption{Feature aggregation process for classification and regression features. $S_r$ and $S_c$ denote the score matries of IoU and classification, respectively.}
    \label{fig:FAM}
\end{figure}

In practice, we find that directly aggregating the collected features in the classification branch and backpropagating the classification loss of the aggregated features results in unstable training. Since the weights of the feature aggregation module (detailed in the next subsection) are randomly initialized, fine-tuning all the networks from the beginning may contaminate the pre-trained weights. To address these issues, we fix the weights in the base detector except for the linear projection layers in the detection head. We further insert two $3 \times 3$ convolutional (Conv) layers into the model neck as a new branch, called the video object classification branch, which generates features for aggregation. Then, we feed the collected features from the video and regression branches into our feature aggregation module. This staged training strategy allows for efficient training of the feature aggregation module. Even when using a strong backbone (\emph{e.g.}, Swin Transformer based version), the feature aggregation module proposed in this paper can be trained on a single 3090 GPU in just 12 hours, and brings competitive or better accuracy compared to the end-to-end training manner.

\subsubsection{FAM: Feature Aggregation Module}

\begin{figure*}[ht]
\centering
\includegraphics[width=1\linewidth]{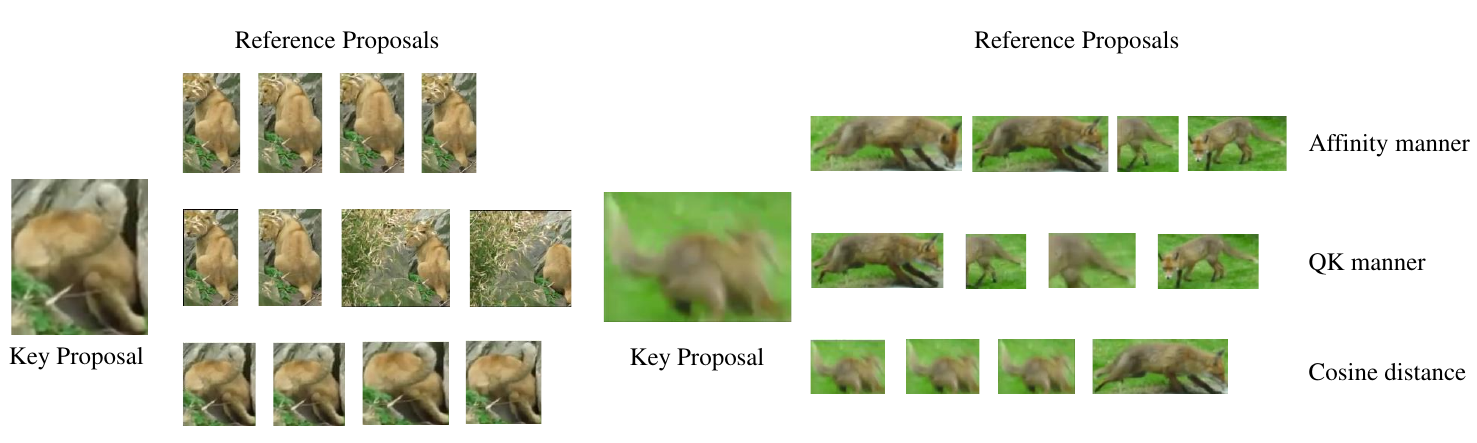}
\caption{Visual comparison between reference proposals selected by three different methods for given key proposals. We display four reference proposals that contribute most in aggregation.}
\vspace{1mm}
\label{fig:conf compare}
\end{figure*}
\begin{figure}[ht]
\centering
\includegraphics[width=1\linewidth]{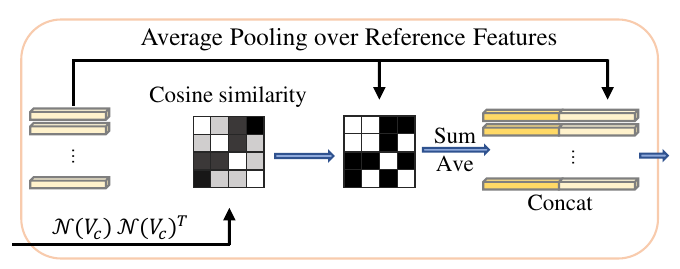}
\caption{Average Pooling over Reference Frame Features}
\label{fig:ave_pooling}
\end{figure}

After obtaining candidate features from FSM, we now proceed to the step of temporal information aggregation. The entire workflow of our feature aggregation module is illustrated in Fig.~\ref{fig:FAM}, which comprises two components: a classification feature aggregation module (FAM$c$) and a regression feature aggregation module (FAM$_r$). Let us start with using multi-head attention to build our feature aggregation module. Given that the base detector's detection head decouples object features into classification and regression parts, we take into account the features of both branches when performing feature aggregation. Let $\mathcal{F} = \left\{{C}_{1}, {C}_{2}, ..., {C}_{f};{R}_{1}, {R}_{2}, ..., {R}_{f}\right\}$ denote the feature set selected by FSM where $C_i\in\mathbb{R}^{d\times a_i} = \left[\mathbf{c}_{i}^{1}, \mathbf{c}_{i}^{2}, ..., \mathbf{c}_{i}^{a_i}\right]$  and  $R_i\in\mathbb{R}^{d\times a_i} = \left[\mathbf{r}_{i}^{1}, \mathbf{r}_{i}^{2}, ..., \mathbf{r}_{i}^{a_i}\right]$  denote the features of the $i$-th frame in $\mathcal{F}$ from the video classification and regression branches, respectively. $d$ and $f$ refer to the feature dimension and the number of related frames, respectively. $a_i$ is a constant value when using the TopK+NMS feature selection pipeline and dynamically changes when adopting the Thresh feature selection pipeline. Similarly to~\cite{vaswani2017attention}, the Query, Key, and Value matrices are constructed and fed into the multi-head attention. For instance, $Q_c$ and $Q_r$ are respectively formed by stacking the features from the classification branch and the regression branch for all proposals in all related frames (\emph{i.e.}, $Q_c\in\mathbb{R}^{n\times d}=\operatorname{LP}([{C}_{1}, {C}_{2}, ..., {C}_{f}]^T)$ and $Q_r\in\mathbb{R}^{n\times d}=\operatorname{LP}([{R}_{1}, {R}_{2}, ..., {R}_{f}]^T)$, where $\operatorname{LP}(\cdot)$ is the linear projection operator and $n$ is the total proposals in $f$ frames, while the others are done similarly. By the scaled dot-product, we obtain the corresponding attention weights through the following:
\begin{equation}
\begin{array}{c}
    \operatorname{A_c}=\operatorname{softmax}\left(\frac{Q_c K_c^{T}}{\sqrt{d}}\right),
    \operatorname{A_r}=\operatorname{softmax}\left(\frac{Q_r K_r^{T}}{\sqrt{d}}\right).
\end{array}
\label{eq:attention_weights}
\end{equation}
The features after aggregation can be obtained via:
\begin{equation}
\begin{array}{c}
\operatorname{SA(\mathcal{F})}= \operatorname{concat}((\operatorname{A_c}+\operatorname{A_r})V/2,V),
\end{array}
\end{equation}
when aggregating features from the classification branch, $V$ refers to $V_c$, and similarly for the regression branch features. In the feature aggregation process, we concatenate $V$ with $(\operatorname{A_c}+\operatorname{A_r})V/2$ to better preserve the initial representations. Following the decoupled design in YOLOX, we use two sets of weights to aggregate the classification and regression features respectively. The aggregated features are then passed through a linear projection to obtain the prediction score.

\textbf{The homogeneity issue.} The generalized cosine similarity is arguably the most widely used metric to compute the similarity between features or the attention weight~\cite{wu2019sequence,shvets2019leveraging,deng2019relation}. Simply referring to the cosine similarity will find features most similar to the target. When the target suffers from some degradation, the selected reference proposals using cosine similarity are very likely to have the same problem. We name this phenomenon \emph{the homogeneity issue}. To overcome the homogeneity issue, not only the similarity between features needs to be considered during the aggregation process of classification features, but also the quality of the reference features themselves, which could be described by the classification score and IoU score of the proposals from the base detector. Thus, we further incorporate predicted confidences into the computation of attention weights. Specifically, the similarity score between the query of the $p$-th proposal and the key of the $q$-th proposal in the feature aggregation module, $(QK^{T})_{pq}/\sqrt{d}$, is updated to $(QK^{T})_{pq}\times S_q/\sqrt{d}$, where $S_q$ is the confidence of the $q$-th proposal. When calculating the similarity of classification features, this confidence is the classification score of the proposal, and when calculating the similarity of regression features, this confidence is the IoU score of the proposal. We denote the updated feature similarity calculation method as the affinity manner. We provide a visual comparison in Fig.~\ref{fig:conf compare} to show the superiority of our affinity manner. They are the lion case with a rare pose and the fox case with motion blur. Without loss of generality, the top $4$ reference proposals are listed for different feature selection modes, including the cosine similarity, QK manner in multi-head attention, and our affinity manner. As previously analyzed, the cosine manner selects proposals most similar to the key proposal but suffering the same degradation problem as the key proposal. Though the QK manner alleviates this problem, it is obviously inferior to the affinity manner. By introducing the confidence scores as guidance, our method selects better proposals and further boosts the detection accuracy with quite limited computational cost. 

Furthermore, given the nature of softmax, a minor portion of references may possess a significant share of weights. This often results in neglecting references with low weights, thereby limiting the diversity of reference features for potential subsequent use. To mitigate such risks, we propose an \textbf{average pooling over reference features (A.P.)} approach, which is depicted in Fig.~\ref{fig:ave_pooling}. Specifically, we select all references with similarity scores exceeding a threshold $\tau$ and apply average pooling to these selected features. It is worth noticing that the similarity in this study is calculated via $\mathcal{N}(V_c) \mathcal{N}(V_c)^{T}$. The operator $\mathcal{N}(\cdot)$ represents layer normalization, ensuring the values fall within a specific range and thus eliminating the impact of scale difference. The average-pooled features and the key features are subsequently input into a linear projection layer for final classification. 

\section{Experimental Validation}

\subsection{Implementation Details}
We employ YOLOX~\cite{ge2021yolox}, PPYOLOE~\cite{xu2022pp} and FCOS~\cite{tian2019fcos} as base detectors to verify the effectiveness of our design. Please notice that, without loss of generality, ablation studies are carried out on YOLOX-S. The ImageNet VID~\cite{russakovsky2015imagenet}, is used to conduct experiments, which contains 3,862 videos for training, and 555 videos for validation. There are 30 categories in total, \emph{i.e.}, a subset of the 200 basic-level categories of the object detection task. The AP50 and inference speed are two metrics to reflect the performance in accuracy and efficiency, respectively.

Similarly to previous works~\cite{wu2019sequence,shi2023yolov}, we also initialize our base detector with COCO pre-trained weights. Only the linear projection layers in YOLOX prediction head, the newly added video object classification branch and the FAM are fine-tuned for the sake of largely excluding the influence from other factors.
Following the settings in ~\cite{chen2020memory,gong2021temporal,zhu2017flow}, we combine the ImageNet VID and the same classes in ImageNet DET~\cite{russakovsky2015imagenet} as our training data to train base detectors.  Considering the redundancy of video frames, we randomly sample 1/10 frames in the VID training set instead of using all of them. The base detectors are trained for 7 epochs by SGD with batch size of 32. As for the learning rate, we adopt the cosine learning rate schedule used in YOLOX with 1 warm-up epoch. When integrating the FAM into the base detectors, we fine-tune them for 150K iterations with batch size of 16 on a single 3090 GPU. Furthermore, our training rule incorporates an initial warm-up phase for the first 15k iterations, followed by a cosine learning rate schedule for subsequent iterations. The foundational learning rate is established at $5 \times 10^{-4}$ for all the models, except for those utilizing Swin-Base~\cite{liu2021swin} and FocalNet-Large~\cite{yang2022focal} backbones, for which the learning rate is adjusted to $1.25 \times 10^{-4}$. When training the FAM, we set the number of frames $f$ to 16. The thresholds for IoU in NMS and confidence are configured to $0.75$ and $0.001$, respectively, facilitating preliminary feature selection via the TopK+NMS and Thresh feature selection methods. Conversely, for the generation of final detection boxes, the NMS threshold is modified to $0.5$ to encompass a broader range of high-confidence candidates.
During the training phase, the image size is randomly altered, ranging from $352 \times 352$ to $672 \times 672$, with an increment of 32 strides. For evaluation, images are uniformly resized to $576 \times 576$. Besides, parallel predictions are conducted following recent works~\cite{zhou2022transvod,deng2023identity} for fair comparison. All the models are evaluated using FP16 precision on a 3090 GPU. 
\setlength{\tabcolsep}{4pt}
\begin{table*}[t]
\begin{center}
\caption{Comparison of accuracy and efficiency on the ImageNet VID dataset. The reported inference speeds of different models are measured on the same device with a 3090 GPU. The missed results (marked `-') in the Without Post-processing group indicate the codes/models of corresponding methods are unavailable.}

\label{table:compare w post}
\begin{tabular}[t]{l|cc|cc|c}
\hline\noalign{\smallskip}
Methods & Backbone & Base detector & AP50 ($\%$) & Time (ms) & Post-processing \\
\hline
\multicolumn{6}{c}{Without Post-processing} \\
\hline
\noalign{\smallskip}
FGFA\textcolor{gray}{\textsubscript{[ICCV17]}}~\cite{zhu2017flow} & ResNet-101 & R-FCN & 76.3 & 62.0 & -\\ 
SELSA\textcolor{gray}{\textsubscript{[ICCV19]}}~\cite{wu2019sequence} & RestNet-101 & Faster R-CNN & 80.3 & 80.0 & -\\
RDN\textcolor{gray}{\textsubscript{[ICCV19]}}~\cite{deng2019relation} & RestNet-101 & Faster R-CNN & 81.8 & 93.1 & -\\
MEGA\textcolor{gray}{\textsubscript{[CVPR20]}}~\cite{chen2020memory} & ResNet-101 & Faster R-CNN & 82.9 & 121.8 & -\\
HVR\textcolor{gray}{\textsubscript{[ECCV20]}}~\cite{han2020mining} & ResNet-101 & Faster R-CNN & 83.2 & - & -\\
MAMBA\textcolor{gray}{\textsubscript{[AAAI21]}}~\cite{sun2021mamba} & ResNet-101 & Faster R-CNN & 84.6 & - & -\\ 
TROIA\textcolor{gray}{\textsubscript{[AAAI21]}}~\cite{gong2021temporal} & ResNet-101 & Faster R-CNN & 82.0 & 133.5 & -\\
TransVOD\textcolor{gray}{\textsubscript{[ACM MM22]}}~\cite{he2021end} & ResNet-101 & Deformable DETR & 81.9 & 345.0 & -\\  
QueryProp\textcolor{gray}{\textsubscript{[AAAI22]}}~\cite{he2022queryprop} & ResNet-101 & Sparse R-CNN & 82.3 & - & -\\
PTSEFormer\textcolor{gray}{\textsubscript{[ECCV22]}}~\cite{ptseformer} & ResNet-101 & Deformable DETR & 88.1 & - & -\\
ClipVID\textcolor{gray}{\textsubscript{[ICCV23]}}~\cite{deng2023identity} & ResNet-101 & Faster R-CNN & 84.7 & - & -\\
Objects\textcolor{gray}{\textsubscript{[ICCV23]}}~\cite{liu2023objects} & ResNet-101 & Deformable DETR & 87.9 & - & -\\
\hline
\noalign{\smallskip}
SELSA\textcolor{gray}{\textsubscript{[ICCV19]}}~\cite{wu2019sequence} & RestNeXt-101 & Faster R-CNN & 83.1 & 76.6 & -\\ 
RDN\textcolor{gray}{\textsubscript{[ICCV19]}}~\cite{deng2019relation} & RestNeXt-101 & Faster R-CNN & 83.2 & - & -\\
MEGA\textcolor{gray}{\textsubscript{[CVPR20]}}~\cite{chen2020memory} & ResNeXt-101 & Faster R-CNN & 84.1 & - & -\\
HVR\textcolor{gray}{\textsubscript{[ECCV20]}}~\cite{han2020mining} & ResNeXt-101 & Faster R-CNN & 84.8 & - & -\\
TROIA\textcolor{gray}{\textsubscript{[AAAI21]}}~\cite{gong2021temporal} & ResNeXt-101 & Faster R-CNN & 84.3 & 143.2 & -\\
ClipVID\textcolor{gray}{\textsubscript{[ICCV23]}}~\cite{deng2023identity} & ResNeXt-101 & Faster R-CNN & 85.8 & - & -\\
\hline
\noalign{\smallskip}
EVOD\textcolor{gray}{\textsubscript{[ECCV22]}}~\cite{sun2022efficient} &  Modified CSP v5 & YOLOX-S & 75.1 & - & -\\
YOLOV\textcolor{gray}{\textsubscript{[AAAI23]}}~\cite{shi2023yolov} & Modified CSP v5 & YOLOX-S & 77.3 & 4.0 & - \\
YOLOV++ & Modified CSP v5 & YOLOX-S & 78.7 & 5.3 & - \\

YOLOV\textcolor{gray}{\textsubscript{[AAAI23]}}~\cite{shi2023yolov} & Modified CSP v5 & YOLOX-L & 83.6 & 6.3 & - \\
YOLOV++ & Modified CSP v5 & YOLOX-L & 84.2 & 7.6 & -\\

\hline
\noalign{\smallskip}
TransVOD Lite\textcolor{gray}{\textsubscript{[TPAMI23]}}~\cite{zhou2022transvod} & SwinTiny & Deformable DETR & 83.7 & 33.8 & -\\
YOLOV++ & SwinTiny & YOLOX-SwinT & 85.7 & 8.4 & -\\ 
TransVOD Lite\textcolor{gray}{\textsubscript{[TPAMI23]}}~\cite{zhou2022transvod} & SwinBase & Deformable DETR & 90.1 & 51.7 & -\\
Objects\textcolor{gray}{\textsubscript{[ICCV23]}}~\cite{liu2023objects} & SwinBase & Deformable DETR & 91.3 & - & -\\
YOLOV++ & SwinBase & YOLOX-SwinB & 90.7 & 15.9 & -\\
YOLOV++ & FocalLarge & YOLOX-FocalL & \textbf{92.9} & 27.6 & -\\

\hline
\multicolumn{6}{c}{With Post-processing} \\
\hline
\noalign{\smallskip}
FGFA\textcolor{gray}{\textsubscript{[ICCV17]}}~\cite{zhu2017flow} & ResNet-101 & R-FCN & 78.4 & - & Seq-NMS~\cite{han2016seq}\\
SELSA\textcolor{gray}{\textsubscript{[ICCV19]}}~\cite{sabater2020robust} & RestNet-101 & Faster R-CNN & 84.2 & - & REPP~\cite{sabater2020robust}\\
RDN\textcolor{gray}{\textsubscript{[ICCV19]}}~\cite{deng2019relation} & RestNet-101 & Faster R-CNN & 83.8 & - & BLR~\cite{deng2019relation}\\
RDN\textcolor{gray}{\textsubscript{[ICCV19]}}~\cite{deng2019relation} & RestNeXt-101 & Faster R-CNN & 84.7 & - & BLR~\cite{deng2019relation}\\
MEGA\textcolor{gray}{\textsubscript{[CVPR2020]}}~\cite{chen2020memory} & ResNet-101 & Faster R-CNN & 84.5 & - & BLR~\cite{deng2019relation}\\
MEGA\textcolor{gray}{\textsubscript{[CVPR2020]}}~\cite{chen2020memory} & ResNeXt-101 & Faster R-CNN & 85.4 & - & BLR~\cite{deng2019relation}\\
HVR\textcolor{gray}{\textsubscript{[ECCV2020]}}~\cite{han2020mining} & ResNet-101 & Faster R-CNN & 83.8 & - & Seq-NMS~\cite{han2016seq}\\
HVR\textcolor{gray}{\textsubscript{[ECCV2020]}}~\cite{han2020mining} & ResNeXt-101 & Faster R-CNN & 85.5 & - & Seq-NMS~\cite{han2016seq}\\
\hline
\noalign{\smallskip}
YOLOV++ & Modified CSP v5 & YOLOX-L & 85.6 & - &  REPP~\cite{sabater2020robust}\\
YOLOV++ & Modified CSP v5 & YOLOX-SwinT & 86.6 & -&  REPP~\cite{sabater2020robust}\\
YOLOV++ & Modified CSP v5 & YOLOX-SwinB & 91.3 & -&  REPP~\cite{sabater2020robust}\\
YOLOV++ & Modified CSP v5 & YOLOX-FocalL & \textbf{93.2} & -&  REPP~\cite{sabater2020robust}\\

\hline
\end{tabular}
\end{center}
\end{table*}

\subsection{Comparison with State-of-the-art Methods}
Table~\ref{table:compare w post} provides a comprehensive comparison of our proposed models against existing competitors in terms of accuracy and inference speed. In the top segment of Tab.~\ref{table:compare w post}, we outline the performance metrics of various competing models without applying any post-processing techniques. Owing to the inherent advantages of the one-stage detection framework and the efficacy of our feature aggregation strategy, our model exhibits a commendable balance of detection accuracy and inference efficiency. 
Notably, when employing YOLOX-S as the base detector, our YOLOV++ achieves an impressive 78.7\% AP50 with an inference time of only 5.3ms. Compared to our previous method, YOLOV, this advanced version further improves AP50 by 1.4\% with marginal additional time cost. When utilizing SwinTiny as the backbone, our model consistently outperforms TransVOD Lite in both accuracy and efficiency, yielding a 2.0\% AP50 improvement and nearly $4\times$ faster. The efficiency of our method allows for the use of more powerful backbones while maintaining a lead in inference speed. For instance, when FocalNet Large is used as the backbone, our model achieves a remarkable 92.9\% AP50 that significantly outperforms other methods, while only spending 27.6ms to process one frame.
The bottom portion of Tab.~\ref{table:compare w post} displays the results further adopted post-processing techniques. Due to the variety of post-processing strategies utilized, we only report the AP50 and omit the inference time. The REPP~\cite{sabater2020robust} further increases the AP50 of our models to 93.2\%, surpassing the previous SOTA method by 1.9\%. This improvement highlights the efficacy of our approach in establishing new benchmarks for accuracy and efficiency in VOD.

\begin{figure*}[hp]
    \centering
    \subfloat{\includegraphics[width=0.24\linewidth,height=0.08\paperheight]{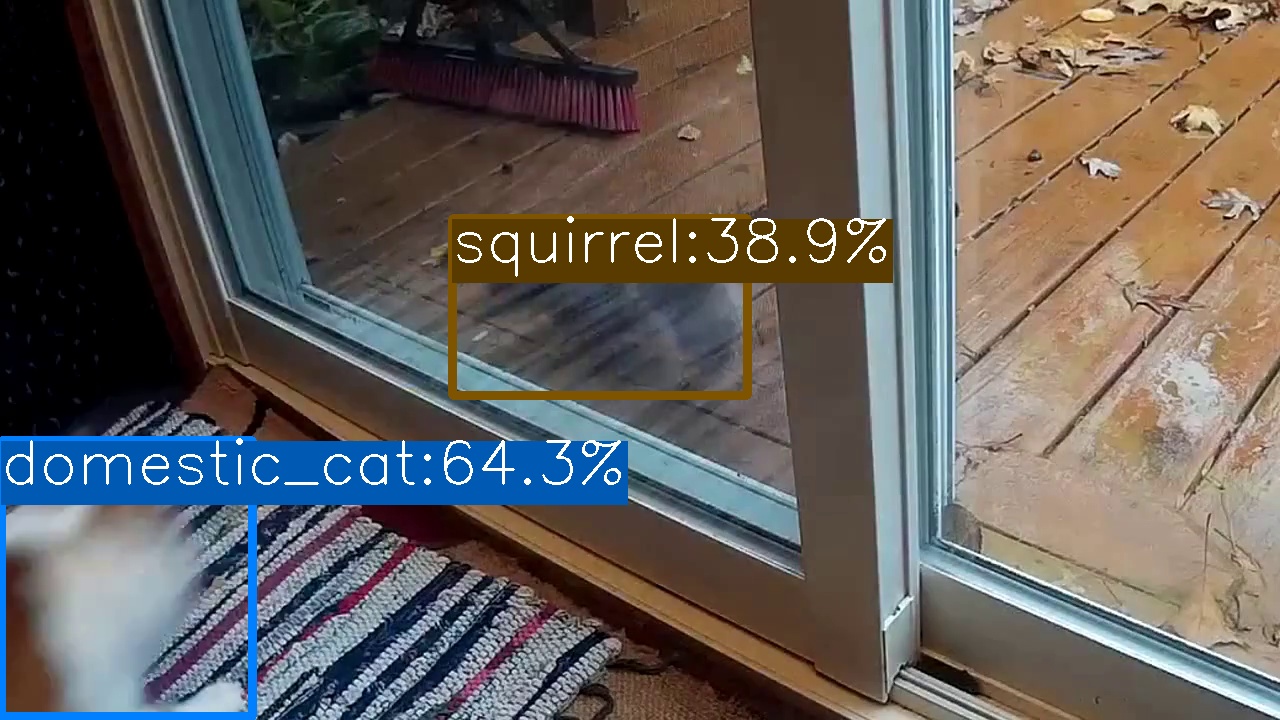}}\hfill
    \subfloat{\includegraphics[width=0.24\linewidth,height=0.08\paperheight]{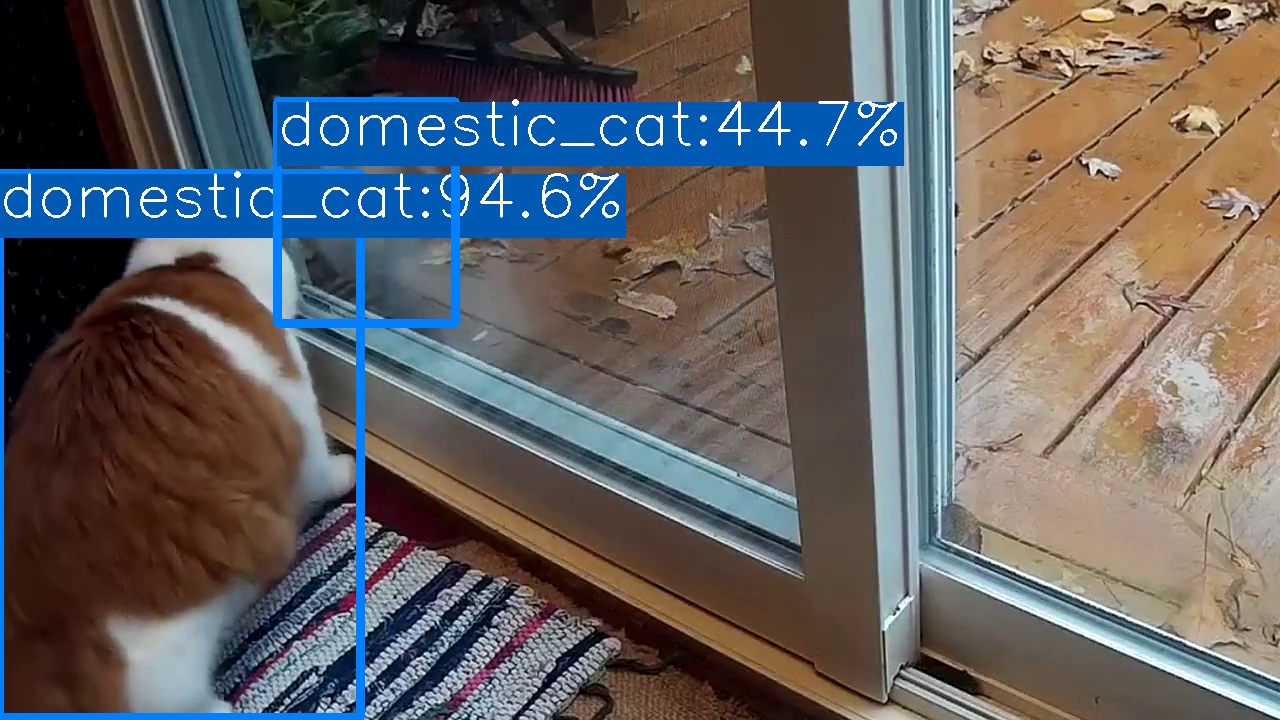}}\hfill
    \subfloat{\includegraphics[width=0.24\linewidth,height=0.08\paperheight]{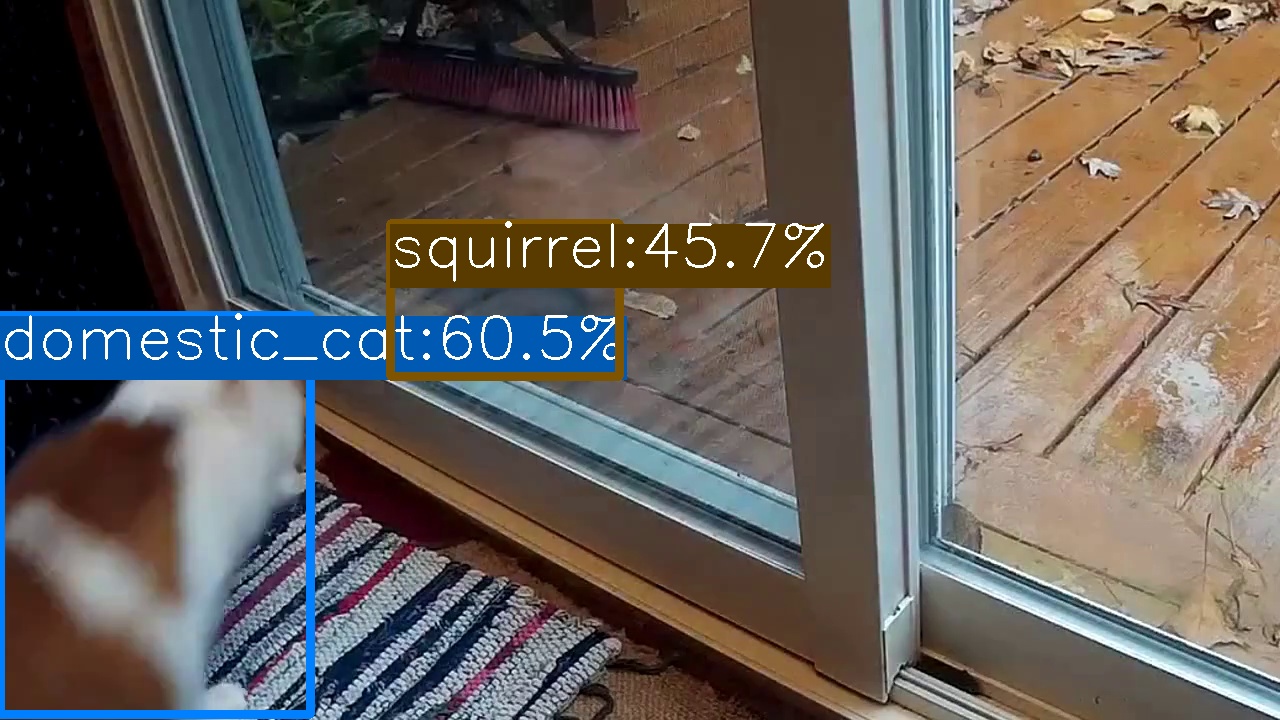}}\hfill
    \subfloat{\includegraphics[width=0.24\linewidth,height=0.08\paperheight]{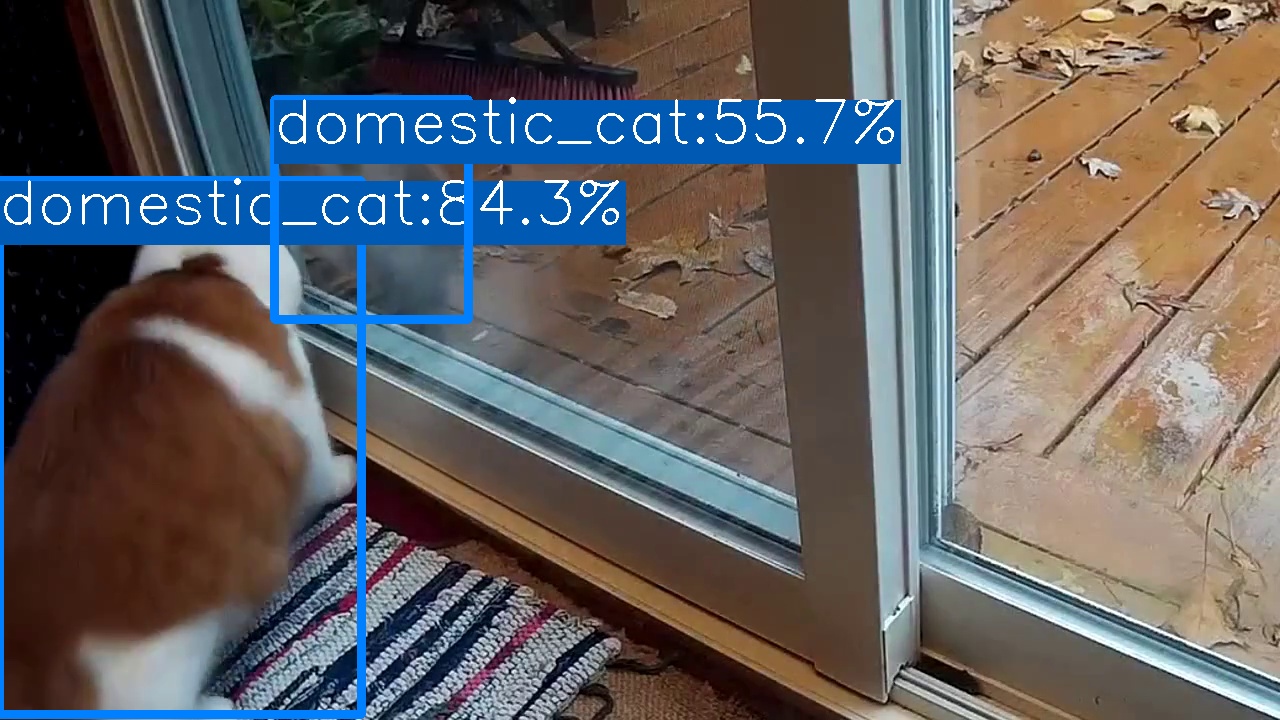}}\\
    \subfloat{\includegraphics[width=0.24\linewidth,height=0.08\paperheight]{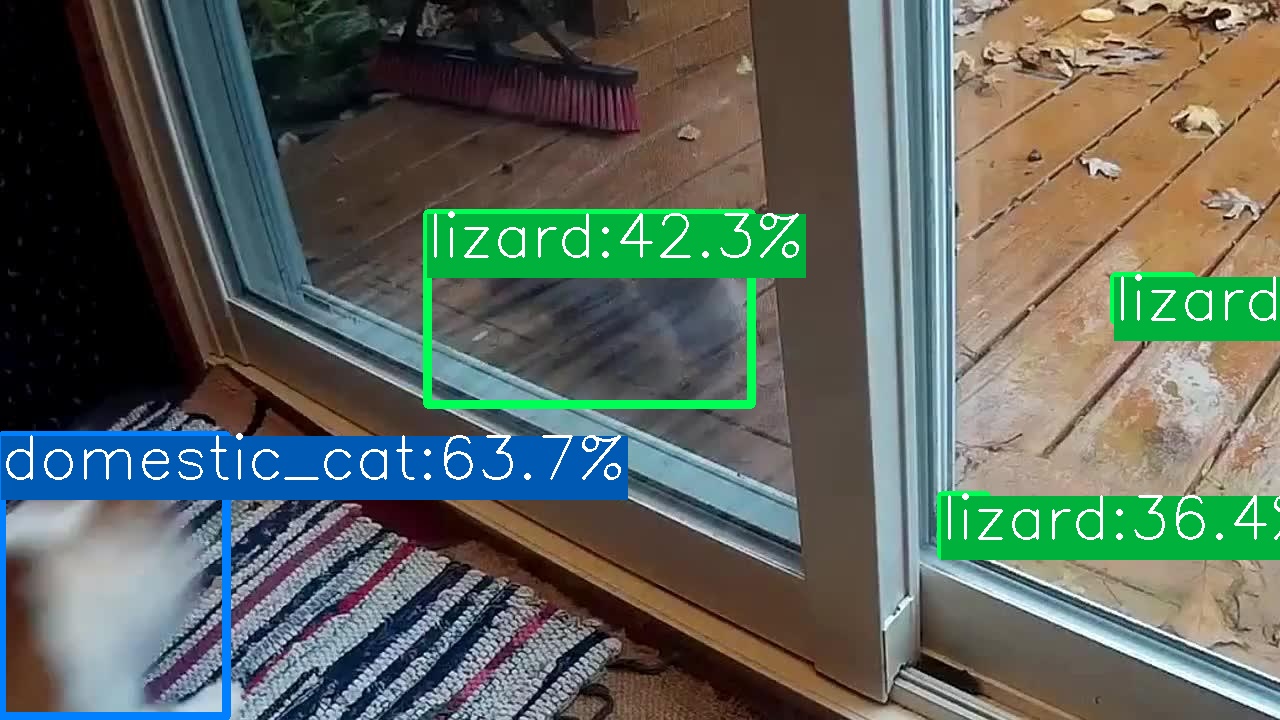}}\hfill
    \subfloat{\includegraphics[width=0.24\linewidth,height=0.08\paperheight]{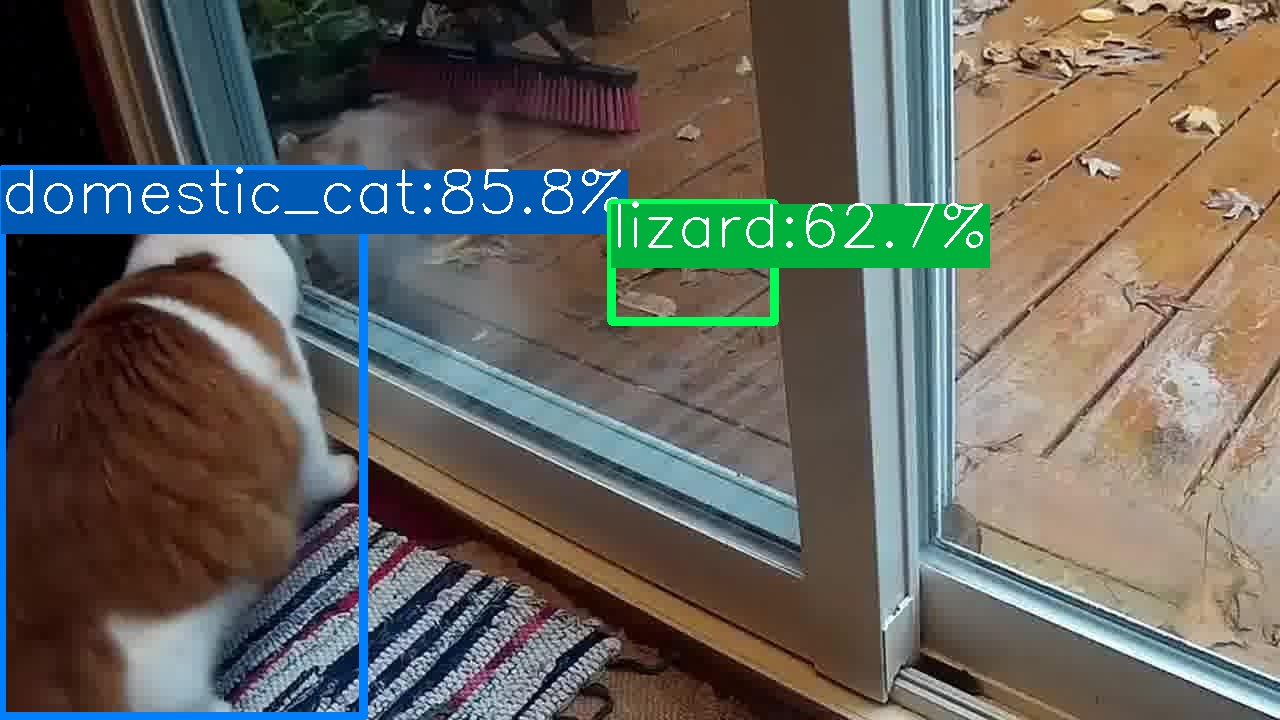}}\hfill
    \subfloat{\includegraphics[width=0.24\linewidth,height=0.08\paperheight]{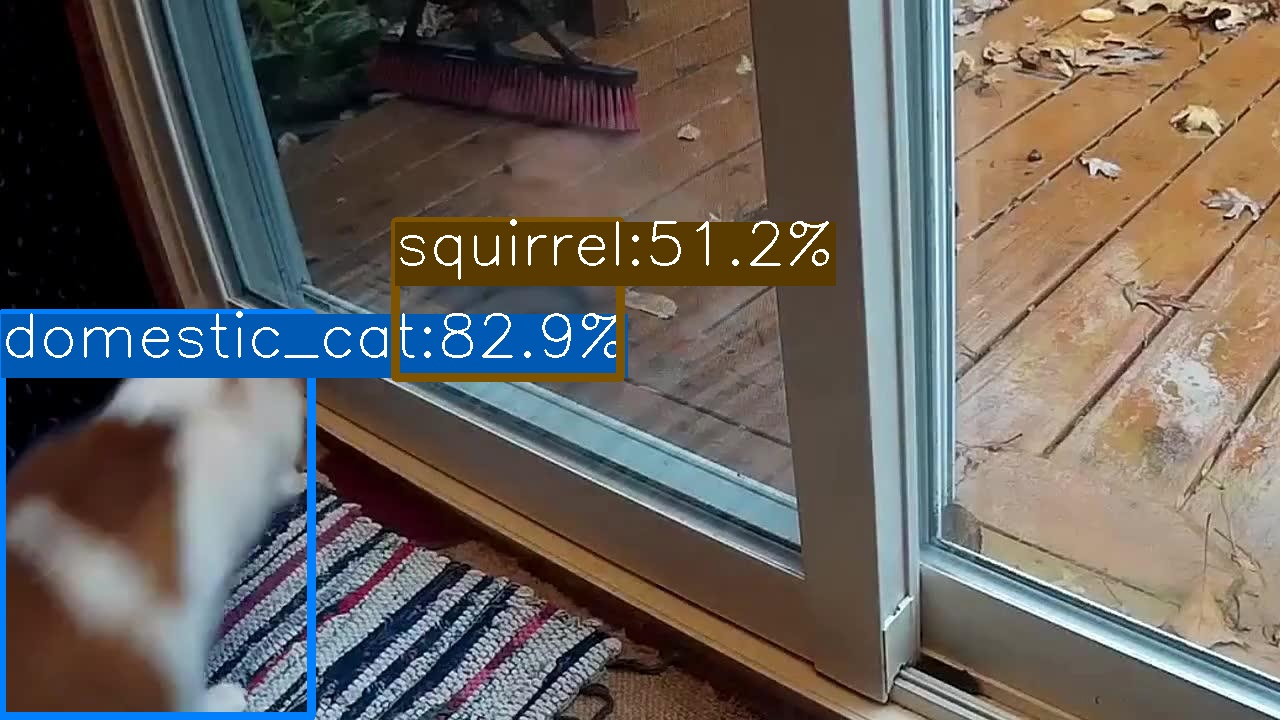}}\hfill
    \subfloat{\includegraphics[width=0.24\linewidth,height=0.08\paperheight]{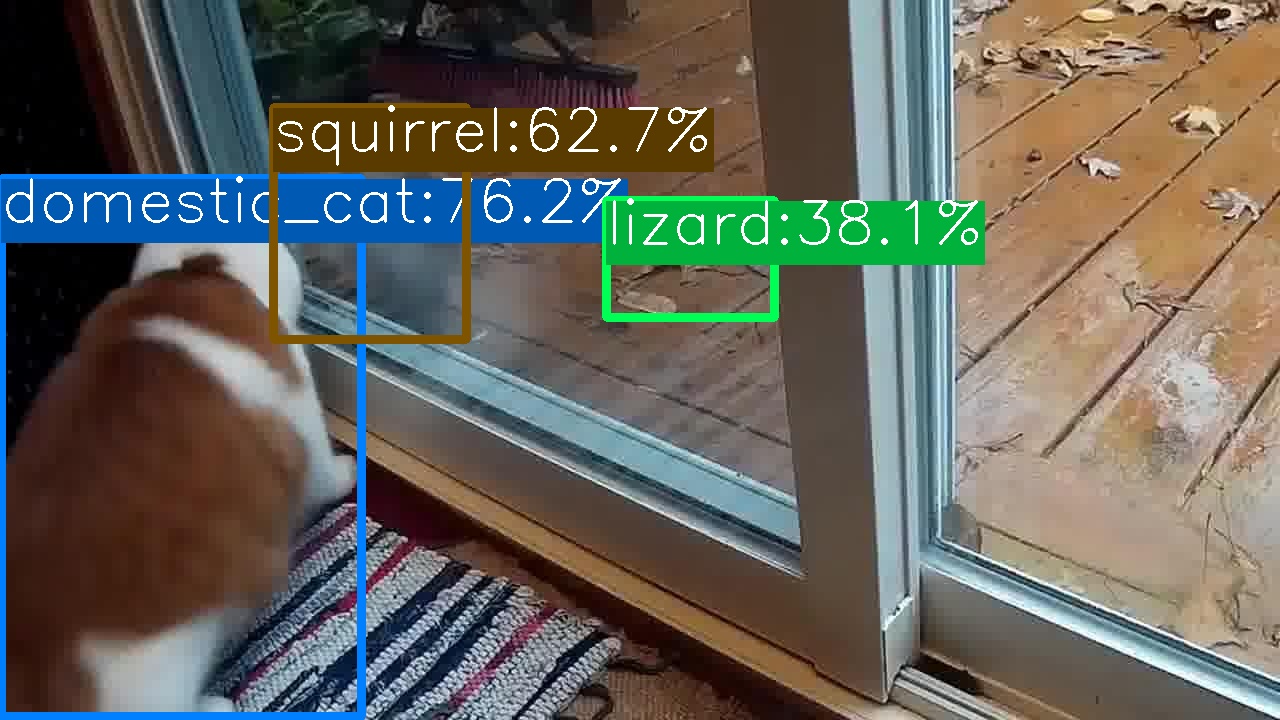}}\\
    \subfloat{\includegraphics[width=0.24\linewidth,height=0.08\paperheight]{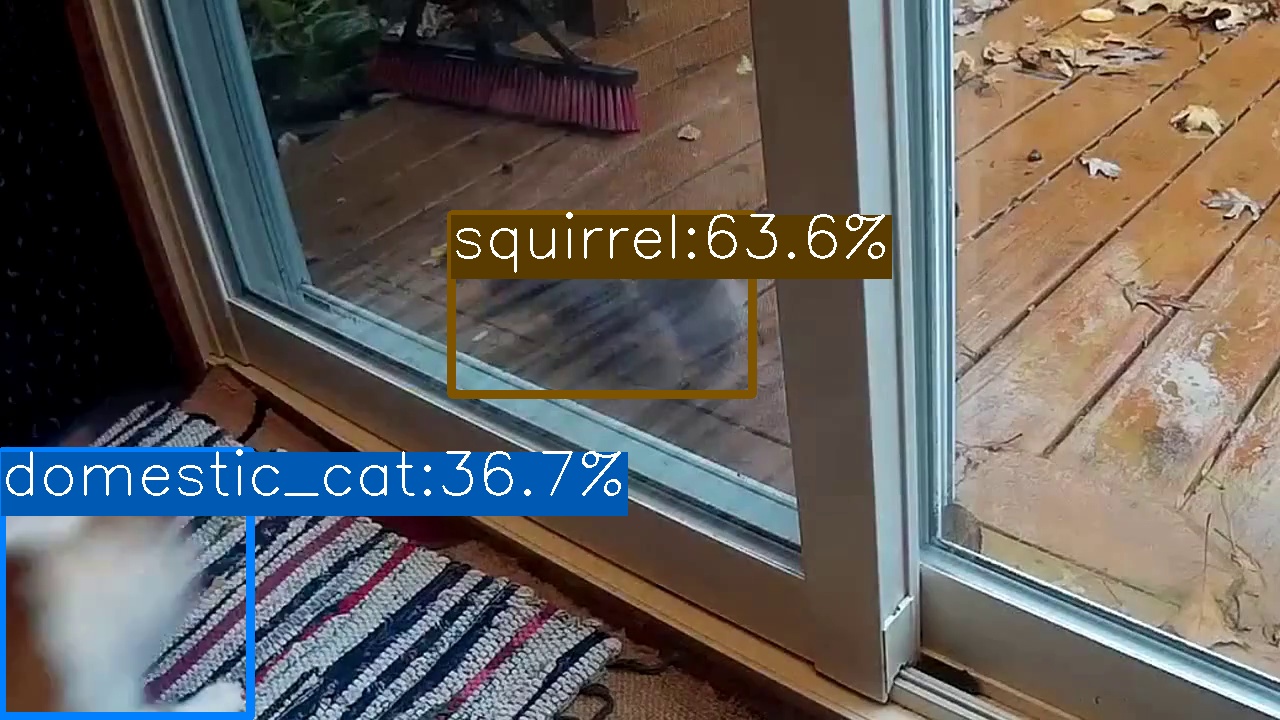}}\hfill
    \subfloat{\includegraphics[width=0.24\linewidth,height=0.08\paperheight]{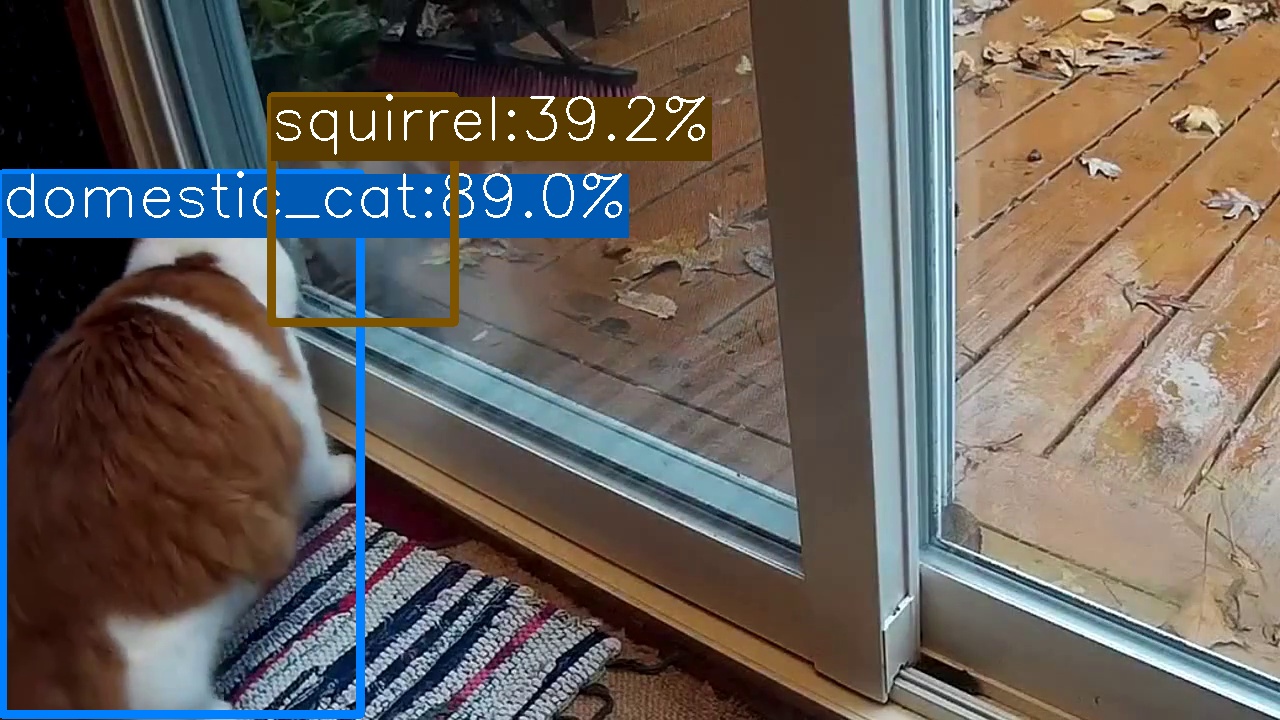}}\hfill
    \subfloat{\includegraphics[width=0.24\linewidth,height=0.08\paperheight]{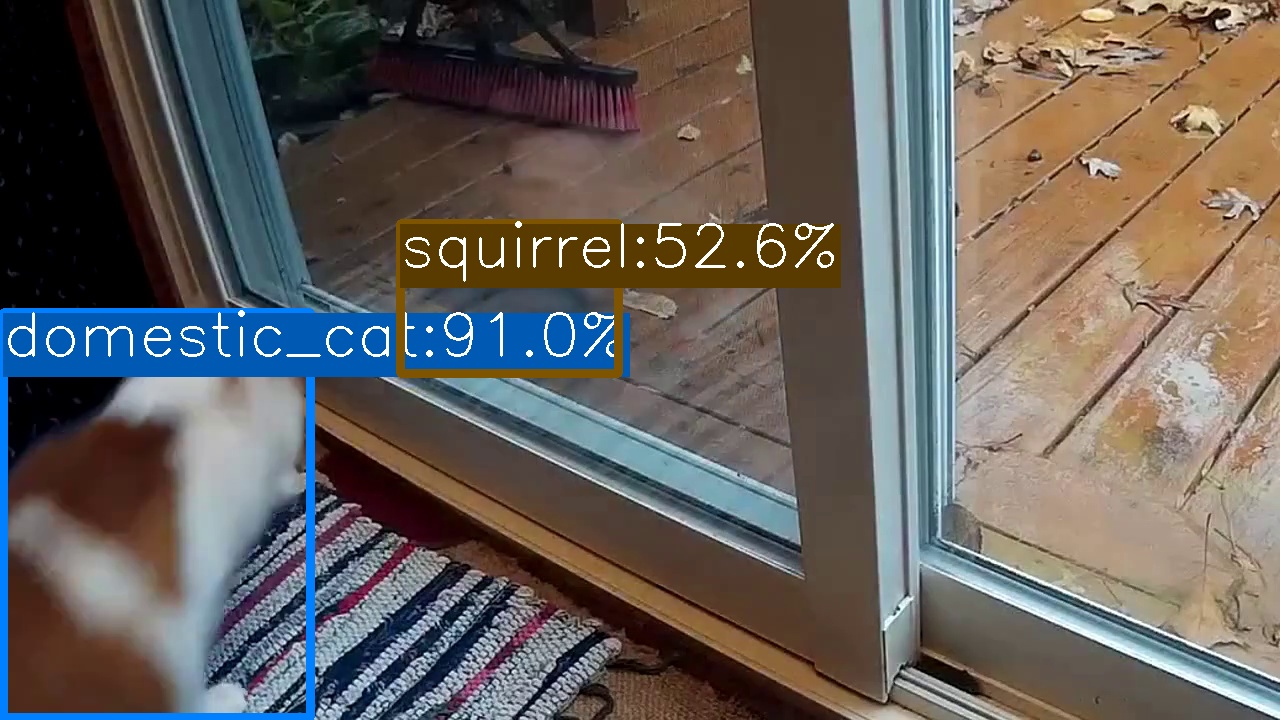}}\hfill
    \subfloat{\includegraphics[width=0.24\linewidth,height=0.08\paperheight]{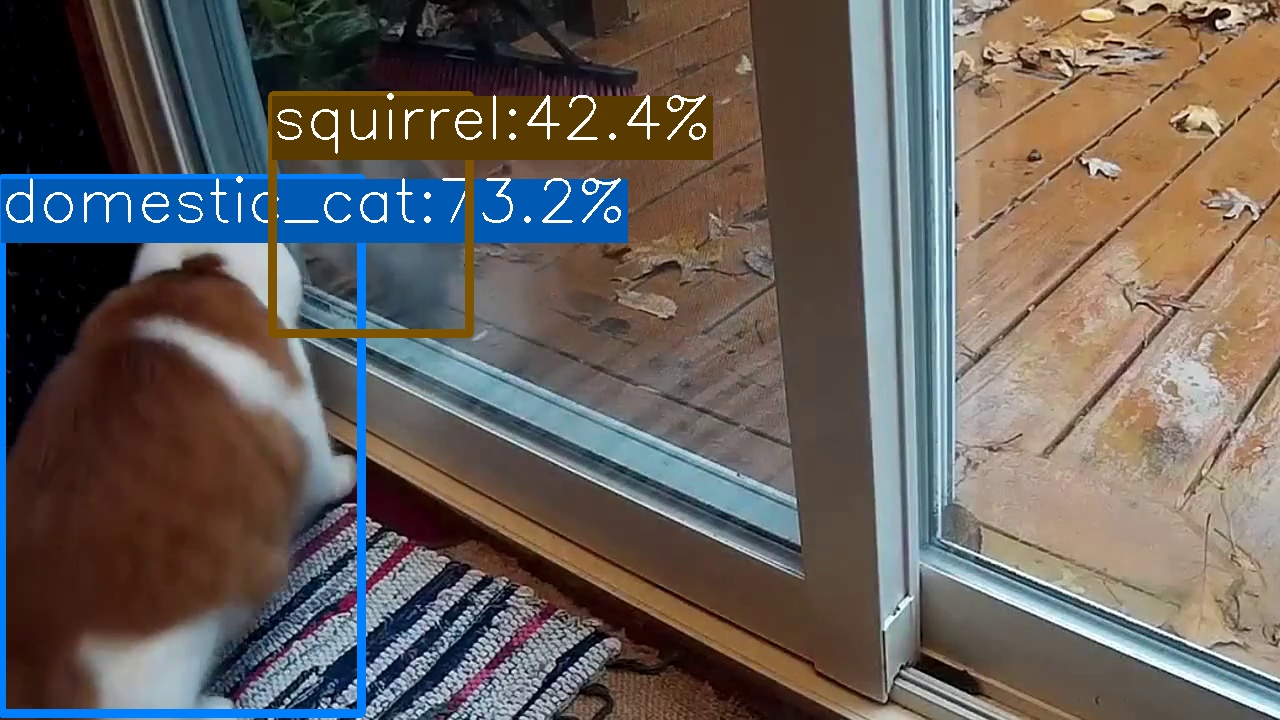}}
    \caption*{(a) Motion Blur}
    \subfloat{\includegraphics[width=0.24\linewidth,height=0.08\paperheight]{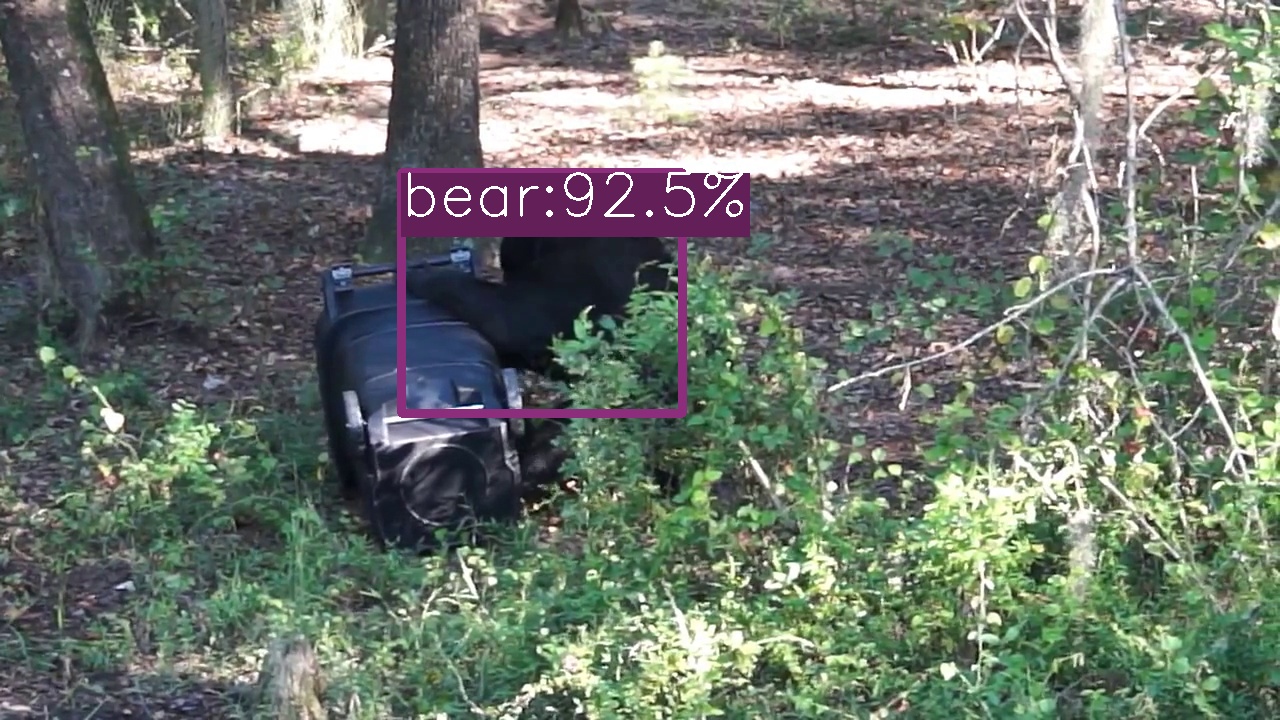}}\hfill
    \subfloat{\includegraphics[width=0.24\linewidth,height=0.08\paperheight]{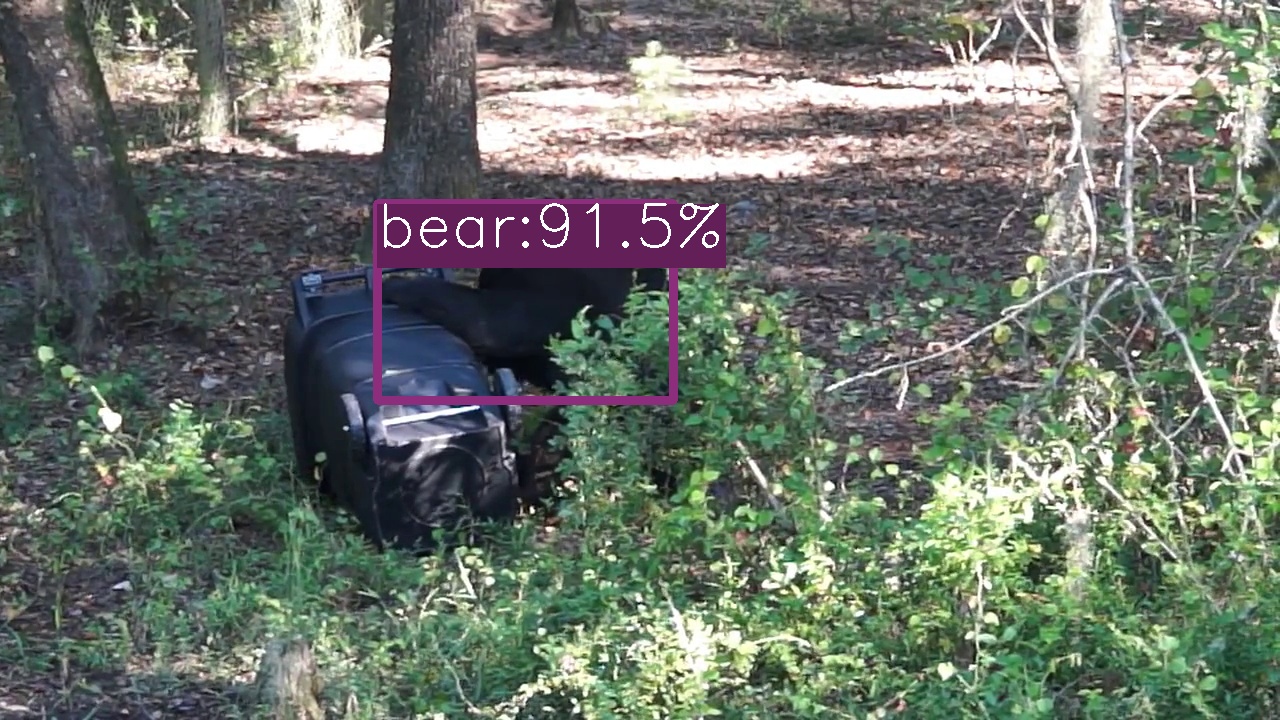}}\hfill
    \subfloat{\includegraphics[width=0.24\linewidth,height=0.08\paperheight]{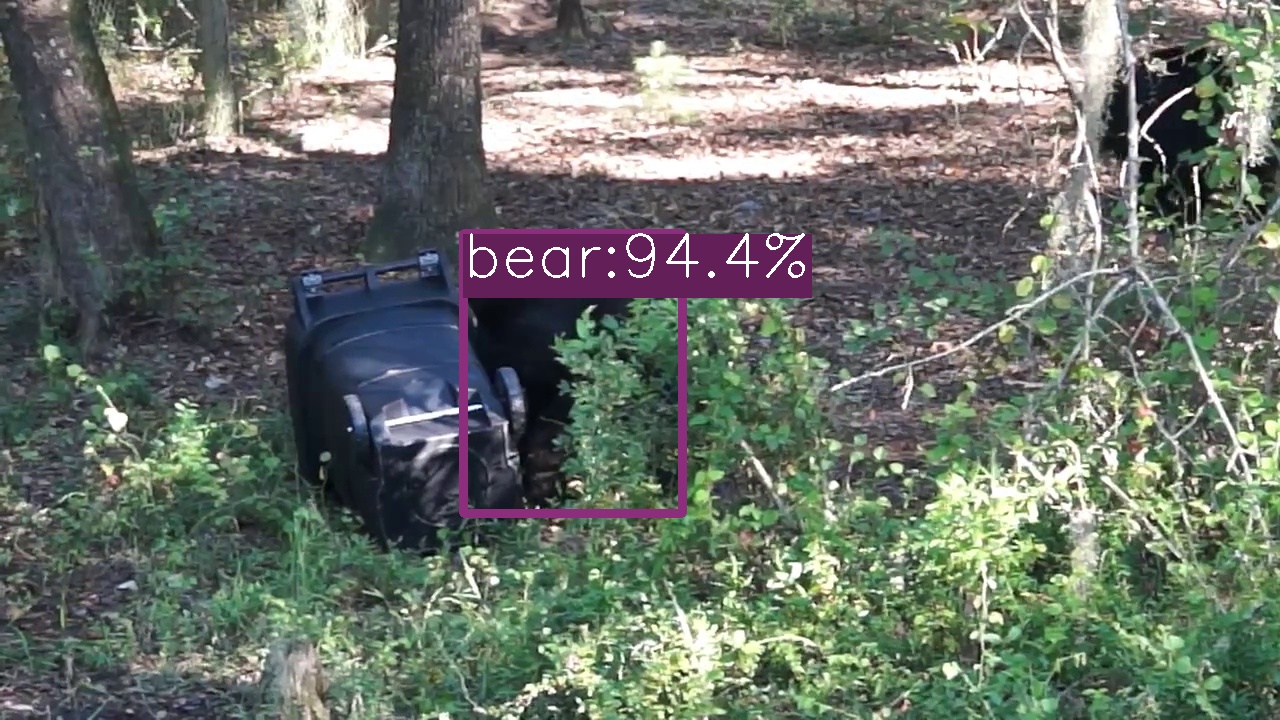}}\hfill
    \subfloat{\includegraphics[width=0.24\linewidth,height=0.08\paperheight]{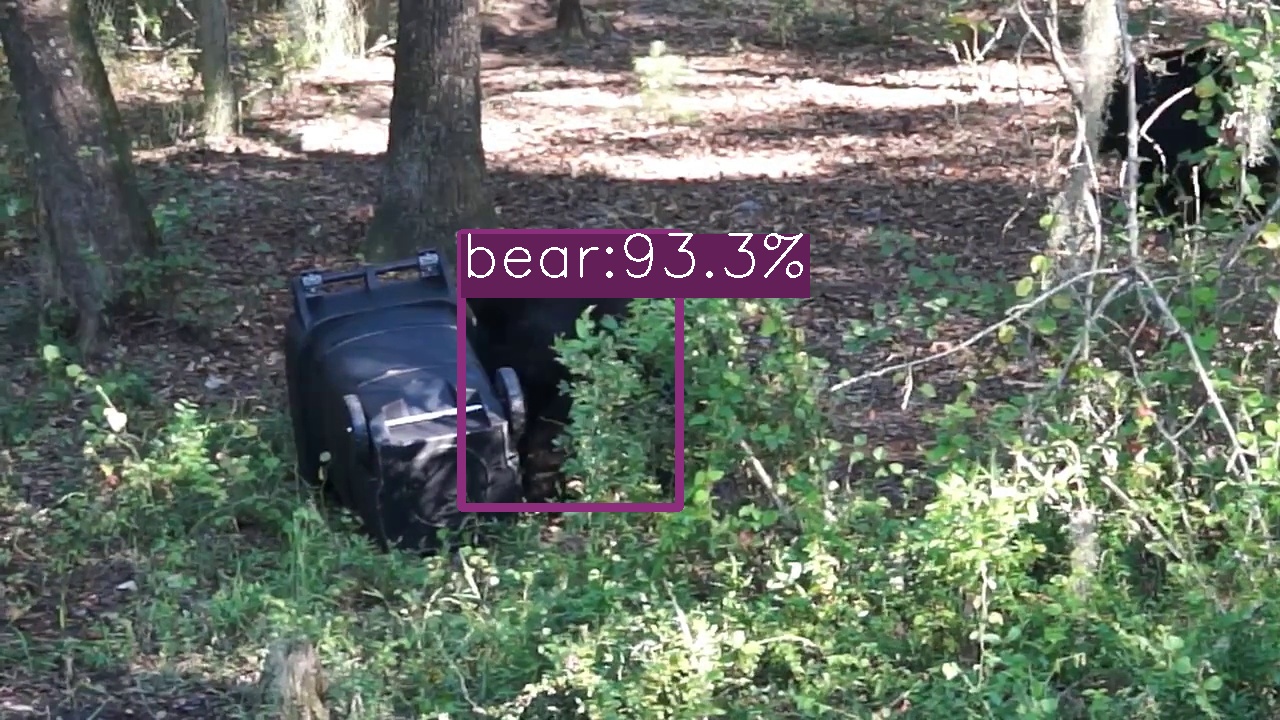}}\\
    \subfloat{\includegraphics[width=0.24\linewidth,height=0.08\paperheight]{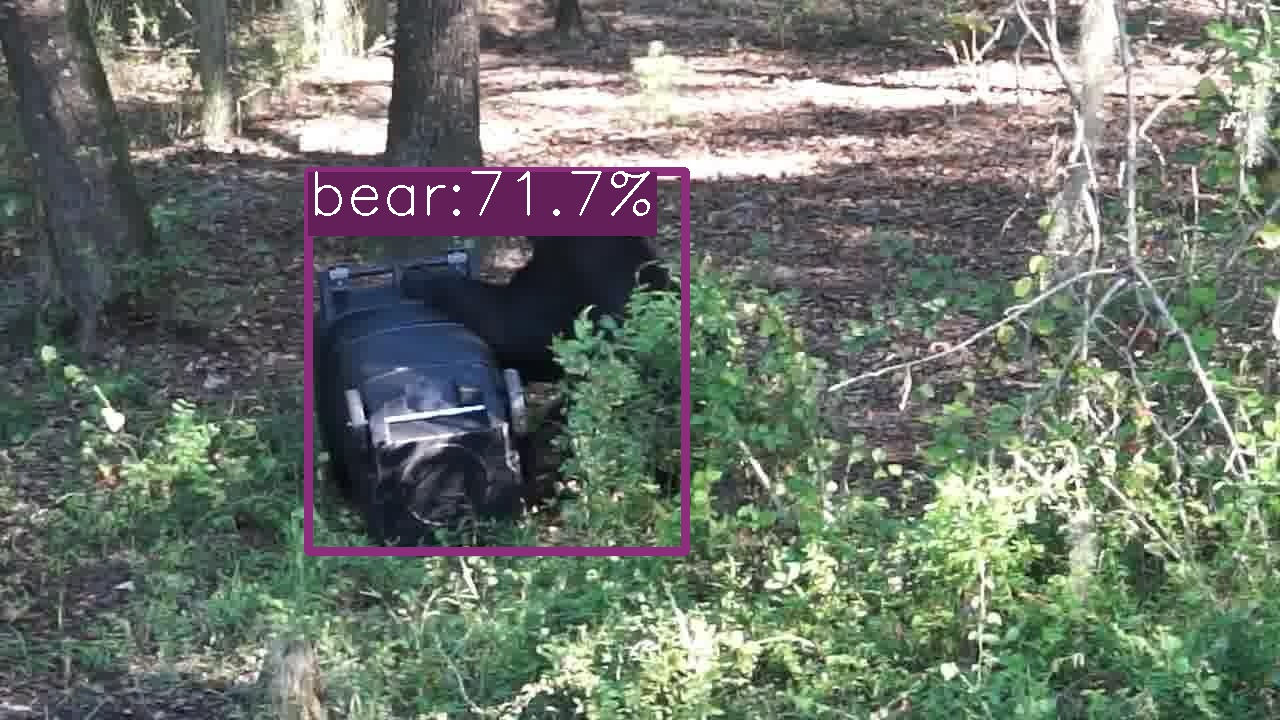}}\hfill
    \subfloat{\includegraphics[width=0.24\linewidth,height=0.08\paperheight]{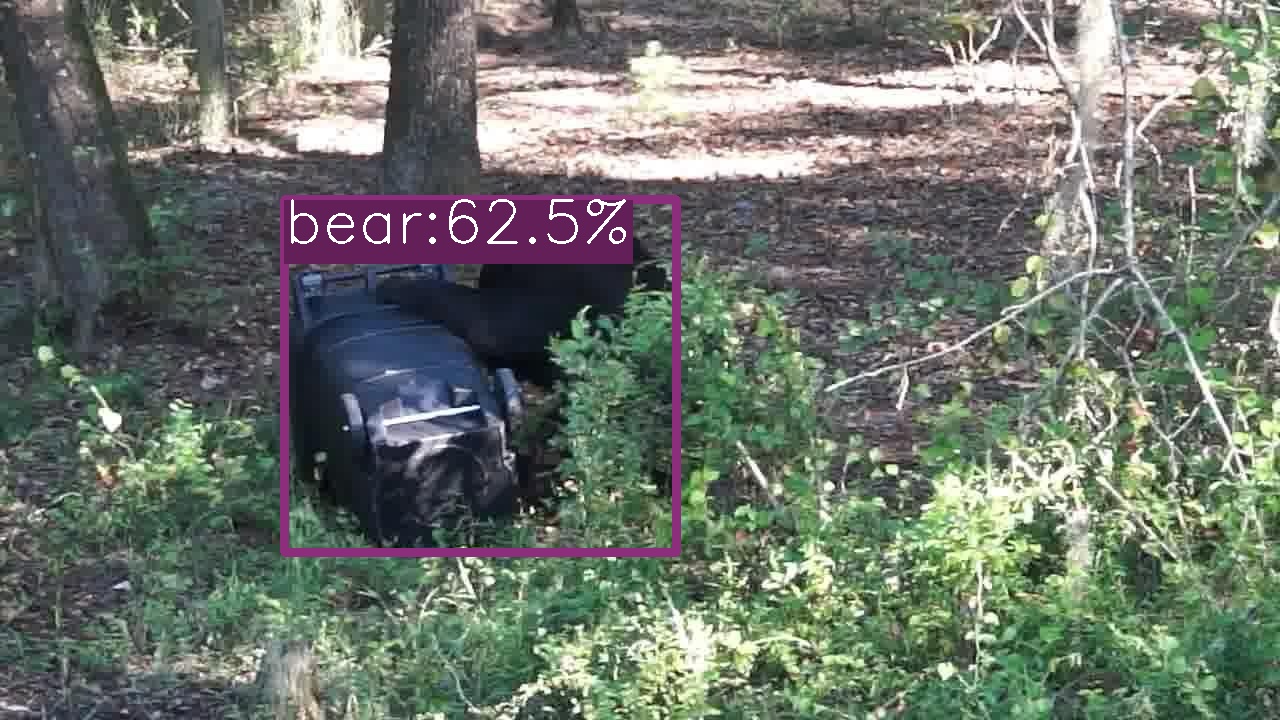}}\hfill
    \subfloat{\includegraphics[width=0.24\linewidth,height=0.08\paperheight]{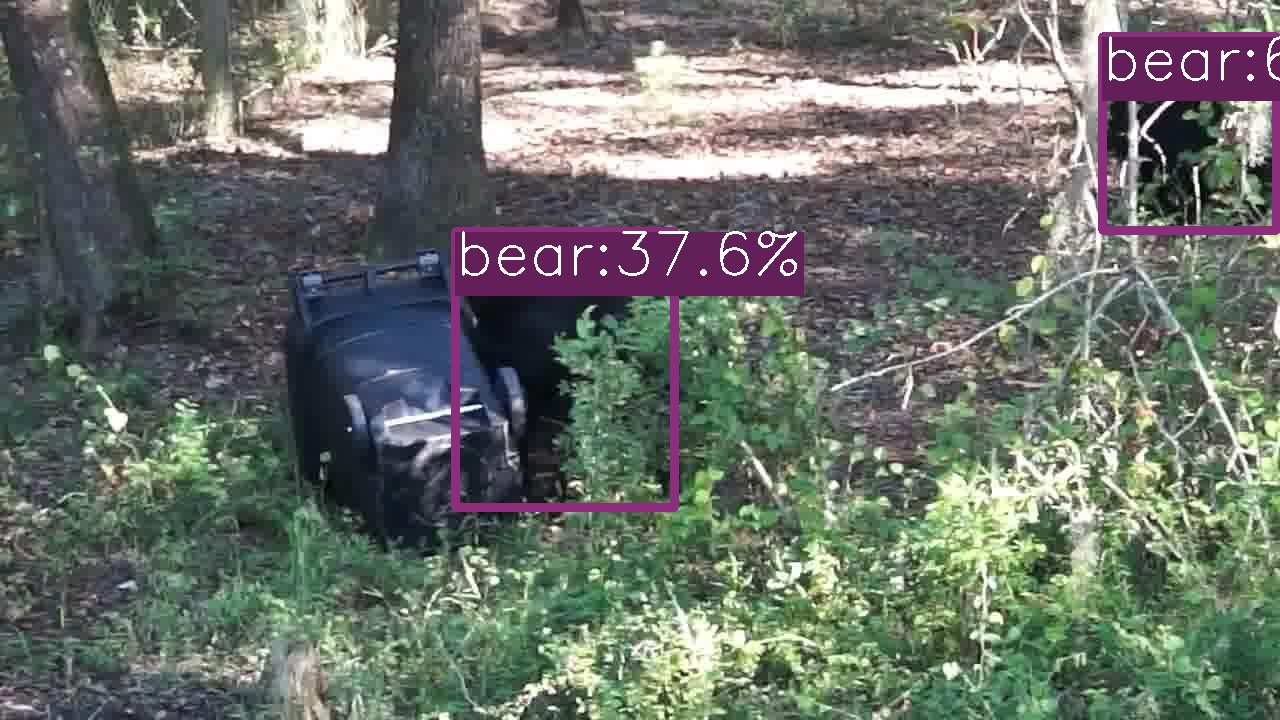}}\hfill
    \subfloat{\includegraphics[width=0.24\linewidth,height=0.08\paperheight]{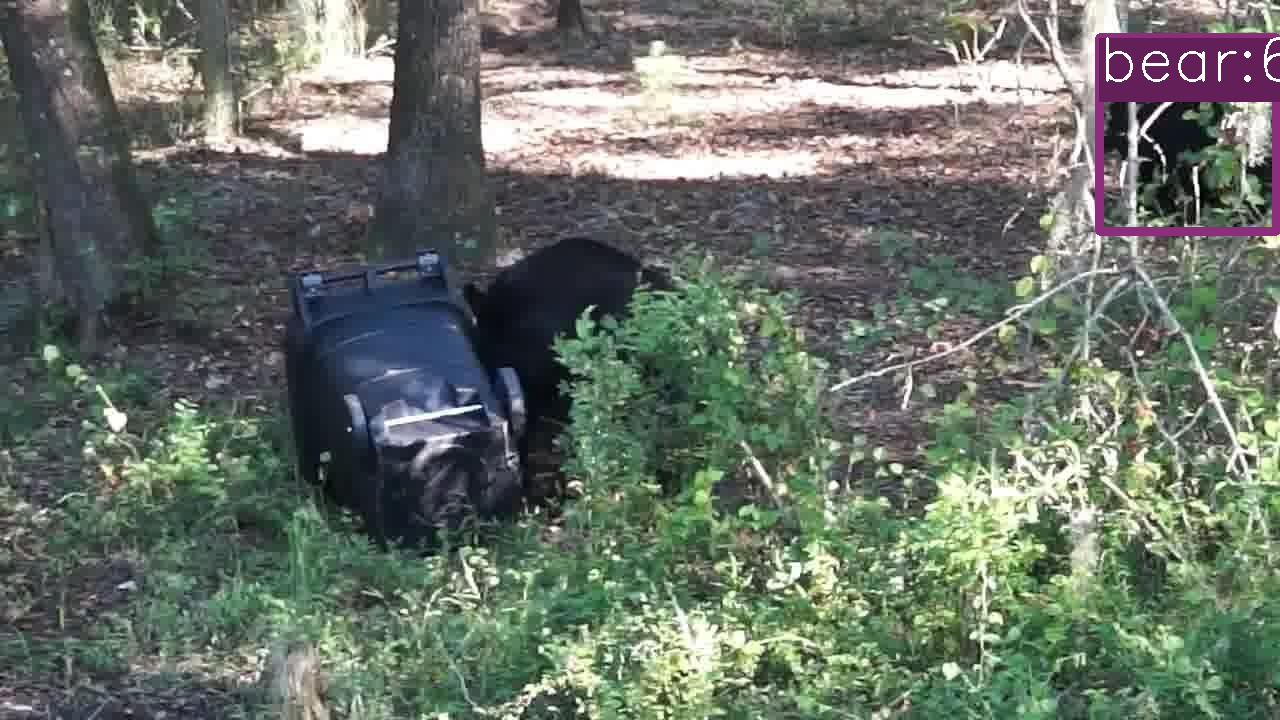}}\\
    \subfloat{\includegraphics[width=0.24\linewidth,height=0.08\paperheight]{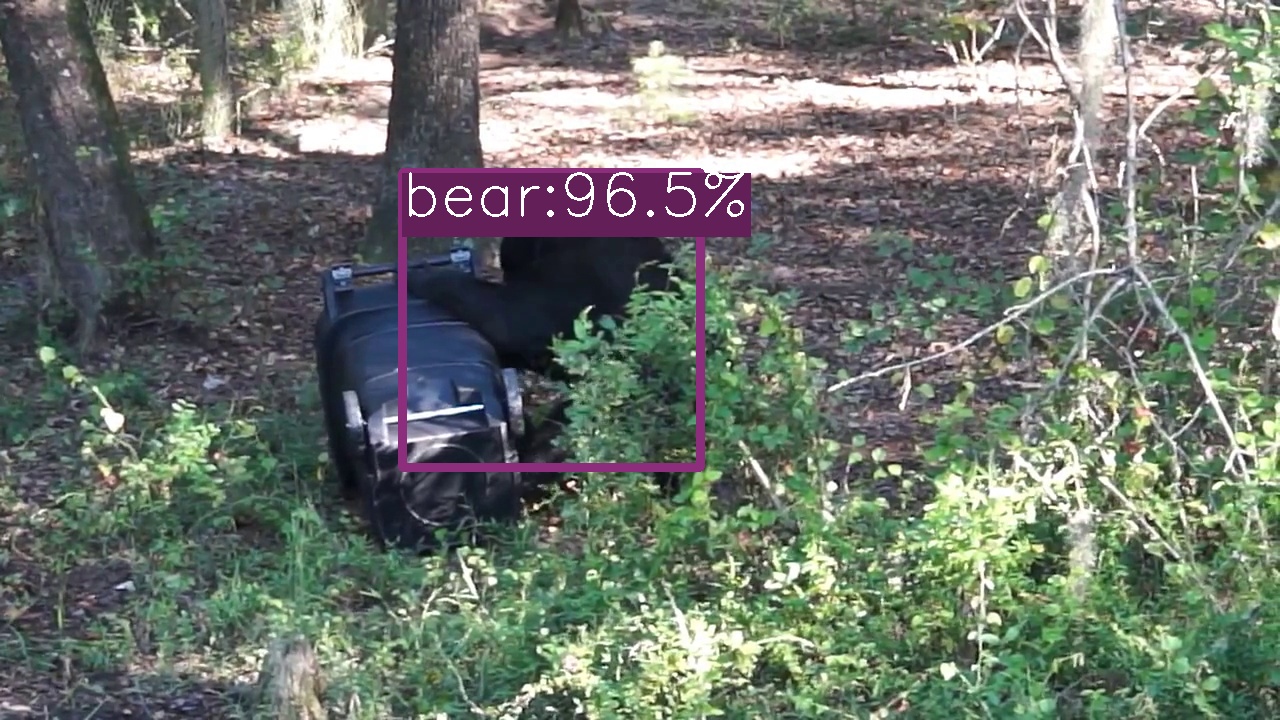}}\hfill
    \subfloat{\includegraphics[width=0.24\linewidth,height=0.08\paperheight]{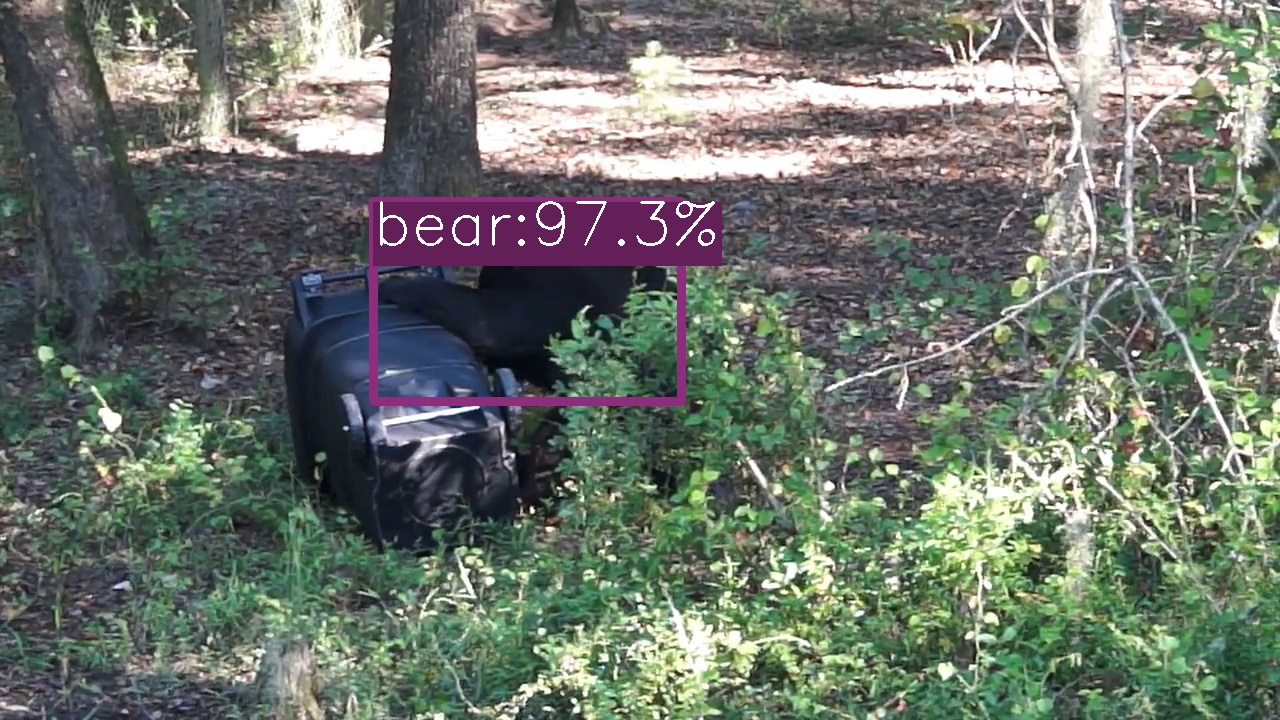}}\hfill
    \subfloat{\includegraphics[width=0.24\linewidth,height=0.08\paperheight]{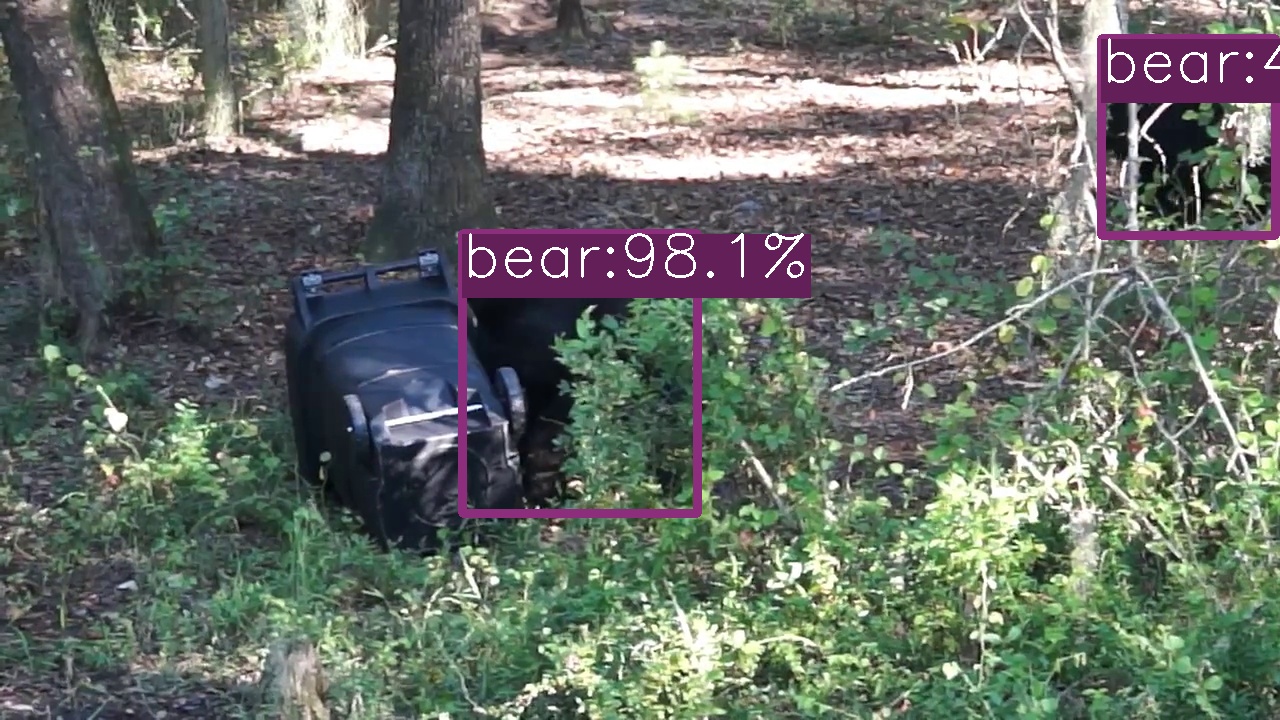}}\hfill
    \subfloat{\includegraphics[width=0.24\linewidth,height=0.08\paperheight]{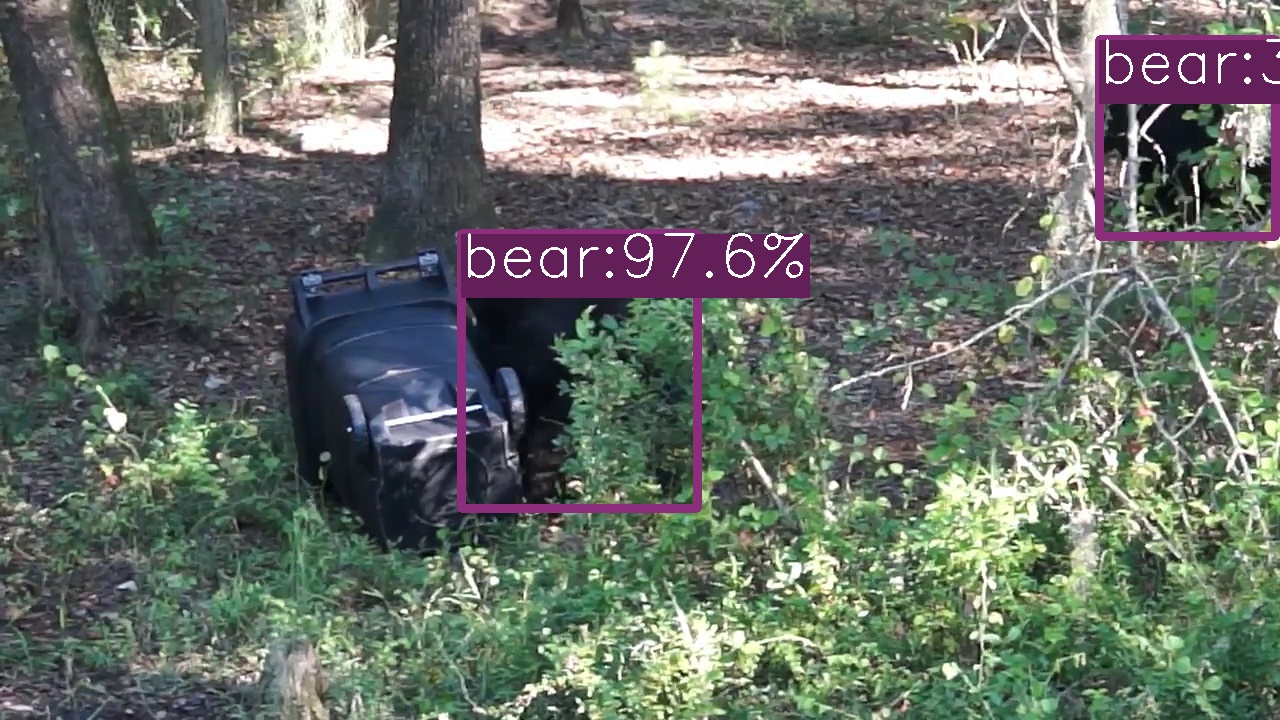}}
    \caption*{(b) Rare Pose}

    \subfloat{\includegraphics[width=0.24\linewidth,height=0.08\paperheight]{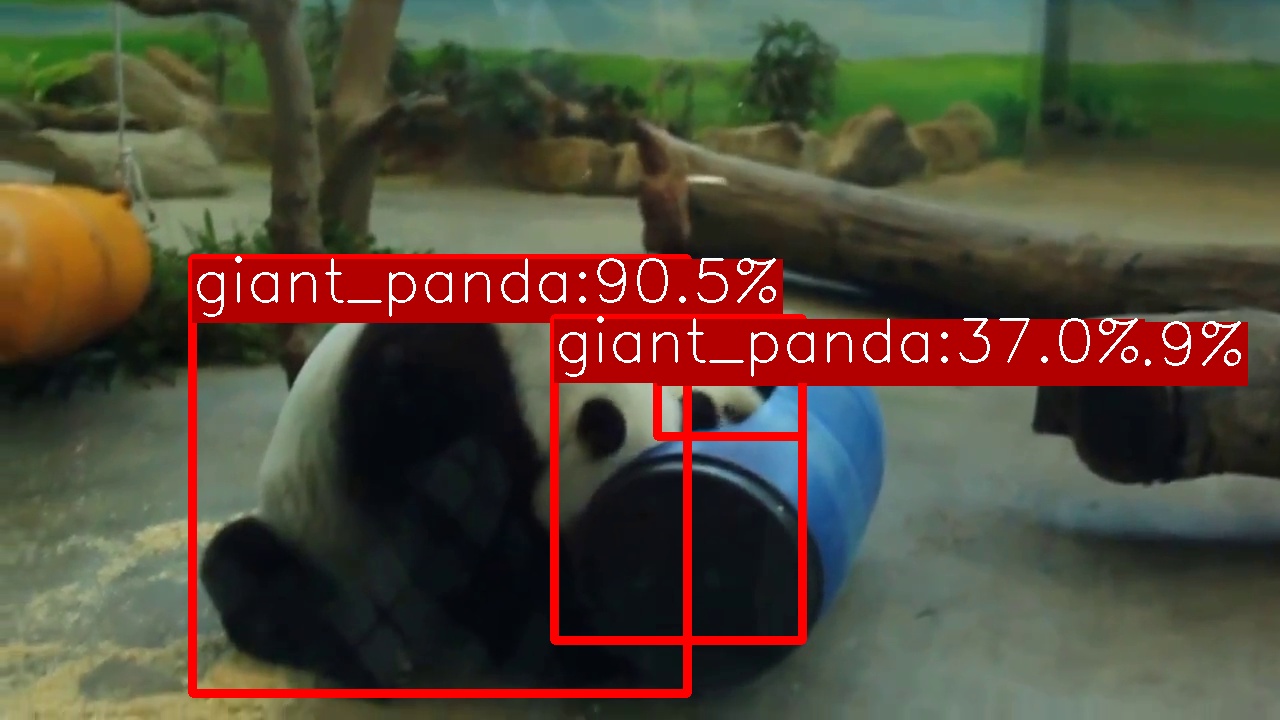}}\hfill
    \subfloat{\includegraphics[width=0.24\linewidth,height=0.08\paperheight]{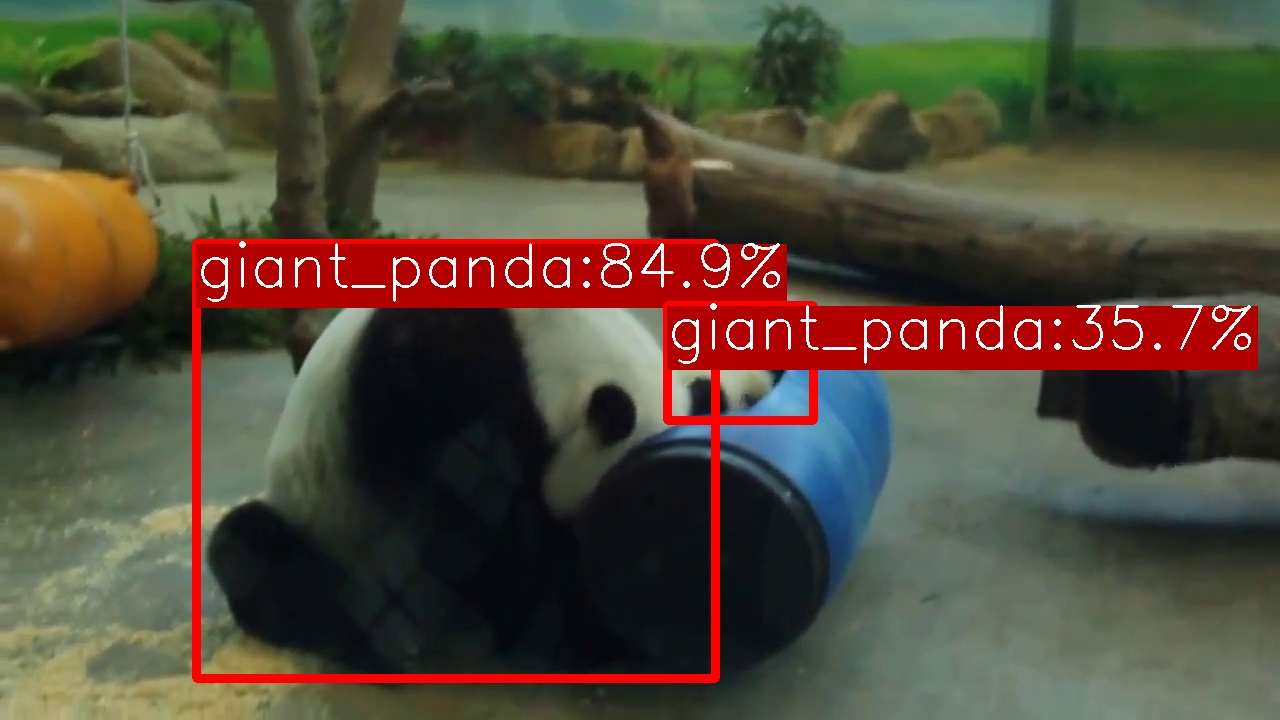}}\hfill
    \subfloat{\includegraphics[width=0.24\linewidth,height=0.08\paperheight]{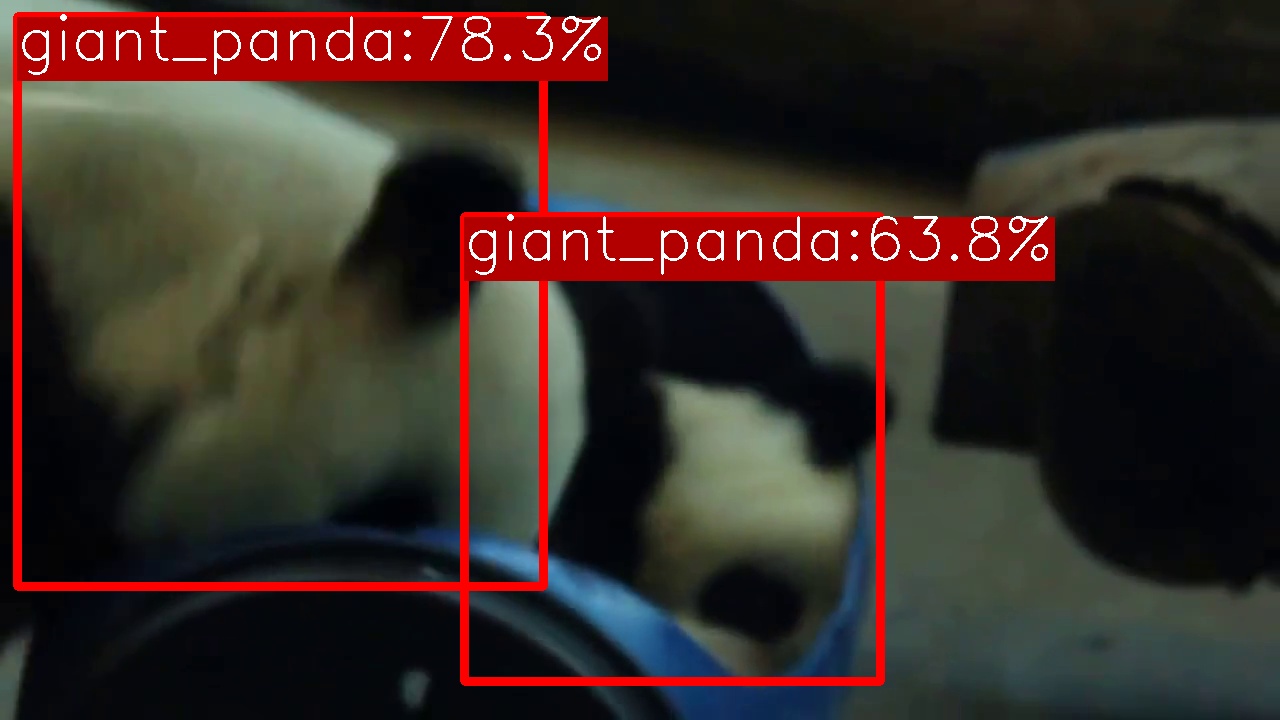}}\hfill
    \subfloat{\includegraphics[width=0.24\linewidth,height=0.08\paperheight]{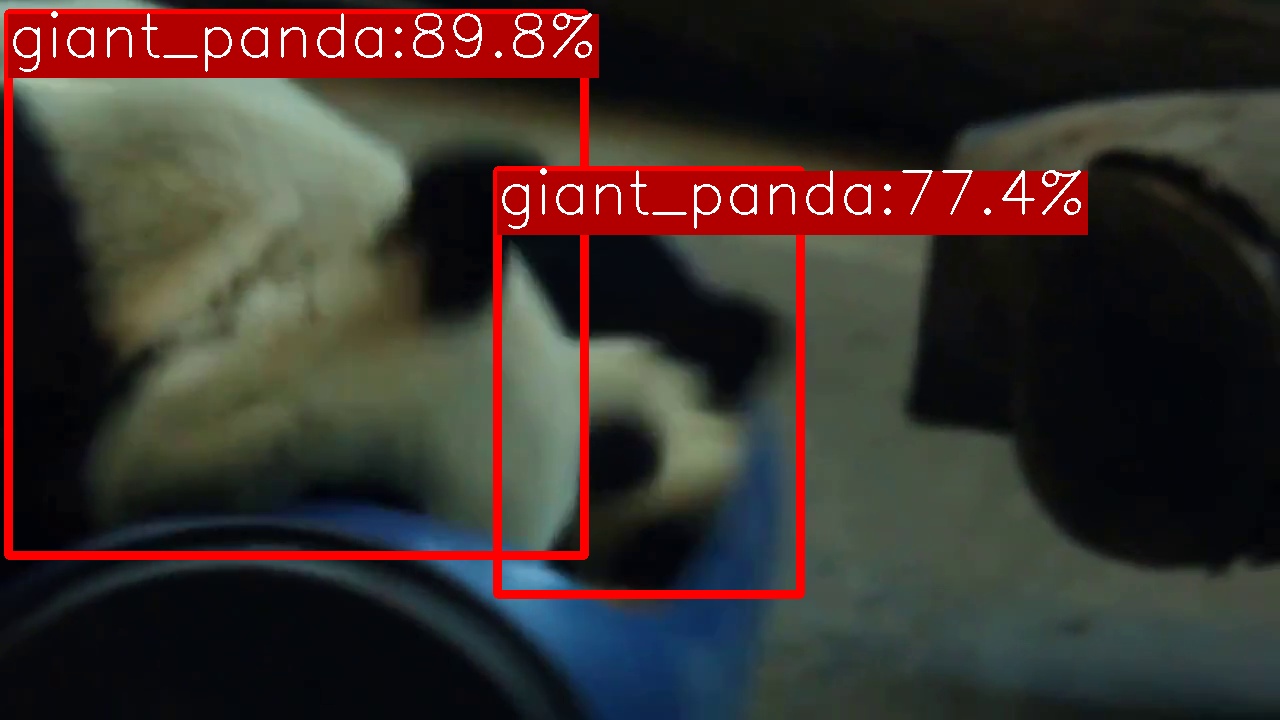}}\\
    \subfloat{\includegraphics[width=0.24\linewidth,height=0.08\paperheight]{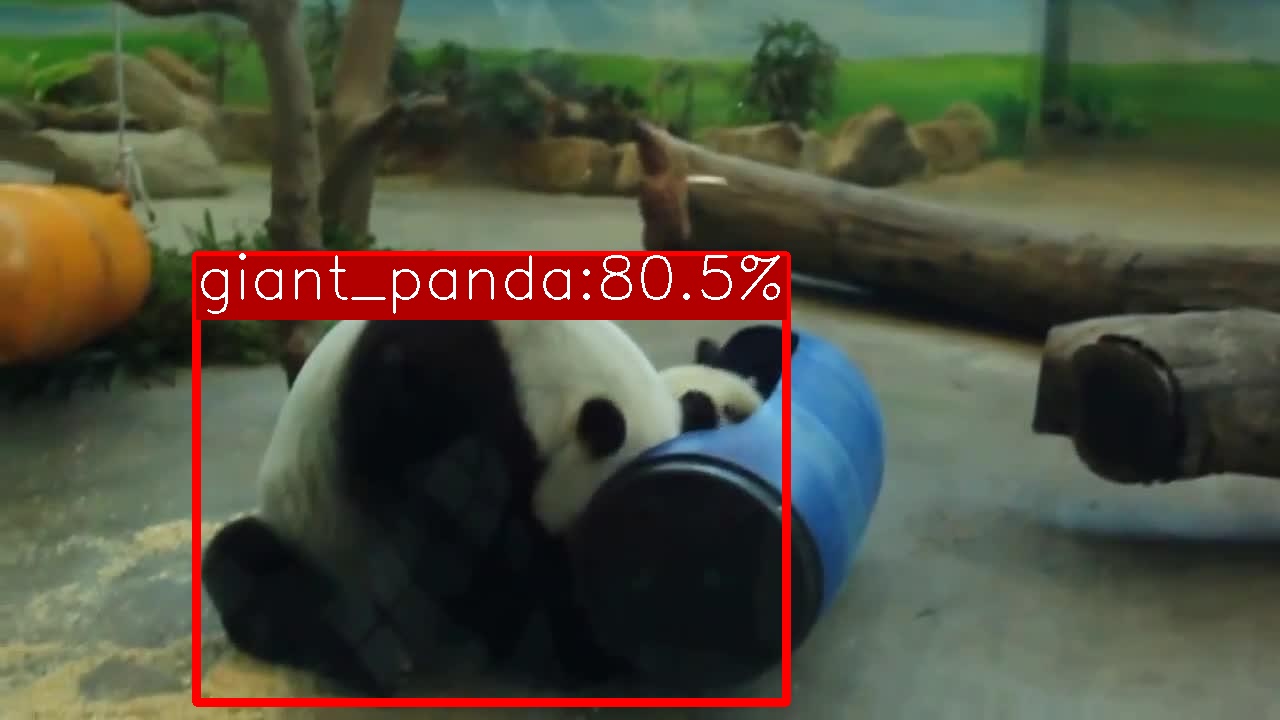}}\hfill
    \subfloat{\includegraphics[width=0.24\linewidth,height=0.08\paperheight]{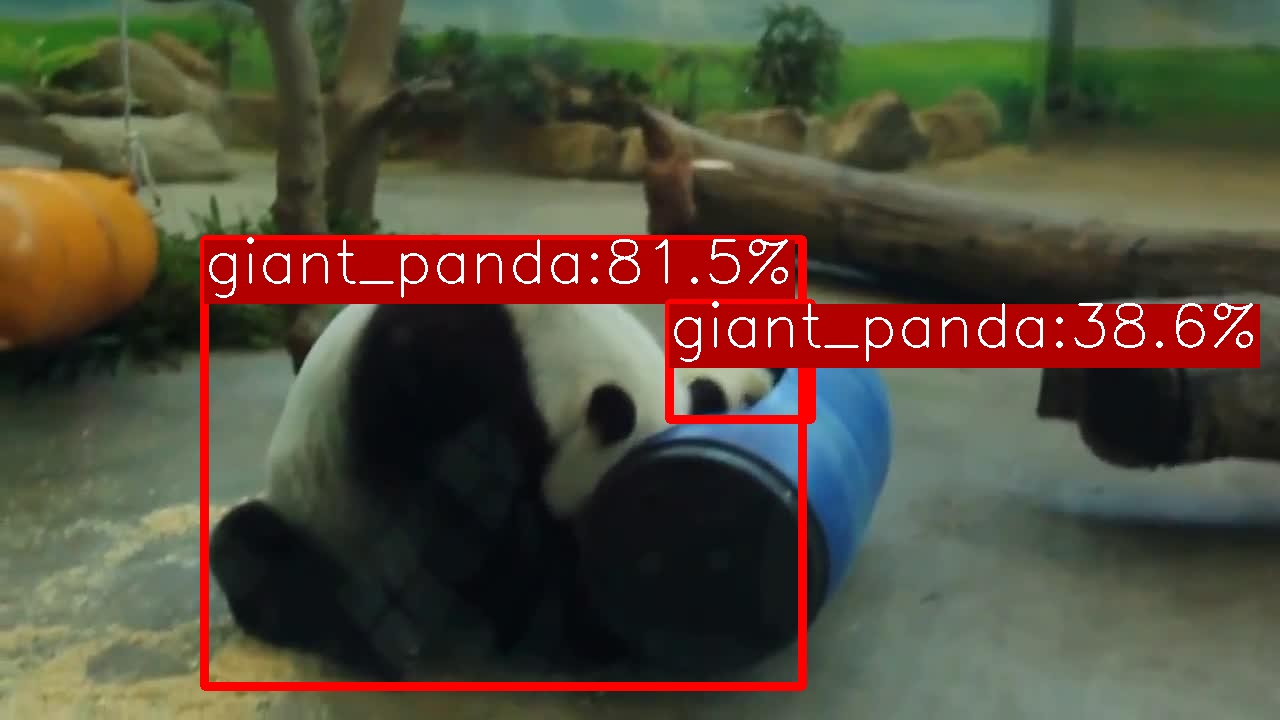}}\hfill
    \subfloat{\includegraphics[width=0.24\linewidth,height=0.08\paperheight]{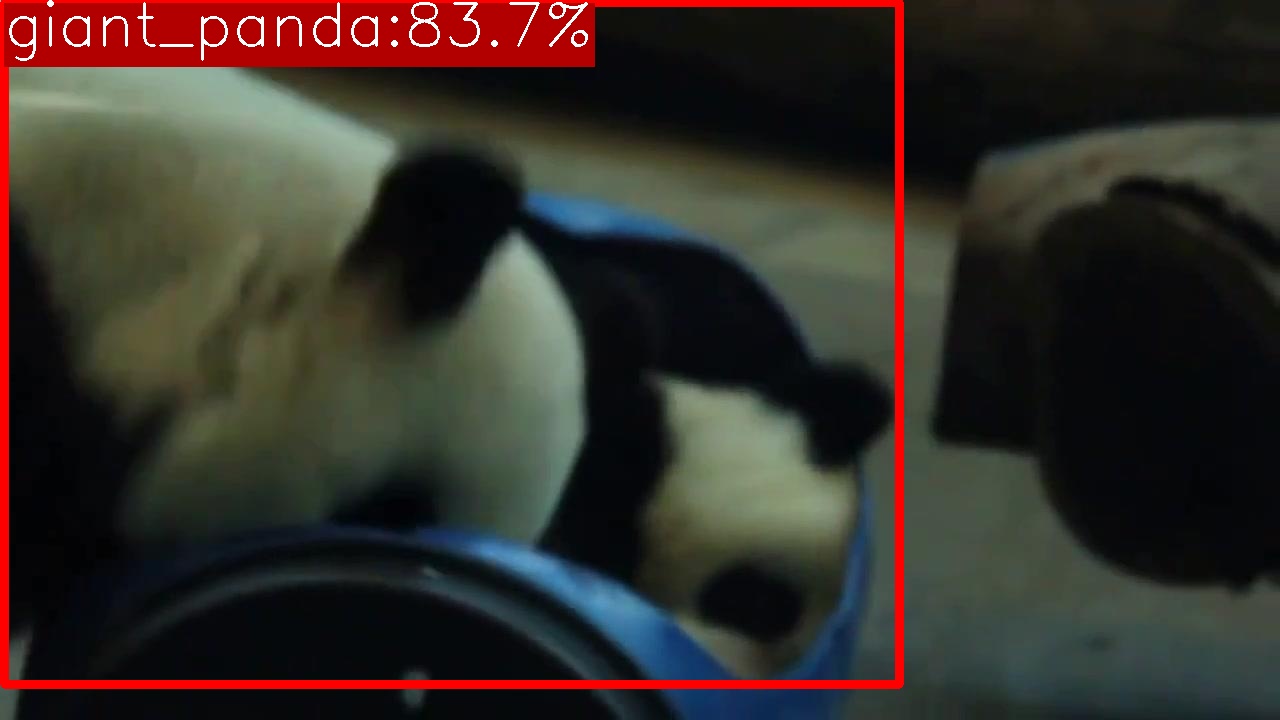}}\hfill
    \subfloat{\includegraphics[width=0.24\linewidth,height=0.08\paperheight]{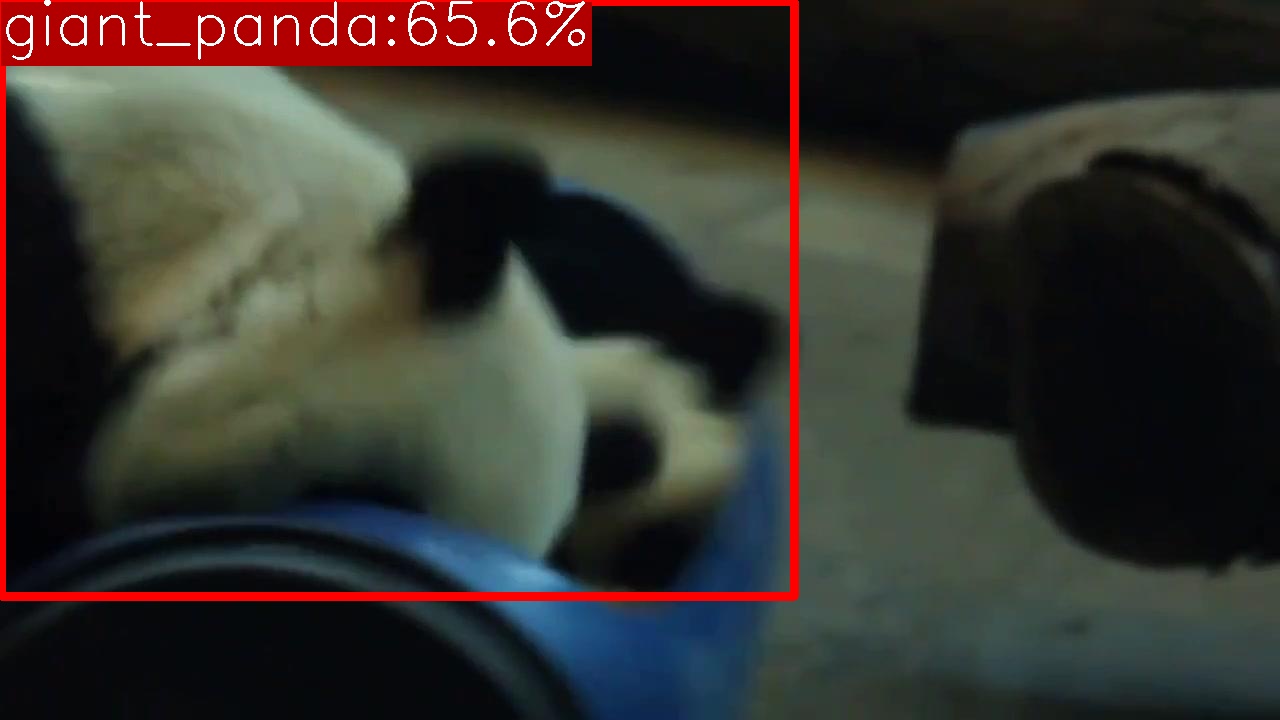}}\\
    \subfloat{\includegraphics[width=0.24\linewidth,height=0.08\paperheight]{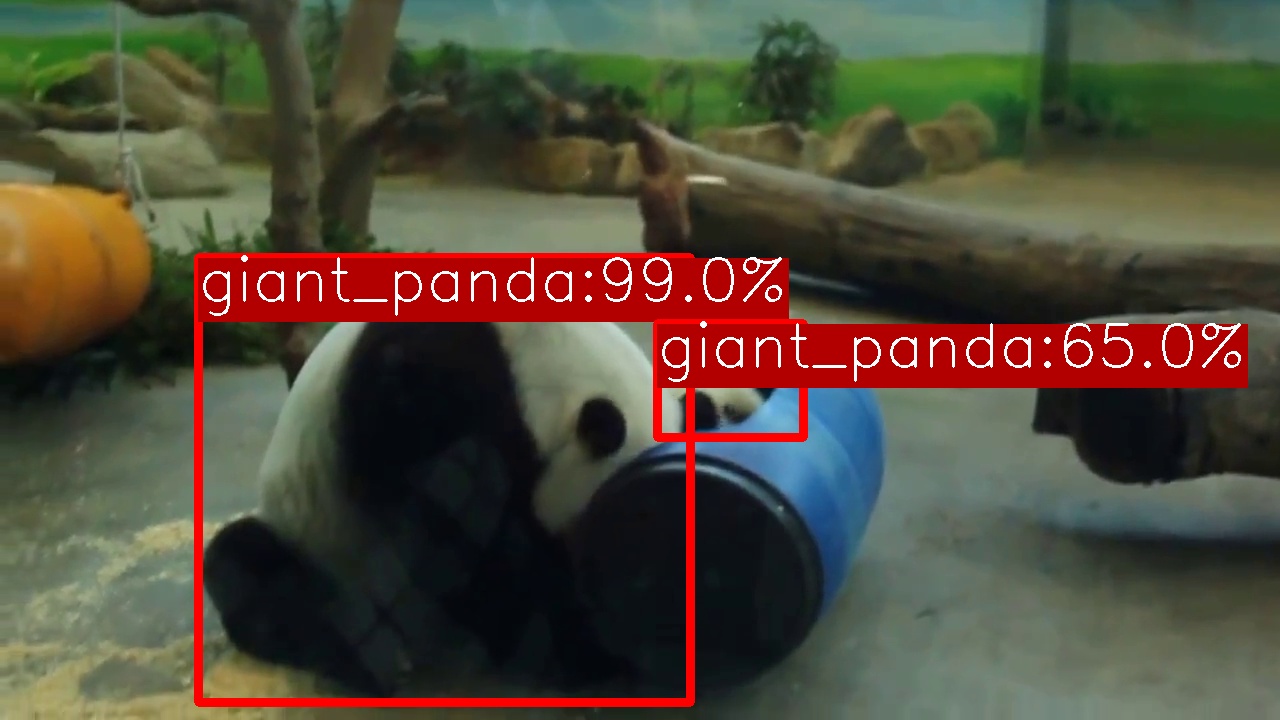}}\hfill
    \subfloat{\includegraphics[width=0.24\linewidth,height=0.08\paperheight]{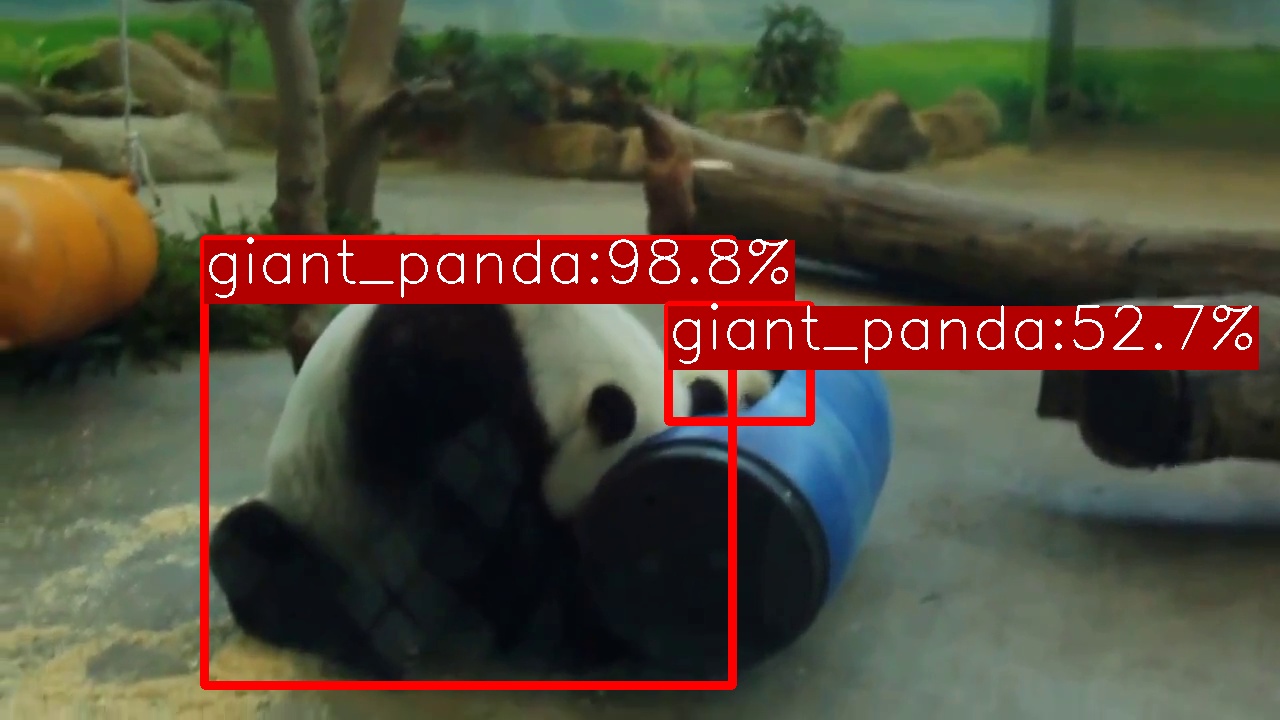}}\hfill
    \subfloat{\includegraphics[width=0.24\linewidth,height=0.08\paperheight]{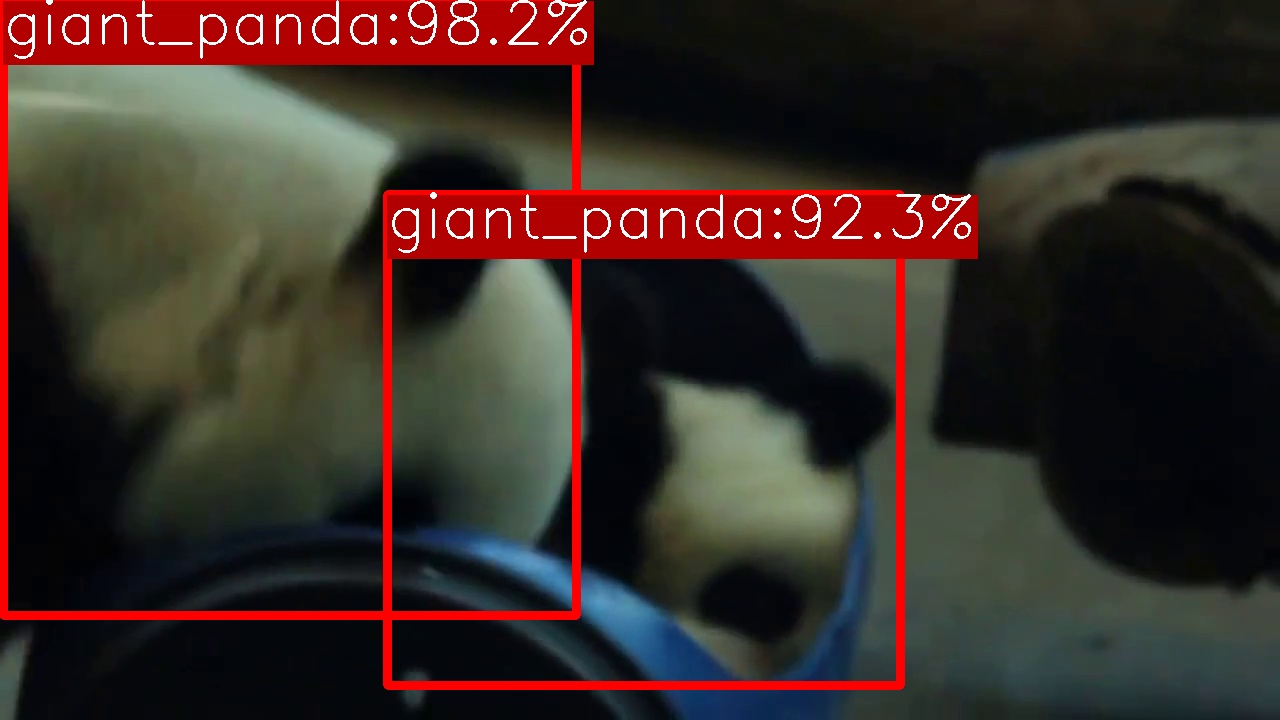}}\hfill
    \subfloat{\includegraphics[width=0.24\linewidth,height=0.08\paperheight]{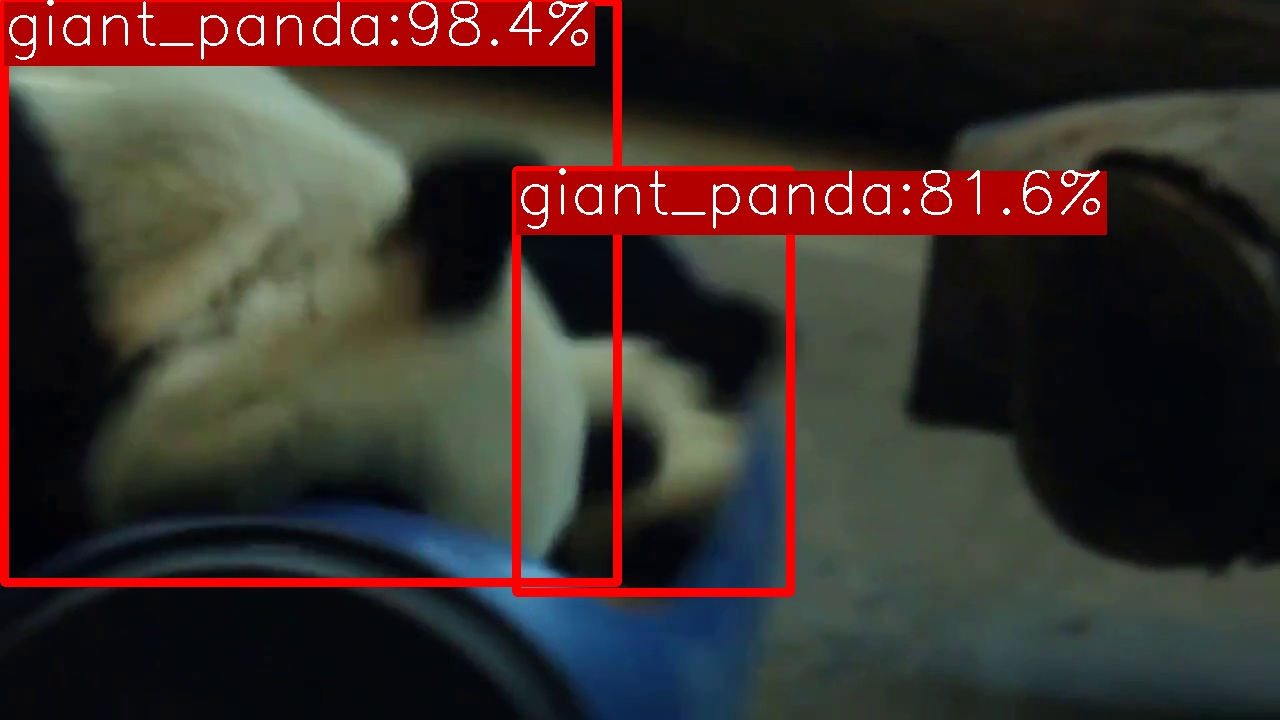}}
    \caption*{(c) Occlusion}

    \caption{Visual comparisons between YOLOV-SwinBase (1st row), YOLOV++-SwinBase (3rd row), and TransVOD-Lite (2nd row) with the same SwinBase backbone. Three cases suffer from different types of degradation: (a) motion Blur, (b) rare pose, and (c) occlusion. Our method exhibits its robustness against these challenging cases.}
    \label{fig:degradation}
\end{figure*}

In addition to quantitative experiments, we offer qualitative visualizations to intuitively show the effectiveness of our method on several samples from the VID dataset using YOLOV++, YOLOV, and the end-to-end Transformer-based TransVOD-Lite with the same SwinBase backbone, as illustrated in Fig.~\ref{fig:degradation}. 
To demonstrate the robustness of our model against various types of degradation, we select three challenging scenarios including (a) motion blur, (b) rare poses, and (c) occlusion. The visual results clearly reveal that our model can significantly boost the precision of predictions. For instance, in the second frame of (b), our model not only accurately predicts the locations but also correctly identifies the categories of objects under rare pose conditions. Additionally, it reliably detects occluded objects with high confidence. We would also like to highlight that our model operates approximately $3\times$ faster than TransVOD-Lite.

\subsection{Ablation Study}
\subsubsection{On the Effectiveness of Feature Selection and Aggregation Strategies} 
To ascertain the effectiveness of the affinity matrix (A.M.), the average pooling over reference frame features (A.P.) and different feature selection strategies, we carry out a comparative analysis on our model with and without these components. Omitting both A.M. and A.P. in FAM degrades our approach to a basic Multi-Head Attention mechanism. The findings, as presented in Tab.~\ref{table:effectiveness of ours}, indicate that incorporating A.M. and A.P. can effectively make the model to aggregate features and capture more nuanced semantic representations from the one-stage detector. Specifically, when utilizing the TopK+NMS feature selection, the inclusion of A.M. and A.P. elevates the AP50 metric from 75.4\% to 77.3\%. However, please note that the selected features by TopK+NMS are sparse and misaligned with the label assignment strategy intended for dense predictions. Consequently, using the IoU score predicted by FAM$r$ leads to a decrease in AP50 to 75.3\%. Our customized feature selection, which uses a specific threshold to select nearly dense foreground proposals, is able to address the misalignment. Thus, under the Threshold feature selection pipeline, refining both the classification and IoU scores concurrently elevates the AP50 from 76.4\% to 78.7\%, compared to refining only the classification score. 

\setlength{\tabcolsep}{4pt}
\begin{table}[t]
\begin{center}
\caption{Effectiveness of different feature selection and feature aggregation strategies}
\label{table:effectiveness of ours}
\begin{tabular}{l|ccccc}
\hline\noalign{\smallskip}
Pipeline & A.M. &A.P.& FAM$_r$ &AP50 ($\%$)&Time(ms)\\
\noalign{\smallskip}
\hline
\noalign{\smallskip}
Baseline & - & - & - &$69.5$ & 1.4 \\
TopK+NMS & - & - & - & $75.4_{\uparrow 5.9 }$ & 3.8\\
TopK+NMS &  \checkmark & - & -  &$76.9_{\uparrow 7.4 }$& 4.0 \\
TopK+NMS &  \checkmark& \checkmark & -  &$77.3_{\uparrow 7.8 }$& 4.0\\
TopK+NMS &  \checkmark& \checkmark &\checkmark  &$75.3_{\uparrow 5.8 }$& 4.5\\
Thresh & \checkmark& \checkmark & -  &$76.4_{\uparrow 6.9 }$& 4.6 \\
Thresh & \checkmark& \checkmark & \checkmark &$78.7_{\uparrow 9.2 }$& 5.3\\
\hline
\end{tabular}
\end{center}
\end{table}
\setlength{\tabcolsep}{1.4pt}

\subsubsection{On the Reference Frame Sampling Strategy}  Effective frame sampling is crucial for VOD. Prior research on two-stage methods has explored various global and local sampling strategies~\cite{wu2019sequence,gong2021temporal,chen2020memory}. Global sampling randomly selects $F_g$ frames from the entire video, whereas local sampling picks out only $F_l$ consecutive frames. To examine the impacts of these two different schemes, we vary the number of reference frames in both global and local settings. As can be seen from Tab.~\ref{table:effectiveness of local and globle number}, the principles of feature aggregation for two-stage detectors remain valid. Notably, using just 4 reference frames in the global mode obtains superior performance over using 39 frames in the local mode, supporting the insight from previous studies~\cite{wu2019sequence,gong2021temporal}. Global frame sampling can more effectively address feature degradation via long-term feature aggregation, which overcomes challenges such as motion blur, rarity, and occlusion. However, incorporating additional reference frames consistently increases both time and memory usage due to the quadratic complexity of self-attention mechanisms. For a trade-off, we employ a global sampling strategy with $F_g= 31$ as default for subsequent experiments. 

\setlength{\tabcolsep}{4pt}

\begin{table}[t]
\begin{center}
\caption{Influence by the number of global reference frames $F_g$ and local reference frames $F_l$}
\label{table:effectiveness of local and globle number}
\begin{tabular}{c|cccccc}
\hline\noalign{\smallskip}
$F_g$ & 4 & 7 & 15 & 23 & 31 & 39\\
\noalign{\smallskip}
\hline
\noalign{\smallskip}
AP50 ($\%$) & 75.7 & 76.9 & 78.0 & 78.4 & 78.7 & 78.7 \\
\hline
\hline\noalign{\smallskip}
$F_l$ & 4 & 7 & 15 & 23 & 31 & 39\\
\noalign{\smallskip}
\hline
\noalign{\smallskip}
AP50 ($\%$) & 70.5 & 71.2 & 72.3 & 73.1 & 73.6 & 74.1 \\
\hline
\end{tabular}
\end{center}
\end{table}

\subsubsection{On the Threshold of Average Pooing of Reference Frame Features}
Here, we test the effect of different thresholds for average pooling over reference features. Table~\ref{table:effectiveness of thresh in average} lists the numerical results. As can be viewed, when a large portion of the features participate in the average pooling, \emph{i.e.} $\tau=0$, the AP50 is merely $77.1\%$. Lifting the selection standard results in better performance. When $\tau$ falls in $[0.65, 0.75]$, the accuracy keeps steady and achieves an AP50 of $78.7\%$. When $\tau=1$, the average pooling is equivalent to only duplicating $\operatorname{SA(\mathcal{F})}$, the accuracy of which drops to $78.3\%$. Dynamically determining the threshold for different cases is desired and left as our future work. For the rest experiments, we adopt $\tau=0.75$ as default. 
\setlength{\tabcolsep}{4pt}
\begin{table}[t]
\begin{center}
\caption{Influence of the threshold $\tau$ in average pooling over reference features}
\label{table:effectiveness of thresh in average}
\begin{tabular}{c|ccccccc}
\hline\noalign{\smallskip}
 $\tau$ & 0 & 0.2 & 0.5 & 0.65 & \textbf{0.75} & 0.85 & 1\\
\noalign{\smallskip}
\hline
\noalign{\smallskip}
AP50 ($\%$) & 77.1 & 77.3 & 78.5 & 78.7 & 78.7 & 78.4 & 78.3 \\
\hline
\end{tabular}
\end{center}
\end{table}

\setlength{\tabcolsep}{4pt}
\begin{table*}[t]
\begin{center}
\caption{Effectiveness of our strategy on models with different sizes}
\label{table:effectiveness of diff model}
\begin{tabular}{l|ccccccc}
\hline\noalign{\smallskip}
Models&Params&GFLOPs&Time (ms)&AP50 ($\%$)&mAP$_{slow}$($\%$)&mAP$_{medium}$($\%$)&mAP$_{fast}$($\%$)\\
\noalign{\smallskip}
\hline
\noalign{\smallskip}
YOLOX-S & 8.95M & 21.63 & 1.4 &$69.5$&$76.8$&$66.8$&$50.0$\\
YOLOV-S & 10.28M & 26.18 & 4.0   &$77.3_{\uparrow 7.8}$&$82.3_{\uparrow 5.5}$&$75.1_{\uparrow 8.3}$&$58.6_{\uparrow 8.6}$       \\
YOLOV++-S & 11.39M & 26.61 & 5.3 & $78.7_{\uparrow 9.2 }$&$83.7_{\uparrow 6.9}$&$77.0_{\uparrow 10.2}$&$62.3_{\uparrow 12.3}$\\
\hline
YOLOX-L & 54.17M & 125.90 & 4.2 &  $76.5$&$81.9$&$74.5$&$58.9$\\
YOLOV-L & 59.45M & 143.10 & 6.3 &$83.6_{\uparrow 7.1}$&$86.6_{\uparrow 4.7}$&$82.6_{\uparrow 8.1}$&$68.9_{\uparrow 10.0}$  \\ 
YOLOV++-L & 63.85M & 143.96 & 7.6 &  $84.2_{\uparrow 7.7}$&$87.3_{\uparrow 5.4}$&$82.9_{\uparrow 8.4}$&$70.6_{\uparrow 11.7}$\\
\hline
YOLOX-SwinT & 45.94M & 110.11 & 5.5 &  $79.2$&$85.7$&$76.3$&$61.8$\\
YOLOV-SwinT & 51.30M & 127.31 & 7.9 &  $85.6_{\uparrow 6.4}$&$90.8_{\uparrow 5.1}$&$84.4_{\uparrow 8.1}$&$71.3_{\uparrow 9.5}$       \\
YOLOV++-SwinT & 55.63M & 128.17 & 8.4 &  $86.6_{\uparrow 7.4}$&$90.9_{\uparrow 5.2}$&$85.8_{\uparrow 9.5}$&$73.9_{\uparrow 12.1}$\\
\hline
YOLOX-SwinB & 135.14M & 337.03 & 11.8 &  $86.5$&$90.0$&$86.8$&$70.1$\\
YOLOV-SwinB & 143.38M & 363.60 & 13.9  &  $89.7_{\uparrow 3.2}$&$92.4_{\uparrow 2.4}$    &$89.7_{\uparrow 2.9}$&$79.9_{\uparrow 9.8}$      \\
YOLOV++-SwinB & 150.89M & 364.74 & 15.9 &  $90.7_{\uparrow 4.2}$&$92.1_{\uparrow 2.1}$&$90.6_{\uparrow 3.8}$&$80.1_{\uparrow 10.0}$\\
\hline
YOLOX-FocalL & 257.42M & 576.96 & 25.7 &  $89.7$&$92.4$&$90.4$&$76.6$\\
YOLOV-FocalL & 262.78M  & 593.90 & 27.1  &$92.5_{\uparrow 2.8}$& $95.4_{\uparrow 3.0}$&$92.2_{\uparrow 1.8}$&$83.9_{\uparrow 7.3}$\\
YOLOV++-FocalL & 267.11M & 594.76 & 27.6 &  $92.9_{\uparrow 3.2}$&$95.4_{\uparrow 3.0}$&$92.9_{\uparrow 2.5}$&$84.1_{\uparrow 7.5}$\\
\hline
\end{tabular}
\end{center}
\end{table*}
\setlength{\tabcolsep}{1.4pt}

\subsection{Validation on Different Model Sizes}
We further conduct a comprehensive comparison between the base model YOLOX, our previous version YOLOV, and the updated version YOLOV++ with different model sizes in terms of parameters, GFLOPs, inference time, and accuracy, as detailed in Tab.~\ref{table:effectiveness of diff model}. Although the newly updated FAM in the regression branch, FAM$_r$, slightly increases the parameter count and computational load, it contributes to an approximate 1\% increase in AP50. Additionally, we report the accuracy of object detection under various motion speeds, following the split used in previous work~\cite{zhu2017flow}. It is evident that YOLOV++ can further improve the detection accuracy across all of three motion speeds compared to YOLOV. As the objects move faster, the advantage of YOLOV++ becomes more obvious. Particularly, at the fast speed, YOLOV++ achieves around 10\% increase in mAP compared to the base detector. 

\setlength{\tabcolsep}{4pt}
\begin{table}[t]
\begin{center}
\caption{Effectiveness of our strategy on other bases}
\label{table:effectiveness of diff detector}
\begin{tabular}{l|ccc}
\hline\noalign{\smallskip}
Model&Params&GFLOPs&AP50 ($\%$)\\
\noalign{\smallskip}
\hline
\noalign{\smallskip}
PPYOLOE-S & 6.71M & 12.40 &$69.5$\\
PPYOLOEV-S& 8.04M & 16.73 &$74.9_{\uparrow 5.4}$\\
PPYOLOEV++-S & 9.37M & 17.16 & $75.6_{\uparrow 6.1}$\\
\hline
PPYOLOE-L & 50.35M & 89.19  &  $76.9$\\
PPYOLOEV-L & 55.63M & 105.49 & $82.0_{\uparrow 5.1}$\\
PPYOLOEV++-L & 60.91M & 106.35 &  $82.9_{\uparrow 6.0}$ \\
\hline
FCOS & 31.00M & 102.84& $67.0$\\
FCOSV & 36.28M & 120.04 & $73.1_{\uparrow 6.1}$\\
FCOSV++ & 41.56M & 120.90 & $74.1_{\uparrow 7.1 }$\\
\hline
\end{tabular}
\end{center}
\end{table}
\setlength{\tabcolsep}{1.4pt}

\subsection{Validation on Other Base Detectors}
In order to validate the generalization ability of the proposed strategy, we also try it on other widely-used one-stage detectors including PPYOLOE~\cite{xu2022pp} and FCOS~\cite{tian2019fcos}. For PPYOLOE with varying channel numbers across different FPN levels, we standardize the channels in the detection head at all scales to facilitate multi-scale feature aggregation. In the case of FCOS that originally includes 5 FPN levels to handle large size images (\emph{e.g.}, $1333\times 800$), we adapt the architecture to maintain 3 FPN levels with a maximum downsampling rate of 32 to better suit the ImageNet VID dataset. We use the training procedures and hyper-parameter settings from YOLOX for consistency. As shown in Tab.~\ref{table:effectiveness of diff detector}, our strategy consistently enhances the performance of different base detectors, gaining over 6\% in AP50. Compared to our previous version, the enhancements in the FSM and IoU score refinement jointly contribute to about 1\% improvement in AP50. It is worth noting that further tuning of hyper-parameters specific to each base detector could potentially yield even better results.

\setlength{\tabcolsep}{4pt}
\begin{table}[t]
\begin{center}
\caption{Evaluation on the OVIS validation set}
\label{table:validation on ovis}
\begin{tabular}{l|ccc}
\hline\noalign{\smallskip}
Model&AP ($\%$)&AP50 ($\%$)&AP75 ($\%$)\\
\noalign{\smallskip}
\hline
\noalign{\smallskip}
YOLOX-S & $39.0$ & $59.8$ & $40.4$\\
YOLOV-S & $41.7_{\uparrow 2.7}$ & $66.1_{\uparrow 6.3} $ & $43.3_{\uparrow 2.9}$\\
YOLOV++-S & $42.9_{\uparrow 3.9}$ & $69.4_{\uparrow 9.6}$ & $43.7_{\uparrow 3.3}$\\
\hline
YOLOX-SwinT & $49.3$ & $72.9$ & $51.8$\\
YOLOV-SwinT & $52.0_{\uparrow 2.7}$ & $77.5_{\uparrow 4.5}$ & $55.0_{\uparrow 3.2}$\\
YOLOV++-SwinT & $53.2_{\uparrow 3.9}$ & $79.2_{\uparrow 6.3}$ & $56.1_{\uparrow 4.3}$\\
\hline
\end{tabular}
\end{center}
\end{table}
\setlength{\tabcolsep}{1.4pt}

\subsection{Additional Experiments on the OVIS Dataset}
In addition to the ImageNet VID dataset, we also test our models on the Occluded Video Instance Segmentation (OVIS) dataset~\cite{qi2022occluded}. This dataset comprises 607 training videos and 140 validation videos, spanning 25 classes. Characterized by an average of 4.72 objects per frame and a significant prevalence of severe occlusions, the OVIS dataset presents a more challenging set of scenarios for VOD.
For the training settings on the OVIS dataset, we fine-tune the COCO pre-trained weights for 10 epochs with batch size of 8 across 4 GPUs. Subsequently, the FAM is fine-tuned for additional 7 epochs. During multi-scale training, the images are randomly resized from $480 \times 720$ to $800 \times 1200$, with a consistent stride of 32 on the shorter side. For the evaluation phase, the images are uniformly resized to $640 \times 960$. All the other experimental settings are kept consistent with those used for the ImageNet VID dataset.

\begin{figure*}[hp]

    \centering
    \subfloat{\includegraphics[width=0.24\linewidth,height=0.12\paperheight]{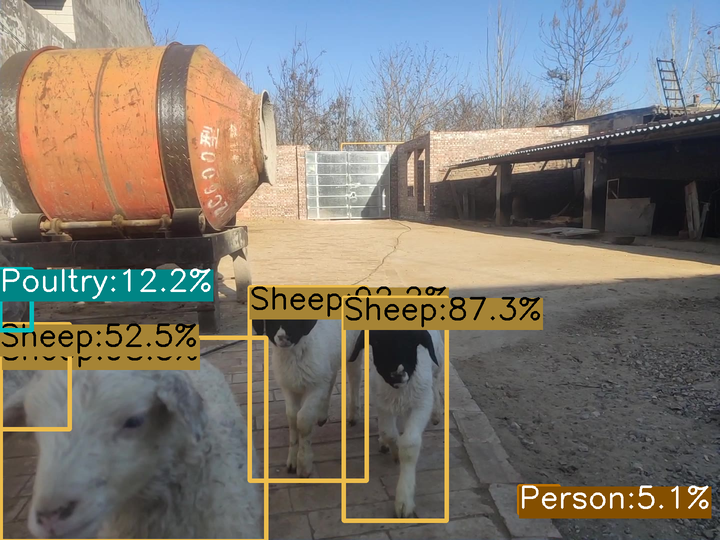}}\hfill
    \subfloat{\includegraphics[width=0.24\linewidth,height=0.12\paperheight]{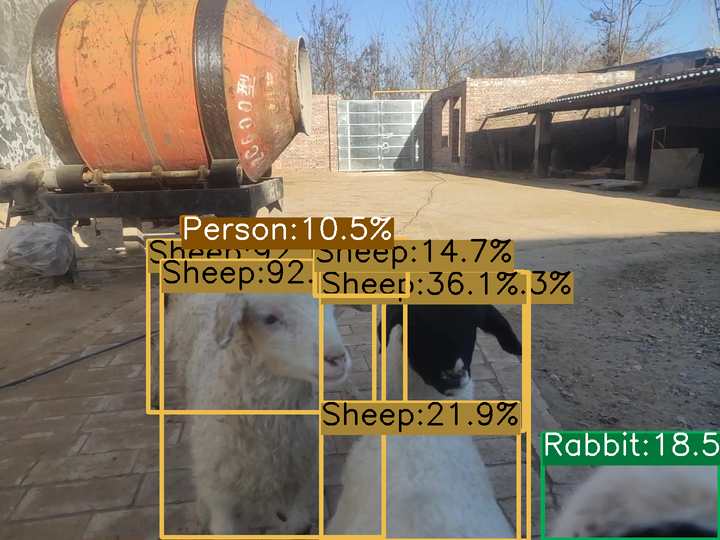}}\hfill
    \subfloat{\includegraphics[width=0.24\linewidth,height=0.12\paperheight]{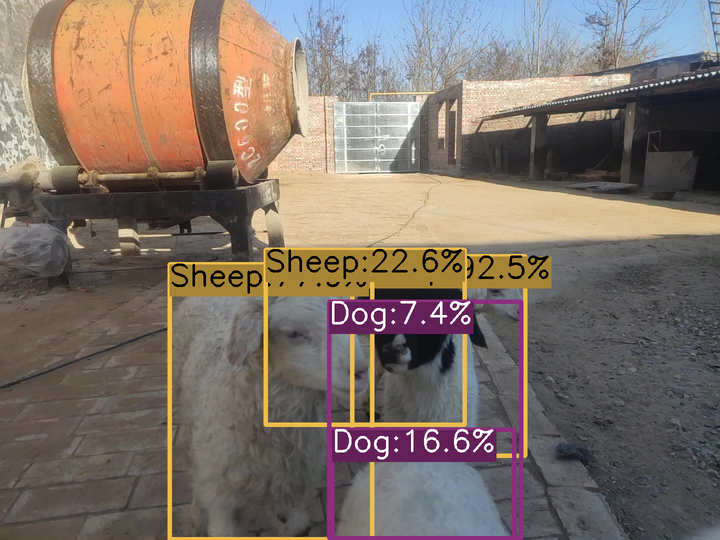}}\hfill
    \subfloat{\includegraphics[width=0.24\linewidth,height=0.12\paperheight]{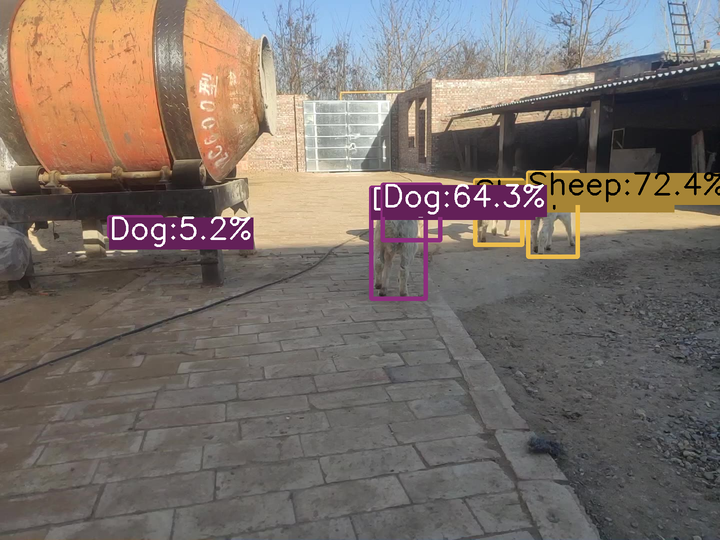}}\\
    \subfloat{\includegraphics[width=0.24\linewidth,height=0.12\paperheight]{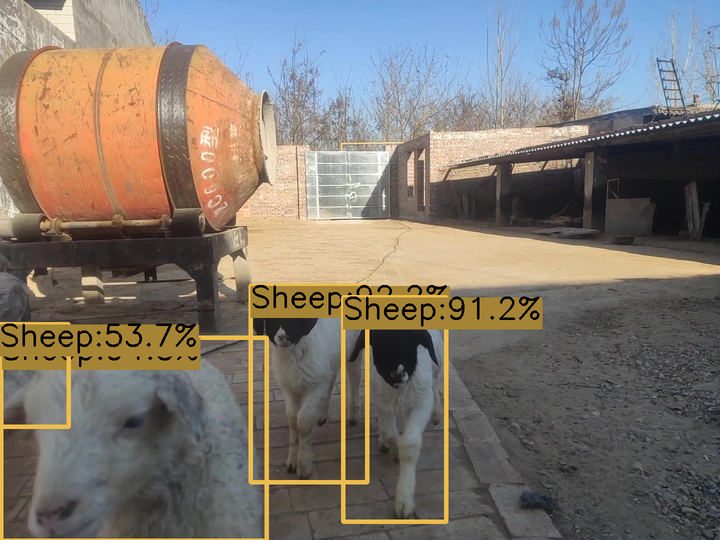}}\hfill
    \subfloat{\includegraphics[width=0.24\linewidth,height=0.12\paperheight]{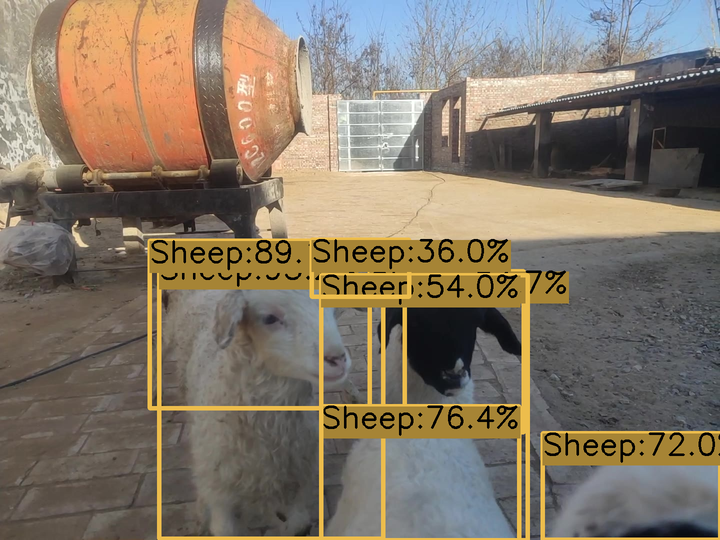}}\hfill
    \subfloat{\includegraphics[width=0.24\linewidth,height=0.12\paperheight]{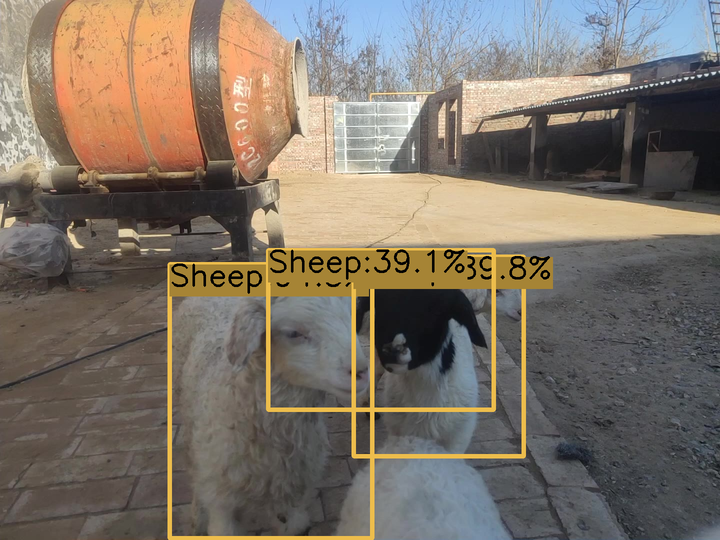}}\hfill
    \subfloat{\includegraphics[width=0.24\linewidth,height=0.12\paperheight]{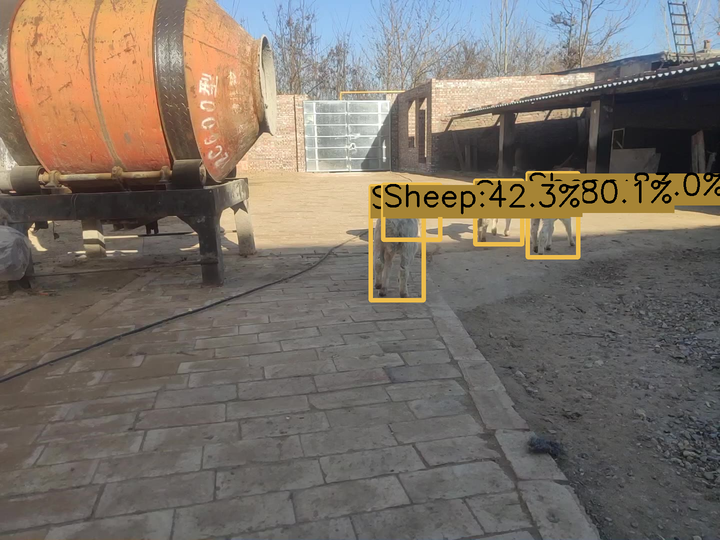}}
    \caption*{(a) Occluded Sheep and Dog}

    \subfloat{\includegraphics[width=0.24\linewidth,height=0.12\paperheight]{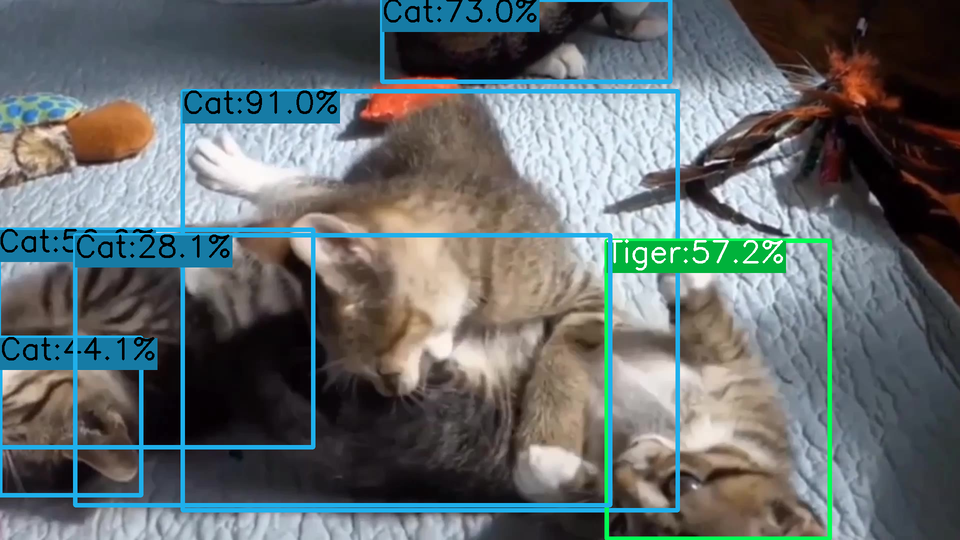}}\hfill
    \subfloat{\includegraphics[width=0.24\linewidth,height=0.12\paperheight]{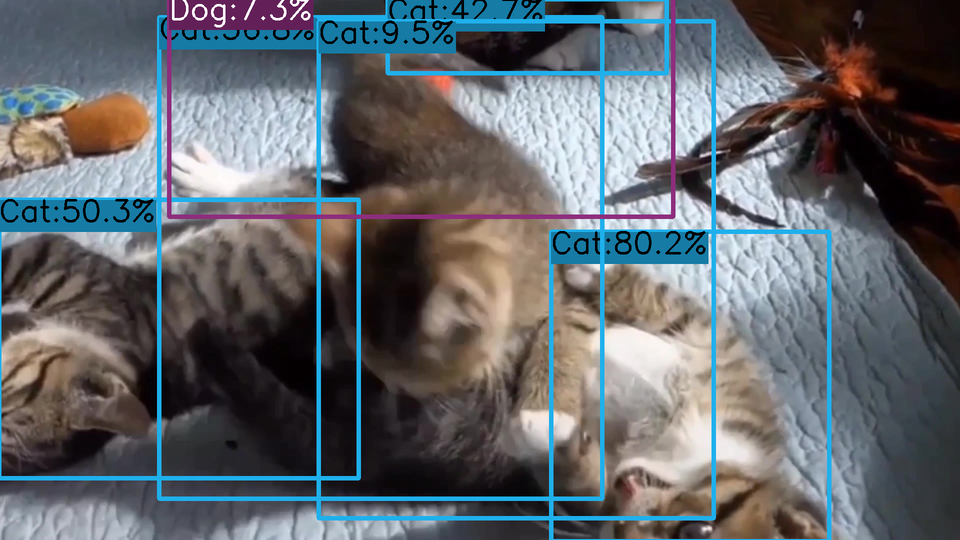}}\hfill
    \subfloat{\includegraphics[width=0.24\linewidth,height=0.12\paperheight]{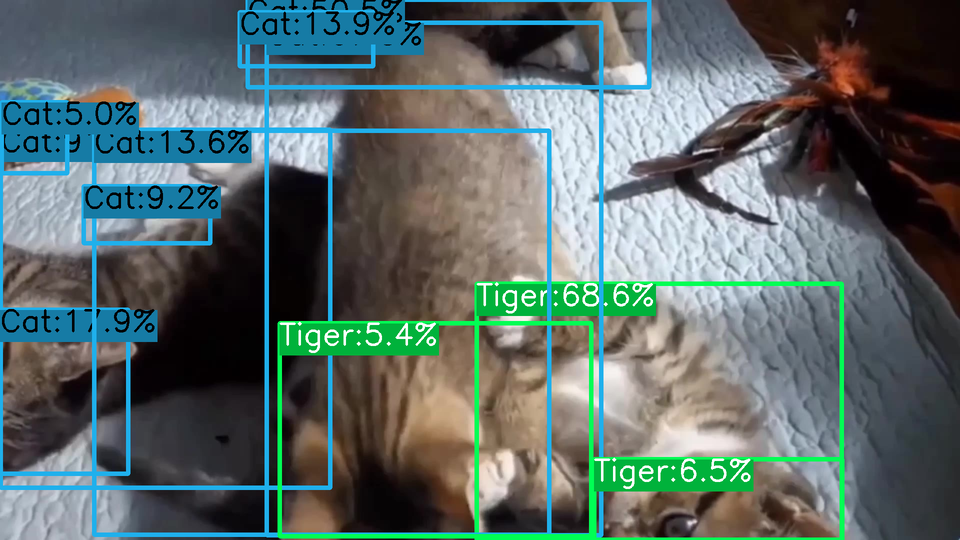}}\hfill
    \subfloat{\includegraphics[width=0.24\linewidth,height=0.12\paperheight]{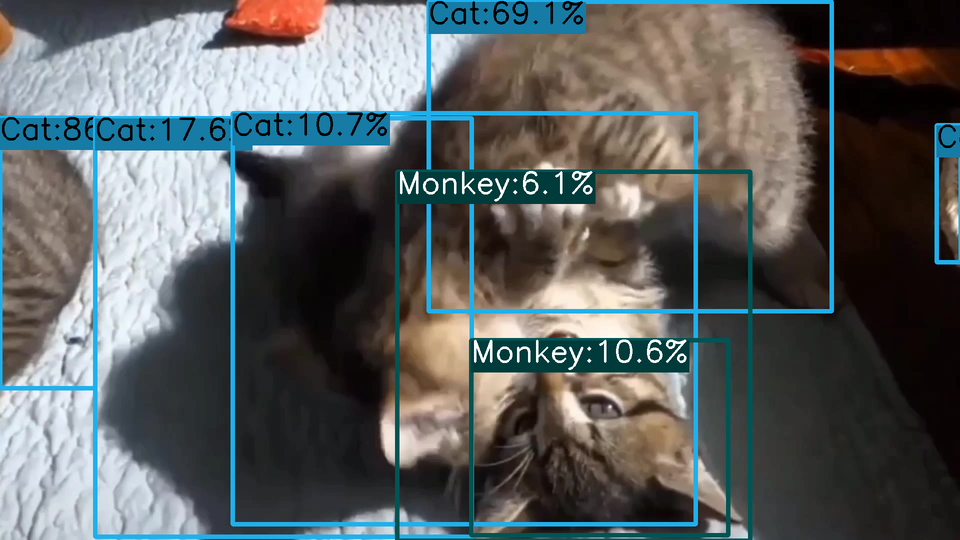}}\\
    \subfloat{\includegraphics[width=0.24\linewidth,height=0.12\paperheight]{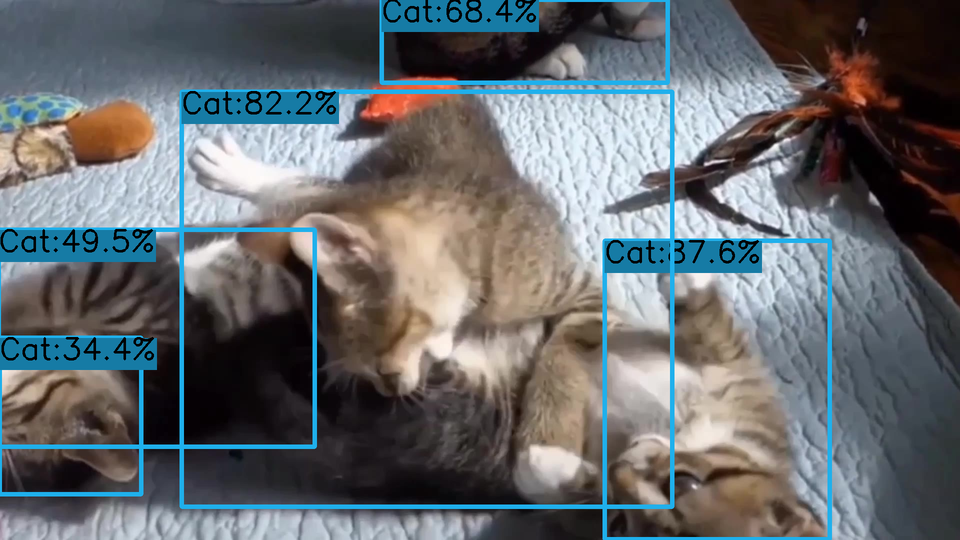}}\hfill
    \subfloat{\includegraphics[width=0.24\linewidth,height=0.12\paperheight]{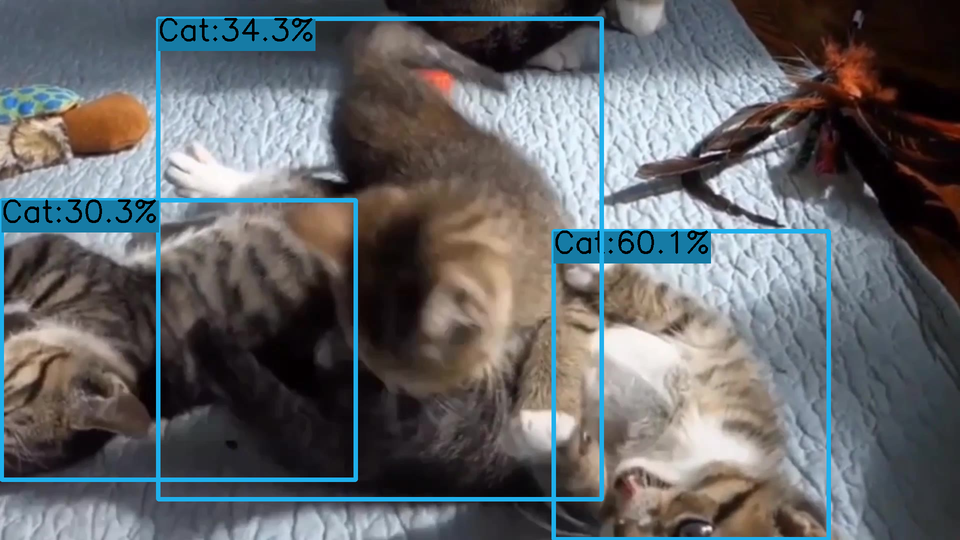}}\hfill
    \subfloat{\includegraphics[width=0.24\linewidth,height=0.12\paperheight]{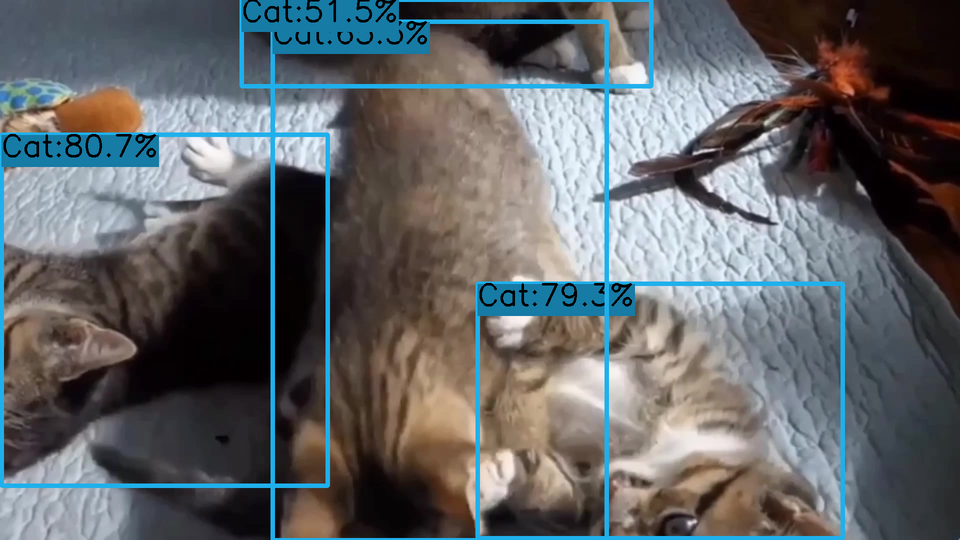}}\hfill
    \subfloat{\includegraphics[width=0.24\linewidth,height=0.12\paperheight]{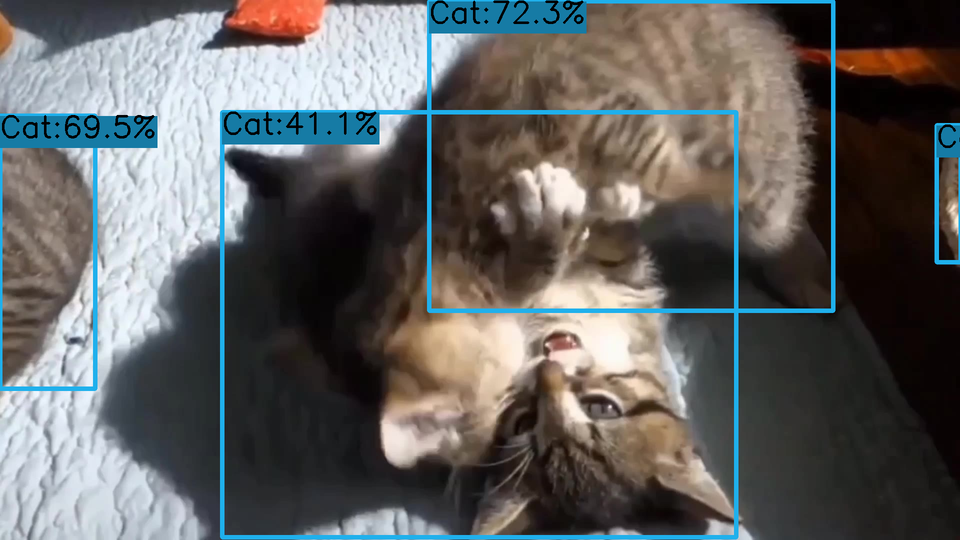}}
    \caption*{(b) Occluded Cat and Tiger}

    \subfloat{\includegraphics[width=0.24\linewidth,height=0.12\paperheight]{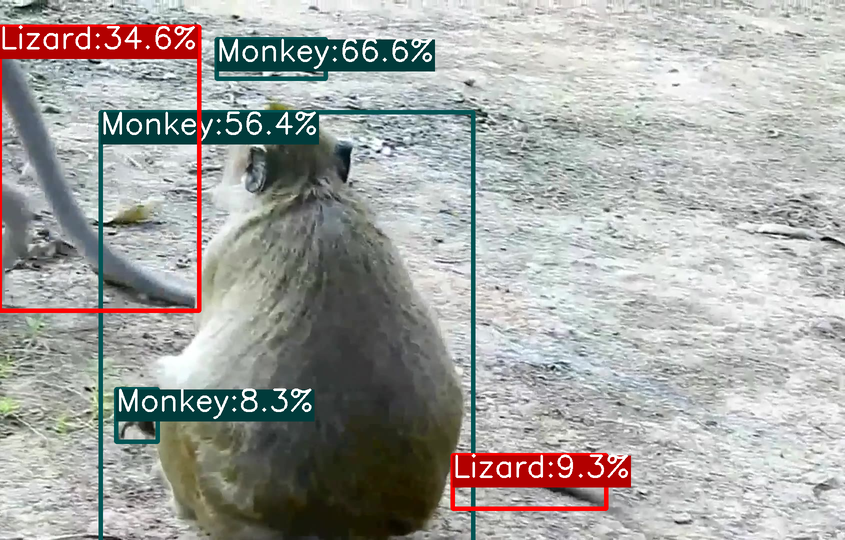}}\hfill
    \subfloat{\includegraphics[width=0.24\linewidth,height=0.12\paperheight]{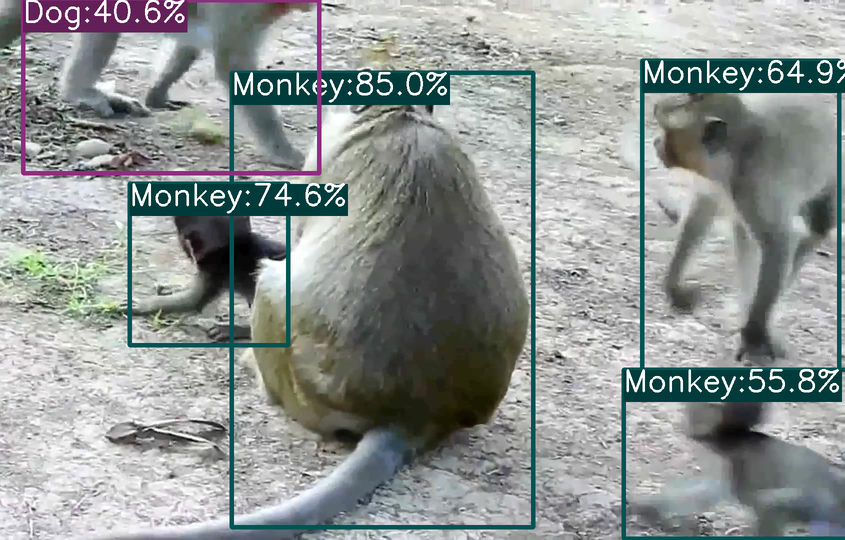}}\hfill
    \subfloat{\includegraphics[width=0.24\linewidth,height=0.12\paperheight]{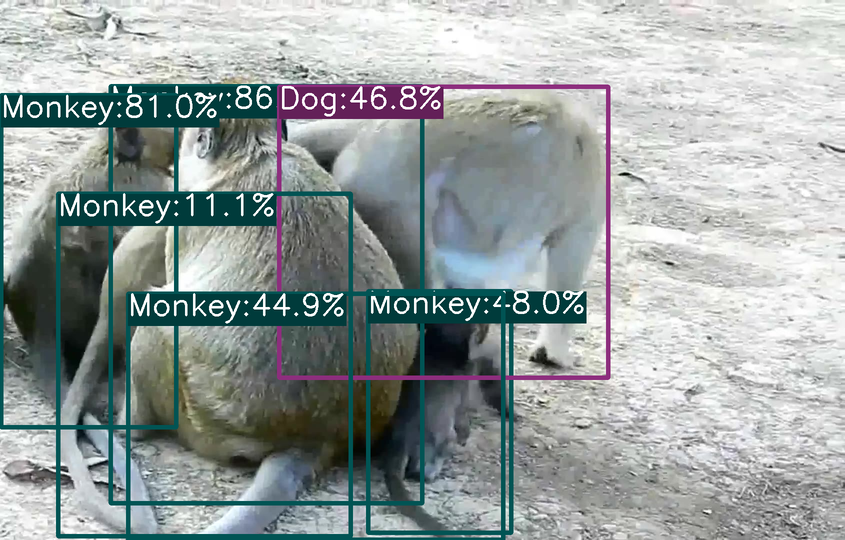}}\hfill
    \subfloat{\includegraphics[width=0.24\linewidth,height=0.12\paperheight]{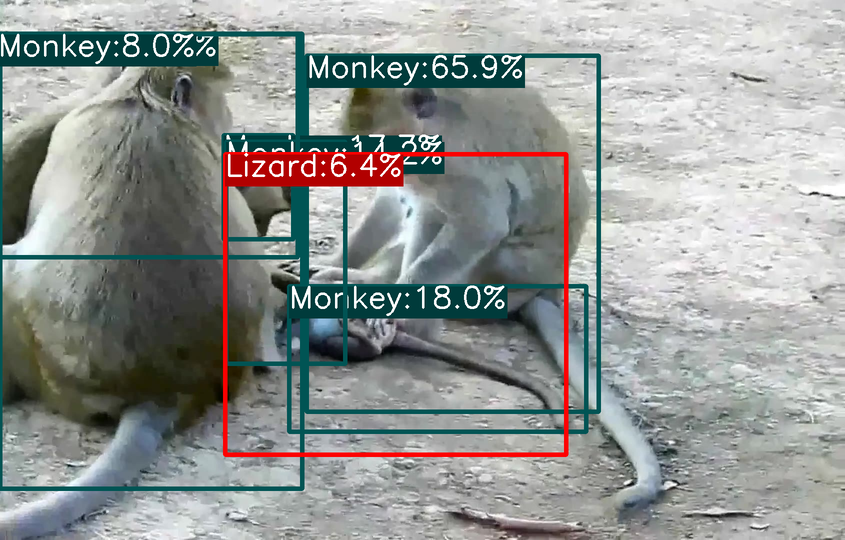}}\\
    \subfloat{\includegraphics[width=0.24\linewidth,height=0.12\paperheight]{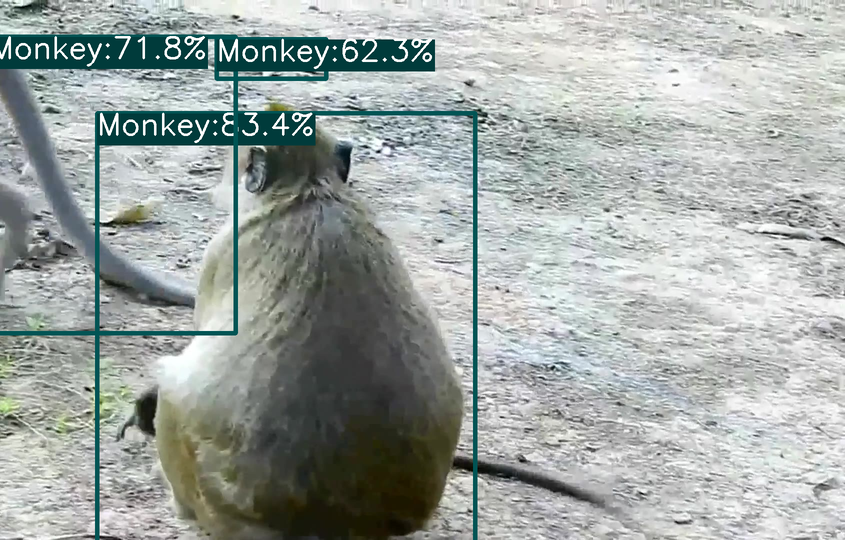}}\hfill
    \subfloat{\includegraphics[width=0.24\linewidth,height=0.12\paperheight]{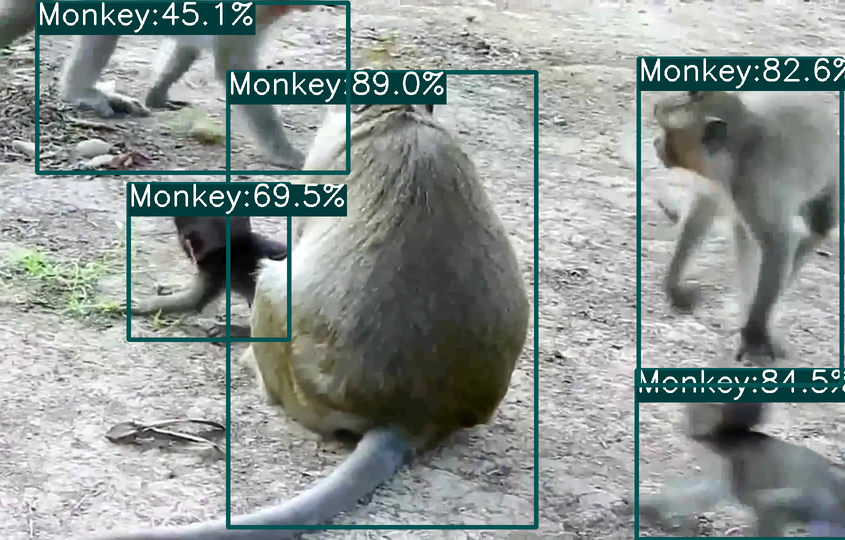}}\hfill
    \subfloat{\includegraphics[width=0.24\linewidth,height=0.12\paperheight]{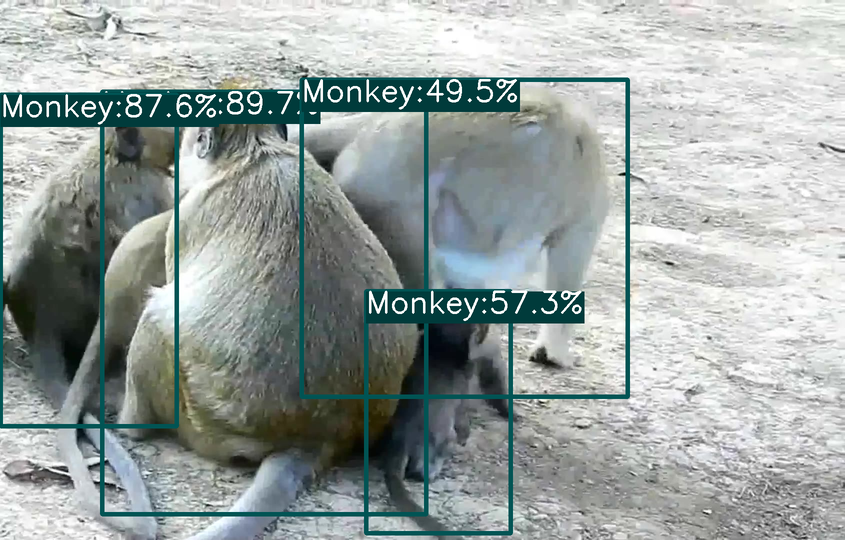}}\hfill
    \subfloat{\includegraphics[width=0.24\linewidth,height=0.12\paperheight]{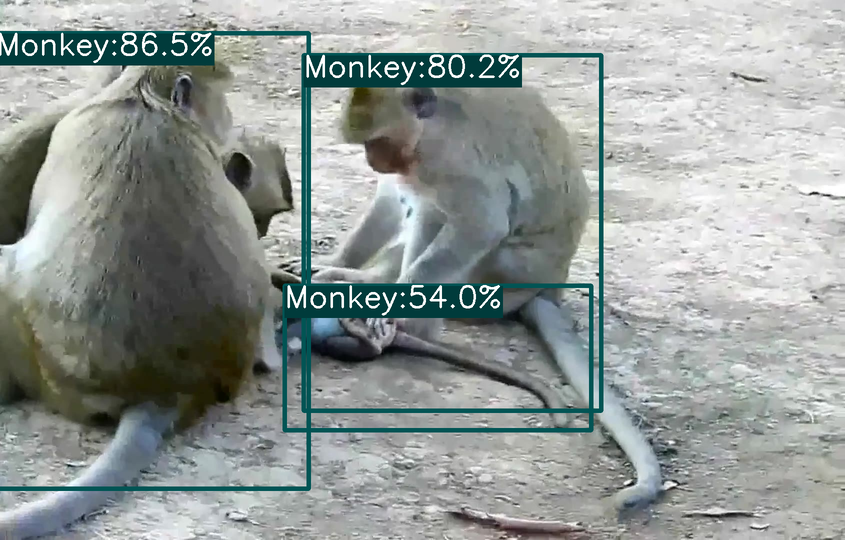}}
    \caption*{(c) Occluded Monkey and Lizard }

    \caption{Visual comparisons between the base detector (upper row) and ours (lower row) on the OVIS dataset.}
    \label{fig:ovis_comparision}
\end{figure*}

Table~\ref{table:validation on ovis} gives the comparison between YOLOV++, YOLOV, and the base detectors on the OVIS validation set. The baseline YOLOX-S model obtains 39.0\% AP. Our previous version YOLOV-S increases YOLOX-S to 41.7\% AP, while YOLOV++-S further elevates the AP to 42.9\%. With the more robust SwinTiny backbone, the base detector YOLOX-SwinT starts at 49.3\% AP. YOLOV-SwinT and  YOLOV++-SwinT respectively achieve 52.0\% and 53.2\% in AP, respectively. These results clearly reflect the advance of our designs over the baseline YOLOX model.
Additionally, we present a qualitative comparison in Fig.~\ref{fig:ovis_comparision}. It is evident that our YOLOV++ model has a distinct advantage over the base detector in scenarios with objects heavily occluded. 

\section{Conclusion}

This paper has built a practical video object detector that jointly considers the detection accuracy and inference efficiency through a feature selection and aggregation manner. A feature aggregation module was designed to effectively absorb temporal information across frames for improving the detection accuracy. While for saving computational resources, different from existing two-stage detectors, we proposed to put the region selection after the (rough) prediction. This subtle change makes our detectors significantly more efficient. 
Experiments and ablation studies have been carried out to verify the effectiveness of our strategy, and its advance over previous arts. It is worth to emphasize that our model has achieved a new record performance, \emph{i.e.}, 92.9\% AP50 at over 30 FPS
on the ImageNet VID dataset on a single 3090 GPU. The core idea is simple and general, which can potentially inspire further research works and broaden the applicable scenarios related to video object detection.

\section*{Data Availability Statement}
The data used in this study can be obtained from the following sources:

\textbf{ImageNet VID and DET}~\cite{russakovsky2015imagenet} support the results in Figures 1, 2, 5, 7, and Tables 1-8. The official page, \emph{i.e.}, the ImageNet Large Scale Visual Recognition Challenge 2015 (ILSVRC2015), can be visited at \url{https://image-net.org/challenges/LSVRC/2015/2015-downloads.php}.

\textbf{OVIS}~\cite{qi2022occluded} is employed to produce the results in Figure 8 and Table 9, which can be found at \url{https://songbai.site/ovis/}.
\bibliographystyle{spmpsci}      
\bibliography{reference}   

%
%

\end{document}